\documentclass{IEEEtaes}

\usepackage{color,amsthm}
\usepackage{array}
\usepackage[table]{xcolor}
\usepackage{graphicx}
\usepackage{siunitx}
\usepackage{amsfonts}
\usepackage{caption}
\usepackage{multirow}
\usepackage{orcidlink}
\usepackage[export]{adjustbox}
\usepackage[noadjust]{cite}
\usepackage{subcaption}

\usepackage{tikzit}

\tikzstyle{mlpstyle}=[fill=cyan!30, draw=black, shape=rectangle, minimum width=1cm, minimum height=1cm, align=center, font={\tiny}]
\tikzstyle{emb}=[fill=red!30, draw=black, shape=rectangle, minimum width=0.75cm, minimum height=1cm, align=center, font={\tiny}]
\tikzstyle{enc}=[fill=green!30, draw=black, shape=rectangle, minimum width=0.75cm, minimum height=1cm, align=center, font={\tiny}]

\tikzstyle{nn_arrow}=[->, >=stealth, thick]
\tikzstyle{nn_connector}=[-, fill=none, thick]

\newcommand{\imgbox}[1]{\includegraphics[width=\linewidth]{#1}}

\begin{document}

\title{Construction of Digital Terrain Maps from Multi-view Satellite Imagery using Neural Volume Rendering}

\author{Josef X. Biberstein\,\orcidlink{0009-0006-9245-3972}}
\affil{Massachusetts Institute of Technology, Cambridge, MA 02139, USA}

\author{Guilherme Cavalheiro}
\affil{Massachusetts Institute of Technology, Cambridge, MA 02139, USA}

\author{Juyeop Han\, \orcidlink{0000-0002-7853-3451}}
\affil{Massachusetts Institute of Technology, Cambridge, MA 02139, USA}

\author{Sertac Karaman\, \orcidlink{0000-0002-2225-7275}}
\member{Member, IEEE}
\affil{Massachusetts Institute of Technology, Cambridge, MA 02139, USA}

\receiveddate{This work has been submitted to the IEEE for possible publication. 
Copyright may be transferred without notice, after which this version may no longer be accessible.
This work was supported in part by the Defense Science and Technology Agency (DSTA) of Singapore.}

\corresp{{\itshape (Corresponding author: Josef X. Biberstein.)}}

\authoraddress{Authors' addresses: Josef X. Biberstein is with Massachusetts Institute of Technology, Cambridge, MA 02139, USA (e-mail: \href{mailto:jxb@mit.edu}{jxb@mit.edu}). Guilherme Cavalheiro is with Massachusetts Institute of Technology, Cambridge, MA 02139, USA (e-mail: \href{mailto:guivecna@mit.edu}{guivecna@mit.edu}). Juyeop Han is with Massachusetts Institute of Technology, Cambridge, MA 02139, USA (e-mail: \href{mailto:juyeop@mit.edu}{juyeop@mit.edu}). Sertac Karaman is with Massachusetts Institute of Technology, Cambridge, MA 02139, USA (e-mail: \href{mailto:sertac@mit.edu}{sertac@mit.edu}).}

\maketitle

\begin{abstract}
  Digital terrain maps (DTMs) are an important part of planetary exploration, enabling operations such as terrain relative navigation during entry, descent, and landing for spacecraft and aiding in navigation on the ground.
  As robotic exploration missions become more ambitious, the need for high quality DTMs will only increase.
  However, producing DTMs via multi-view stereo pipelines for satellite imagery, the current state-of-the-art, can be cumbersome and require significant manual image preprocessing to produce satisfactory results.
  In this work, we seek to address these shortcomings by adapting neural volume rendering techniques to learn textured digital terrain maps directly from satellite imagery.
  Our method, neural terrain maps (NTM), only requires the locus for each image pixel and does not rely on depth or any other structural priors.
  We demonstrate our method on both synthetic and real satellite data from Earth and Mars encompassing scenes on the order of \qty{100}{\kilo\meter\squared}.
  We evaluate the accuracy of our output terrain maps by comparing with existing high-quality DTMs produced using traditional multi-view stereo pipelines.
  Our method shows promising results, with the precision of terrain prediction almost equal to the resolution of the satellite images even in the presence of imperfect camera intrinsics and extrinsics.
\end{abstract}


\section{INTRODUCTION}
Recent and near future planetary exploration missions are pushing new bounds in terms of their scale and complexity.
For example, the Perseverance rover, launched in 2020, carried with it the Ingenuity helicopter~\cite{balaram_ingenuity_2021}, which demonstrated the first powered flight on another planet.
Following on, the upcoming Dragonfly mission~\cite{lorenz_dragonfly_2018} to Titan seeks to conduct long, autonomous flights during its exploration of the moon.
Missions are also embracing more unconventional designs, such as EELS~\cite{vaquero_eels_2024}, and architectures, such as CADRE~\cite{de_la_croix_multi-agent_2024}.
Additionally, the entry of commercial companies into space exploration~\cite{schonfeld_summary_2023} is set to drastically increase the number of missions launched in the coming decades.
\par
For all of these missions, access to high quality digital terrain maps (DTMs) is key for success.
Whether for terrain relative navigation during entry, descent, and landing~\cite{michaelson_terrain-relative_2024,jung_digital_2020} or for aiding navigation while traversing a planetary surface~\cite{kim_autonomous_2004}, high quality DTMs enable missions when other methods of precise localization, such as GPS, are unavailable.
However, current methods of producing DTMs from satellite imagery with multi-view stereo (MVS) reconstruction can be cumbersome to use, requiring significant image preprocessing to get high quality results and lacking the ability to utilize the gamut of satellite imagery available due to struggling with effects such as differing scene illumination, presence of transient phenomena, or differing camera models between images.
Therefore, there is a desire to explore alternate DTM construction methods to support the increased need for these products in future space missions.
\par
\begin{figure}[htbp]
  \centering
  \begin{subfigure}{\linewidth}
    \centering
    \includegraphics[width=\linewidth]{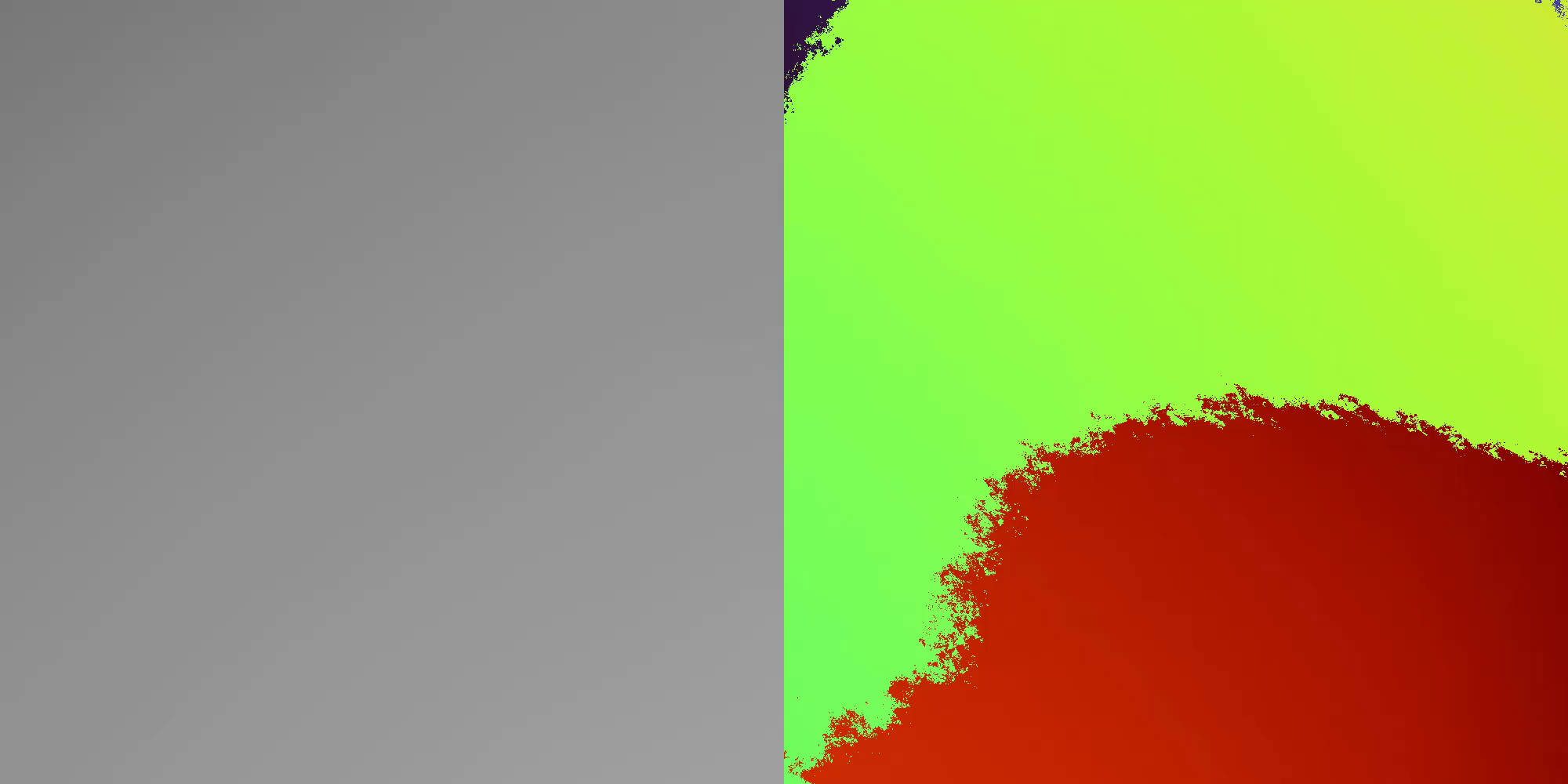}
    \caption{Nerfacto~\cite{tancik_nerfstudio_2023}}
    \label{fig:subfig1}
  \end{subfigure}
  \vspace{2mm}
  \begin{subfigure}{\linewidth}
    \centering
    \includegraphics[width=\linewidth]{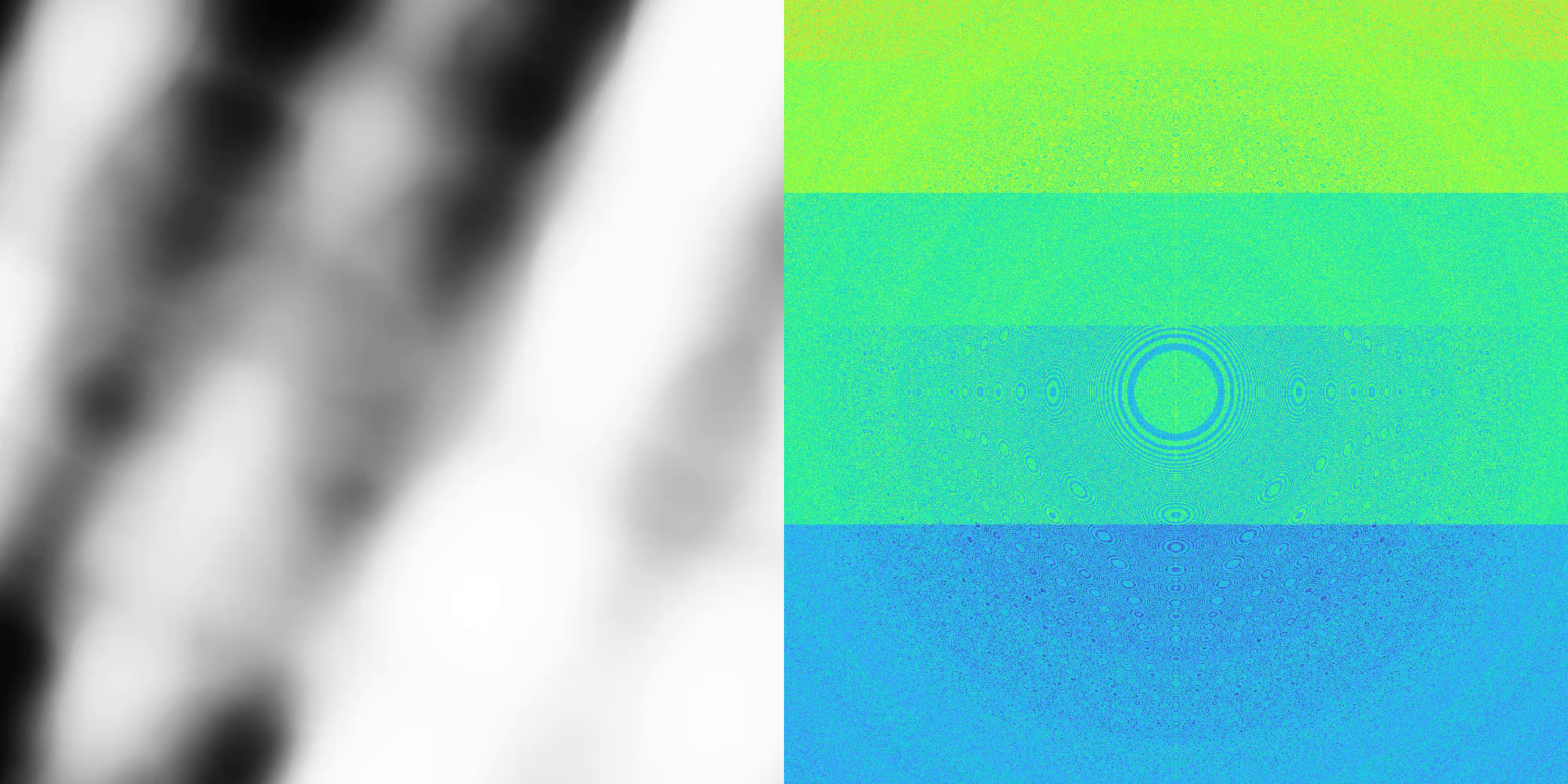}
    \caption{NeuS~\cite{wang_neus_2023}}
    \label{fig:subfig2}
  \end{subfigure}
  \vspace{2mm}
  \begin{subfigure}{\linewidth}
    \centering
    \includegraphics[width=\linewidth]{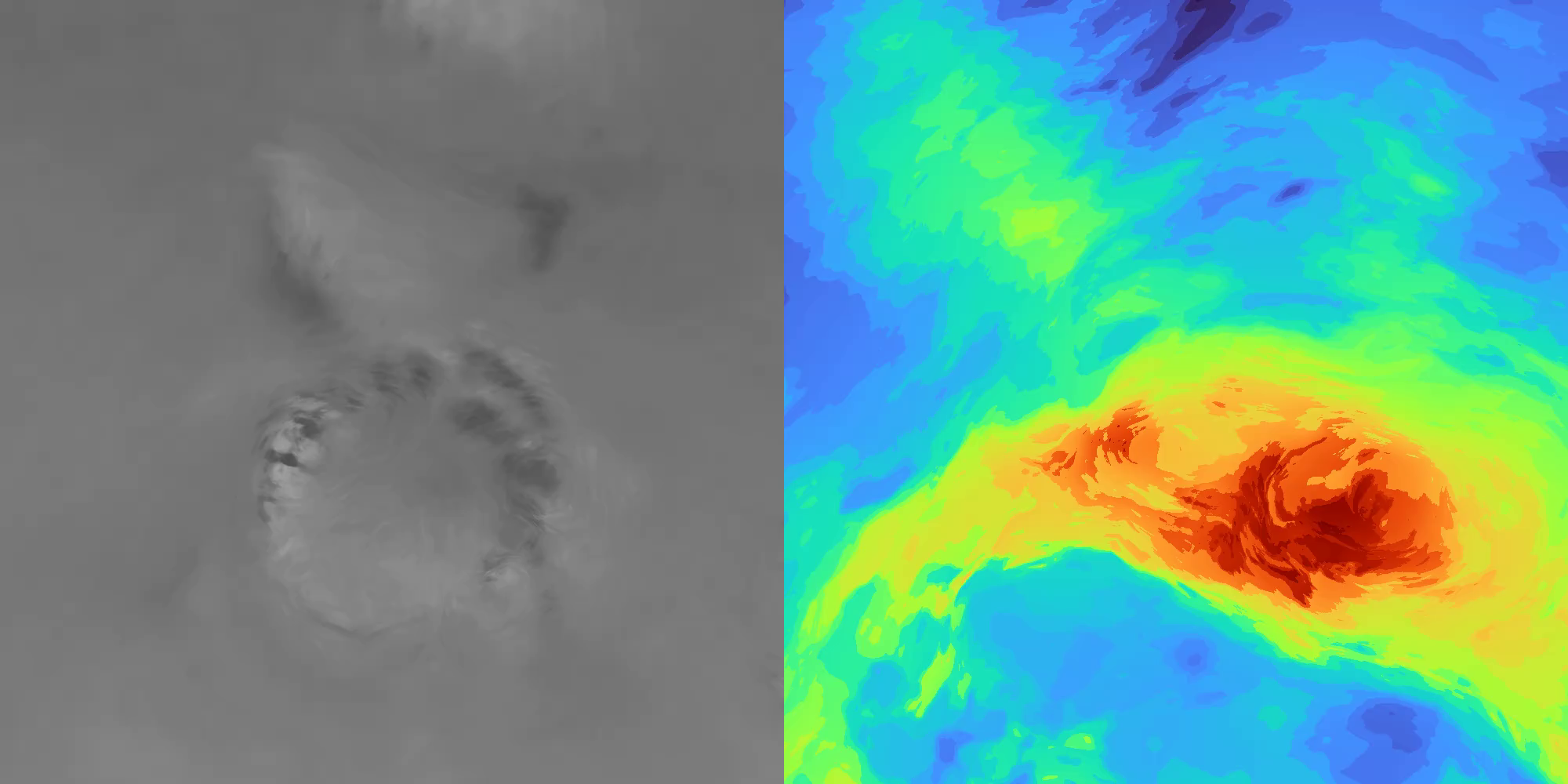}
    \caption{NEMo~\cite{dai_neural_2024}}
    \label{fig:subfig3}
  \end{subfigure}
  \vspace{2mm}
  \begin{subfigure}{\linewidth}
    \centering
    \includegraphics[width=\linewidth]{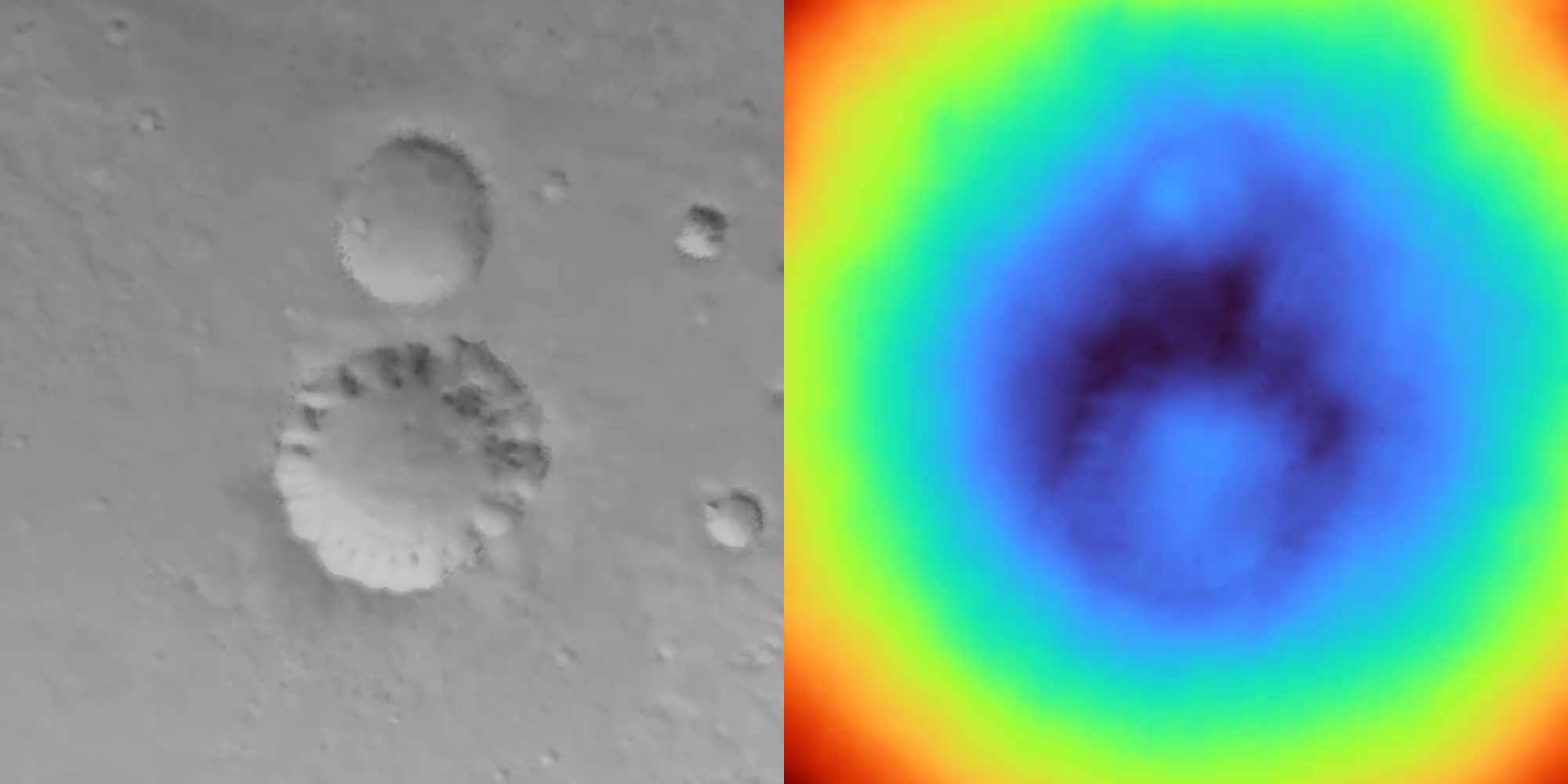}
    \caption{\textbf{NTM (ours)}}
    \label{fig:subfig4}
  \end{subfigure}
  \caption{A comparison of different neural volume rendering methods visual (left) and normalized depth (right) reconstruction on a closeup of a small crater in the Gale Crater CTX scene.
    The view is a nadir perspective view at approximately \qty{19}{\kilo\meter} altitude with a \qty{20}{\degree} field of view.
  Previous methods fail to produce a coherent reconstruction of the terrain due to the low disparity signal in the scene, while NTM succeeds.}
  \label{fig:other-method-compare}
\end{figure}

Recently, neural volume rendering methods, such as neural radiance fields (NeRFs)~\cite{bell_iii_calibration_2013}, have become a popular method for scene representation and novel view reconstruction tasks.
These methods excel at fusing large amounts of data from various sources into high quality representations but, as demonstrated in Fig.~\ref{fig:other-method-compare}, struggle with planar image distributions with low disparity as often seen in satellite imagery of terrain.
Therefore, in this work, we explore how neural volume rendering can be adapted to address this issue and to provide a new way of generating DTMs while avoiding the pitfalls of typical MVS pipelines.
Out contributions are threefold:
\begin{enumerate}
  \item We introduce Neural Terrain Maps (NTM), a volume rendering procedure that can be used to directly optimize textured DTMs from multi-view satellite imagery.
  \item We integrate NTM into the popular Nerfstudio framework~\cite{tancik_nerfstudio_2023} to enable rapid experimentation and development.
  \item We demonstrate the effectiveness of NTM by creating DTMs using datasets from two different orbital imaging instruments and compare the results against simulated imagery and existing high quality reference DTMs.
\end{enumerate}
The rest of this article is organized as follows.
In Section~\ref{sec:related-work}, we review related work in terrain mapping and neural volume rendering.
In Section~\ref{sec:methodology}, we describe the rendering procedure for NTM.
In Section~\ref{sec:implementation}, we provide the implementation details of the networks used in NTM, and also describe the datasets used and how they are processed.
In Section~\ref{sec:results}, we present quantitative and qualitative results of our method on both synthetic and real satellite imagery datasets, including renderings of novel views and georeferenced DTM outputs.
Finally, in Section~\ref{sec:conclusion}, we conclude and outline future work.

\section{RELATED WORK}\label{sec:related-work}

\subsection{Terrain Mapping with Multi-view Stereo}
The use of traditional MVS techniques to produce DTMs from satellite imagery has been extensively studied.
Some established software solutions have emerged such as the Ames Stereo Pipeline~\cite{beyer_ames_2018}, the satellite stereo pipeline (S2P)~\cite{de_franchis_automatic_2014}, and COLMAP~\cite{schonberger_pixelwise_2016,schonberger_structure--motion_2016}.
One significant challenge that arises with these methods is that satellite imagery is not typically provided with a perspective camera model as it is usually acquired from linescan cameras, which capture images in a series of lines as the satellite moves along its orbit.
Therefore, either a linescan model is used or a rational polynomial coefficient (RPC) model is used to abstract away the camera model by fitting a rational polynomial function from latitude and longitude points to pixel coordinates or vice versa.
Because of this, MVS on satellite imagery cannot rely on the standard epipolar geometry assumptions that are typically used in stereo matching.
This can be solved by trying to match directly with non-linear epipolar curves in linescan image pairs~\cite{lee_extraction_2003}, approximating the linescan camera model with a perspective camera model locally and proceeding with normal stereo rectification and matching~\cite{wang_epipolar_2011}, or constructing epipolar curves for each image in a piecewise manner~\cite{oh_piecewise_2010}.
Learning methods can also be used to overcome this issue, either by replacing the disparity calculation step of the pipeline~\cite{zhang_ga-net_2019} or by learning MVS stereo end to end~\cite{gu_cascade_2020}.
\par
Traditional MVS methods also struggle with dynamic qualities of a scene, such as changes in lighting between images and transient phenomena such as clouds and seasonal changes.
Learning methods using GAN~\cite{singh_cloud-gan_2018,pan_cloud_nodate} and transformers~\cite{wu_cloudformer_2022,christopoulos_cloudtran_2025} have been proposed for removing clouds in image space.
MVS can be combined with principles of photometry to produce DTMs in the presence of varying lighting conditions.
The so-called Shape-from-Shading (SfS) approach can be used to refine areas of an already-constructed DTM which is in shadow~\cite{alexandrov_multiview_2018} or can be used end-to-end by alternating between stereo matching and SfS during optimization~\cite{langguth_shading-aware_2016}.
Semantic segmentation approaches have been used to help ameliorate the effects of seasonal transients in satellite imagery~\cite{bosch_semantic_2019}.
If enough data is available, it is also possible to avoid this pitfall by intelligently selecting images with less variation to use in the reconstruction process~\cite{facciolo_automatic_2017}.

\subsection{Neural Volume Rendering}
NeRFs have revolutionized the field of inverse graphics, enabling high-quality scene reconstruction from images with known intrinsics and extrinsics.
Since the original work~\cite{mildenhall_nerf_2020}, much progress has been made in improving the training speed, robustness, and quality of NeRFs.
For instance, InstantNGP~\cite{muller_instant_2022} introduced a new encoding scheme based on multi-resolution hash tables which drastically improved training speed, BARF~\cite{lin_barf_2021} introduced a new method of jointly optimizing camera poses and scene representations to improve reconstruction quality, and NeRF-W~\cite{martin-brualla_nerf_2021} introduced appearance and transient embeddings to allow for better handling of dynamic scenes.
NeRFs only implicitly encode the structure of the scene under consideration, which is inconvenient for applications in which, along with novel views, the geometry of the scene needs to be extracted.
Therefore, additional processing steps, such as marching cubes, point cloud tesselation, or using depth images to infer geometry, are required to extract 3D geometry from a trained NeRF.
To address this, a family of methods~\cite{wang_neus_2023,wang_neus2_2023} have been proposed to adapt the volume rendering procedure used in NeRFs to use a signed distance function (SDF) instead of a density field, allowing the surface to be trivially extractable.
Some work has been done on extracting height maps from NeRFs as well, such as NEMo~\cite{dai_neural_2024} which seeks to extract height maps from NeRFs using quantile regression and Neural Height Maps~\cite{logothetis_neural_2024} which uses volume rendering on a height map to improve depth estimation for binocular photometric stereo.
\par
Also particularly germane to this work are NeRFs trained on satellite imagery.
S-NeRF~\cite{derksen_shadow_2021} demonstrates the use of NeRFs to reconstruct urban scenes from satellite imagery in the presence of different lighting conditions by modeling albedo instead of radiance and introducing the dependency of solar direction.
Sat-NeRF~\cite{mari_sat-nerf_2022} and EO-NeRF~\cite{mari_multi-date_2023} extend this idea to model shadows explicitly by adding a single ray tracing step to the rendering procedure, allowing the scene to be relit by physical shadows.
While urban scenes, which often contain tall buildings (which provide significant disparity queues between images), are well suited to NeRFs, large-scale scenes of terrain appear to be more challenging~\cite{prosvetov_illuminating_2025,hansen_analyzing_2024,van_kints_neural_2024}.
These scenes are usually composed of slowly varying terrain features which do not provide clear disparity queues.
This makes it challenging for a regular NeRF, which permits any 3D scene structure, to properly constrain the representation to a height map.
In Fig.~\ref{fig:other-method-compare}, we demonstrate this issue by comparing NTM with three other neural volume rendering methods on a scene from the Gale Crater CTX dataset.
Nerfacto~\cite{tancik_nerfstudio_2023} is a state-of-the-art, general purpose NeRF, NeuS~\cite{wang_neus_2023} is a method for rendering surfaces from SDFs, and NEMo~\cite{dai_neural_2024} is a method for using a NeRF to construct a height map from a scene using quantile regression.
As demonstrated by depth renderings, all three of these methods fail to produce a coherent reconstruction of the terrain in the scene, while NTM succeeds.
As such, we choose to evaluate NTM by comparing it against existing high quality DTMs produced using traditional MVS pipelines.

\section{METHODOLOGY}\label{sec:methodology}

\subsection{Neural Radiance Fields and Volume Rendering}
We begin with a brief explanation of NeRFs and their training as a prelude to our method.
A NeRF typically consists of two fields over the domain of a given three-dimensional scene.
These fields are the color field $c_{\theta}: \mathbb{R}^3 \to \mathbb{R}^d$ and the density field $\sigma_{\lambda}: \mathbb{R}^3 \to \mathbb{R}$ which are each represented by neural networks with parameters $\theta$ and $\lambda$ respectively.
Here, $d$ is the number of color channels, which is three for RGB images and one for grayscale images.
We note that the color field may also be additionally parameterized over the viewing direction $\mathbf{d}$, i.e., $c_{\theta}: \mathbb{R}^3 \times \mathbb{R}^3 \to \mathbb{R}^d$.
However, in this work and for the purpose of large scale satellite terrain mapping, the assumption that color is independent of viewing direction is reasonable and so we omit this dependency.
\par
Given a set of $N$ input training images $I = \{I_n\}_{n=1}^N$ with corresponding camera models $K = \{K_n: \mathbb{Z}^2 \to \mathbb{R}^3 \times \mathbb{R}^3\}_{n=1}^N$ the NeRF can be trained to predict new views of the scene through a differentiable volume rendering procedure.
Here the camera model $K_n$ is a function that maps pixel coordinates to a 3D pixel locus $(\mathbf{o}, \mathbf{d})$ (i.e., a ray with an origin $\mathbf{o}$ and a direction $\mathbf{d}$) in the world space of the scene.
This can be a simple pinhole model or a less common model such as an orthographic camera, a fisheye, or, as is demonstrated in this work, a linescan camera.
To train the network, a random batch of $M$ pixels is sampled from $I$ and their ground truth color $\bar{c}_m$ and loci $\mathbf{r}_m = (\mathbf{o}_m, \mathbf{d}_m)$ are obtained from $I$ and $K$ respectively.
A set of $N_s$ sample points $\mathbf{x}_{m, i} = \mathbf{o}_m + t_i \mathbf{d}_m$ is then sampled along the ray defined by the pixel locus, where $t_i$ can be determined in a number of ways (e.g., uniform sampling, importance sampling, proposal networks~\cite{li_nerfacc_2023}) but are typically preferred to be clustered around existing areas of high density in the scene for sample efficiency.
The color $c_{m, i}$ and density $\sigma_{m, i}$ of each sample point is then determined by evaluating the network.
To formulate the predicted color for each pixel, the density values are first used to calculate integration weights for a quadrature rule
\begin{equation}
  w_i = T_i \left( 1 - \exp(-\sigma_i \delta_i) \right)
\end{equation}
where we drop the subscript $m$ here and subsequently for clarity.
The term $T_i$ is the accumulated transmittance of the ray up to the $i$-th sample point, which is defined as
\begin{equation}
  T_i = \exp\left(-\sum_{j=1}^{i-1} \sigma_j \delta_j\right)
\end{equation}
and $\delta_i = t_{i+1} - t_i$ is the distance between the $i$-th and $(i+1)$-th sample points.
The predicted color for the pixel is then given by a weighted sum of each sample point's color
\begin{equation}
  \hat{c}_m = \sum_{i=1}^{N_{s}} w_i c_i.
\end{equation}
Finally, an appropriate loss function is used to compare the predicted color $\hat{c}_m$ with the ground truth color $\bar{c}_m$.
While MSE is often used, in this work, we use L1 loss, which is empirically found to offer good stability and better robustness to outliers~\cite{wang_neus_2023}
\begin{equation}
  \mathcal{L}_m = \frac{1}{m} \sum_{m \in M} \left\| \hat{c}_m - \bar{c}_m \right\|_1.
\end{equation}

\subsection{Neural Volume Rendering for Surfaces}
As noted previously, NeRFs only implicitly encode the structure of the scene under consideration.
While novel views can be extracted trivially from a trained model, surface extraction must be done using a separate procedure such as using marching cubes to construct a level set of the density field, using tesselation of a point cloud sampled from the scene, or inferring from rendered depth images.
To address this, a family of methods~\cite{wang_neus_2023,wang_neus2_2023} have been proposed to adapt the volume rendering procedure used in NeRFs to work directly on an SDF $f_{\lambda}: \mathbb{R}^3 \to \mathbb{R}$.
The SDF replaces the density field and allows for unambiguous extraction of surfaces from the model as the zero-level set of the field as opposed to an arbitrary density threshold used in a NeRF.
This is accomplished by defining a new density based on the SDF $\sigma(\mathbf{x}) = \phi(f_{\lambda}(\mathbf{x}))$ where $\phi_s(x) = s e^{-s x} \big/ \left( 1 + e^{-s x} \right)^2$ is the logistic function parameterized by scale $s$ and is the derivative of the sigmoid function $\Phi_s(x)$.
Importantly, the scale $s$ is a learnable parameter that is adjusted automatically during training.
We can then calculate the opacity of a sample point as
\begin{equation}
  \alpha_i = \max\left( \frac{\Phi_s\left(f(\mathbf{x}_i)\right) - \Phi_s\left(f(\mathbf{x}_{i+1})\right)}{\Phi_s\left(f(\mathbf{x}_i)\right)},\, 0 \right).
  \label{eq:sdf-opacity}
\end{equation}
Accumulated transmittance is then calculated as
\begin{equation}
  T_i = \prod_{j=1}^{i-1} (1 - \alpha_j)
\end{equation}
and a reformulated version of the color rendering equation can be used to calculate the predicted color
\begin{equation}
  \hat{c}_m = \sum_{i=1}^{N_s} T_i \alpha_i c_i.
\end{equation}
To enforce that $f_{\lambda}$ is a signed distance function, a regularizing loss term is added to enforce the eikonal property of this field~\cite{wang_neus_2023}
\begin{equation}
  \mathcal{L}_{\mathrm{eik}} = \frac{1}{N_s} \sum_{i=1}^{N_s} \left( \left\| \nabla f(\mathbf{x}_{i}) \right\|^2 - 1 \right)^2.
\end{equation}

\subsection{Neural Terrain Maps}
\begin{figure}[thbp]
  \centering
  \includegraphics[width=\linewidth]{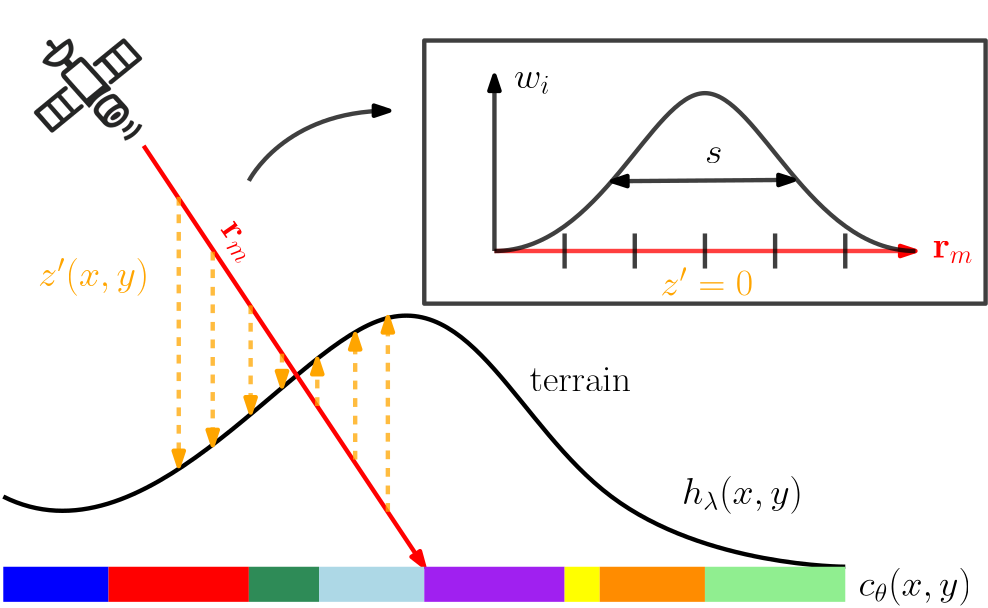}
  \caption{Overview of the neural terrain map (NTM) rendering procedure. The color field $c_{\theta}$ and height map $h_{\lambda}$ are queried to render terrain appearance and geometry from satellite imagery.}
  \label{fig:ntm-render-procedure}
\end{figure}

In this work, we modify neural volume rendering for surfaces for use on textured terrain maps directly.
We begin by redefining our base fields over the x-y plane of the domain as a color $c_{\theta}: \mathbb{R}^2 \to \mathbb{R}^d$ and a height field $h_{\lambda}: \mathbb{R}^2 \to \mathbb{R}$, where the height map is a scalar field that defines the height of the terrain at a point.
For every point $\mathbf{x}_i$ sampled along a locus during training, we consider the 2D point $\mathbf{p}_i = (x_i, y_i)$ in the x-y plane and the height of the sample $z_i$.
Colors and predicted heights are obtained by sampling the networks as $c_i = c_{\theta}(\mathbf{p}_i)$ and $h_i = h_{\lambda}(\mathbf{p}_i)$.
The altitude of the sample point above the terrain is then given by $z^\prime_i = z_i - h_i$ and is used in place of the signed distance $f$ in~(\ref{eq:sdf-opacity}) to calculate the opacity of the sample point.
A diagram of this procedure is shown in Fig.~\ref{fig:ntm-render-procedure}.
Additionally, the eikonal loss term is dropped as a height map is not constrained to be eikonal.
The scale parameter $s$ is initialized to a large value.
In the early stages of the training, this allows the network to learn a rough approximation of the terrain.
As training progresses, the network learns to decreases $s$ leading to a more refined surface estimate.

\section{IMPLEMENTATION}\label{sec:implementation}

\subsection{Network Architecture}
\begin{figure}[htbp]
  \centering
  {
    \tikzstyle{every picture}=[tikzfig]
    \resizebox{8cm}{!}{\begin{tikzpicture}
	\begin{pgfonlayer}{nodelayer}
		\node [style=mlpstyle] (0) at (0, 2) {Elev\\MLP\\1};
		\node [style=mlpstyle] (1) at (3, 2) {Elev\\MLP\\2};
		\node [style=mlpstyle] (2) at (0, -2) {Color\\MLP};
		\node [style=enc] (3) at (-3, 2) {Hash\\Enc};
		\node [style=emb] (4) at (-3, -2) {Ap\\Emb};
		\node [style=none] (5) at (-2.25, 2) {};
		\node [style=none] (6) at (-1, 2) {};
		\node [style=none] (7) at (-1.5, 2.25) {};
		\node [style=none] (8) at (-1.5, 3.5) {};
		\node [style=none] (9) at (1.5, 3.5) {};
		\node [style=none] (10) at (1.5, 2.75) {};
		\node [style=none] (11) at (2, 2.75) {};
		\node [style=none] (12) at (1, 2) {};
		\node [style=none] (13) at (2, 2) {};
		\node [style=none] (14) at (4, 2.5) {};
		\node [style=none] (15) at (4.75, 2.5) {};
		\node [style=none] (16) at (4, 1.5) {};
		\node [style=none] (17) at (4.75, 1.5) {};
		\node [style=none] (18) at (4.75, 0) {};
		\node [style=none] (19) at (-1.5, 0) {};
		\node [style=none] (20) at (-1.5, -1.25) {};
		\node [style=none] (21) at (-1, -1.25) {};
		\node [style=none] (22) at (-2.25, -2.75) {};
		\node [style=none] (23) at (-1, -2.75) {};
		\node [style=none] (24) at (-5.25, 2) {$\mathbf{x}_{m, i}$};
		\node [style=none] (26) at (-4.25, 1.75) {};
		\node [style=none] (27) at (-4.25, 0) {};
		\node [style=none] (28) at (-2, 0) {};
		\node [style=none] (31) at (-2, -2) {};
		\node [style=none] (32) at (-1, -2) {};
		\node [style=none] (33) at (-5, -2) {$I_n$};
		\node [style=none] (34) at (-4.5, 2) {};
		\node [style=none] (35) at (-3.75, 2) {};
		\node [style=none] (36) at (-4.5, -2) {};
		\node [style=none] (37) at (-3.75, -2) {};
		\node [style=none] (38) at (5.25, 2.5) {};
		\node [style=none] (39) at (5.25, 2.5) {$z_i$};
		\node [style=none] (40) at (1, -2) {};
		\node [style=none] (41) at (2.25, -2) {$c_i$};
		\node [style=none] (42) at (1.75, -2) {};
	\end{pgfonlayer}
	\begin{pgfonlayer}{edgelayer}
		\draw [style={nn_connector}] (26.center) to (27.center);
		\draw [style={nn_connector}] (27.center) to (28.center);
		\draw [style={nn_arrow}] (31.center) to (32.center);
		\draw [style={nn_arrow}] (20.center) to (21.center);
		\draw [style={nn_arrow}] (22.center) to (23.center);
		\draw [style={nn_connector}] (19.center) to (20.center);
		\draw [style={nn_connector}] (19.center) to (18.center);
		\draw [style={nn_connector}] (16.center) to (17.center);
		\draw [style={nn_connector}] (17.center) to (18.center);
		\draw [style={nn_arrow}] (14.center) to (15.center);
		\draw [style={nn_arrow}] (12.center) to (13.center);
		\draw [style={nn_arrow}] (5.center) to (6.center);
		\draw [style={nn_connector}] (7.center) to (8.center);
		\draw [style={nn_connector}] (8.center) to (9.center);
		\draw [style={nn_connector}] (9.center) to (10.center);
		\draw [style={nn_arrow}] (10.center) to (11.center);
		\draw [style={nn_connector}] (28.center) to (31.center);
		\draw [style={nn_arrow}] (34.center) to (35.center);
		\draw [style={nn_arrow}] (36.center) to (37.center);
		\draw [style={nn_arrow}] (40.center) to (42.center);
	\end{pgfonlayer}
\end{tikzpicture}}
  }
  \caption{NTM network architecture.
  Inputs to each component are concatenated together.}
  \label{fig:net-arch}
\end{figure}
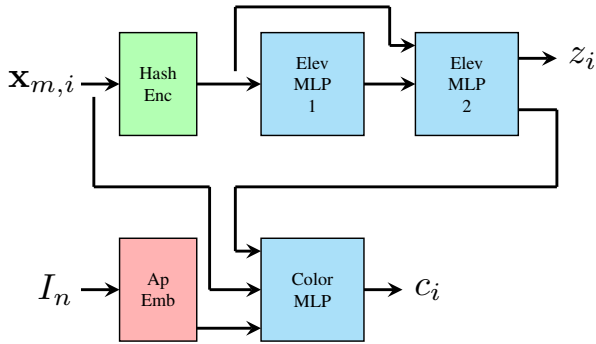

Our network architecture, visualized in Fig.~\ref{fig:net-arch}, is based on that used in the original NeRF.
Both the height field $h_{\lambda}$ and color field $c_{\theta}$ are implemented as fully connected networks with softplus activation functions between layers.
For our experiments, we found that 8 layers for the height network and 4 for the color network with 128 hidden units per layer worked well.
The height network has a skip connection between the fourth and fifth layers to help mitigate the depth of the network and combat vanishing gradients.
The output activation for the color network is a sigmoid function, and the height network has no output activation as we found experimentally that restricting the output in this way led to significantly degraded results.
Additionally, as is common practice with NeRFs, we use a positional encoding on the input samples to combat the tendency of neural networks to bias learning toward low frequency data.
We use a multi-resolution hash encoding~\cite{muller_instant_2022} with 16 levels, a base resolution of 8 and a maximum resolution of 5000.
The $\log_2$ hash grid size is set to 19, and the number of features per level is set to 2.
\par
Each image locus is sampled according to the proposal sampler architecture used in Nerfacto~\cite{tancik_nerfstudio_2023}.
Input coordinates are normalized to the range $[0, 1]$ before being passed to the network.
This is done using a precomputed bounding box which is estimated for each scene using rough image footprints.
The bounding box is referenced to a local East-North-Up coordinate frame with the origin at the estimated mean center of the scene, and height bounds are set according to the known maximum and minimum terrain heights on the planetary body in question.
Experimentally, we found that normalizing the bounding box and camera positions by the maximum bounding box values and then scaling up by a factor of 10 worked to mitigate numerical issues during training which caused divergence.

\subsection{Datasets and Processing}
In this work, we make use of of satellite imagery from two active missions: the Mars Reconnaissance Orbiter (MRO) Context Camera (CTX)~\cite{bell_iii_calibration_2013} and the Advanced Spaceborne Thermal Emission and Reflection Radiometer (ASTER)~\cite{nichols_aster_1994}.
We also use synthetic data generated from Google Earth Studio (GES)~\footnote{https://www.google.com/earth/studio/} as a baseline dataset with perfect intrinsics and extrinsics and no variation in illumination.
In this section, we describe each of these datasets, the related instrument that produced them, and how they were processed for use in NTM.
For a complete list of images used in this work, including unique identifiers for each one, see Table~\ref{tab:image-ids}.

\subsubsection{Mars Reconnaissance Orbiter Context Camera}
CTX is a linescan camera carried by the MRO spacecraft.
From the MRO nominal orbital altitude of \qty{300}{\kilo\meter}, CTX images have an approximate ground sample distance (GSD) of \qty{6}{\meter} per pixel over a swath of approximately \qty{30}{\kilo\meter} in width.
The image detector is \qty{5000}{pixels} wide and the images used in this work are between \qty{5000}{pixels} and \qty{10000}{pixels} in height (some CTX images have significantly larger row counts, but counts over 10000 were excluded in order to maintain consistent overlap of the scene and to conserve GPU memory).
The CTX instrument has a high focal length corresponding to a field of view of \qty{5.7}{\degree}.
It images in the red band with an effective wavelength of \qty{611}{\nano\meter} and a passband full width half maximum of \qty{189}{\nano\meter}.
The MRO orbit is sun-synchronous, meaning that the camera always images the surface at the same local solar time.
This guarantees some level of illumination consistency between images which makes it easier for MVS pipelines to find feature correspondences between images and for NTM to find a single surface which produces a visually consistent rendering across all input views.
However, the sun-synchronous orbit also guarantees that the same regions are in shadow in each image, which makes it hard to reconstruct detail in these areas.

\begin{figure}[h]
  \centering
  \includegraphics[width=\linewidth]{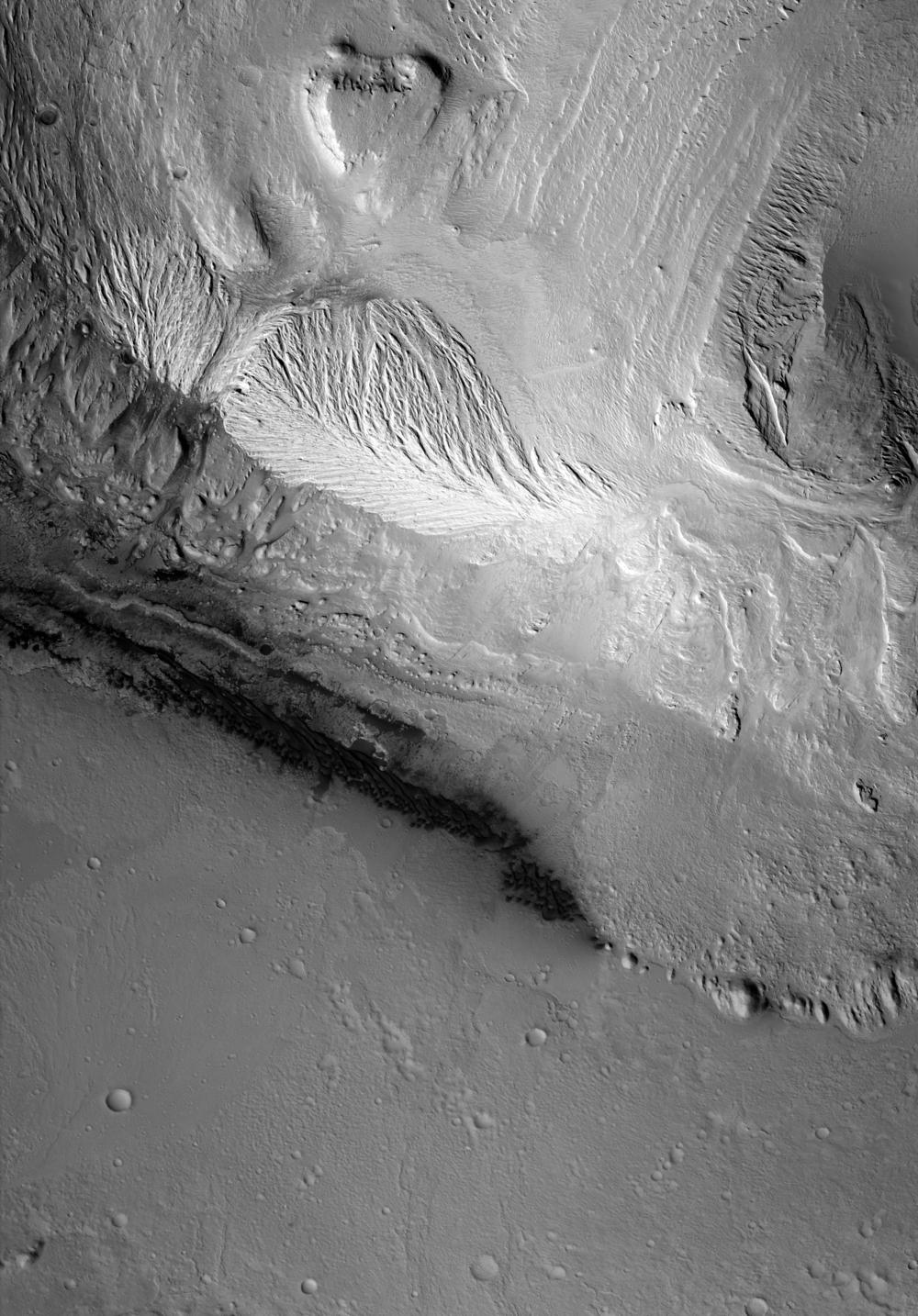}
  \caption{Example CTX image from Gale Crater.}
\end{figure}

\subsubsection{Advanced Spaceborne Thermal Emission and Reflection Radiometer}
ASTER is a multi-spectral linescan camera carried by the Terra satellite.
The data we use in this work is from the third band of the visible and near-infrared (VNIR) subsystem which produces linescan images in the \qty{760}{\nano\meter} to \qty{860}{\nano\meter} band.
The GSD of the VNIR instrument is approximately \qty{15}{\meter} per pixel and it images a swath of \qty{60}{\kilo\meter}.
The orbital altitude of the Terra satellite is approximately \qty{705}{\kilo\meter} and is also sun-synchronous.
The VNIR detector is \qty{5000}{pixels} wide and images are approximately \qty{4000}{pixels} tall.
The third VNIR band is unique in that each area it surveys is imaged twice in a single pass, once with a nadir pointed camera and then again with a backwards pointing camera.
This guarantees that there is sufficient disparity between the two images to allow for stereo reconstruction.

\begin{figure}[h]
  \centering
  \includegraphics[width=\linewidth]{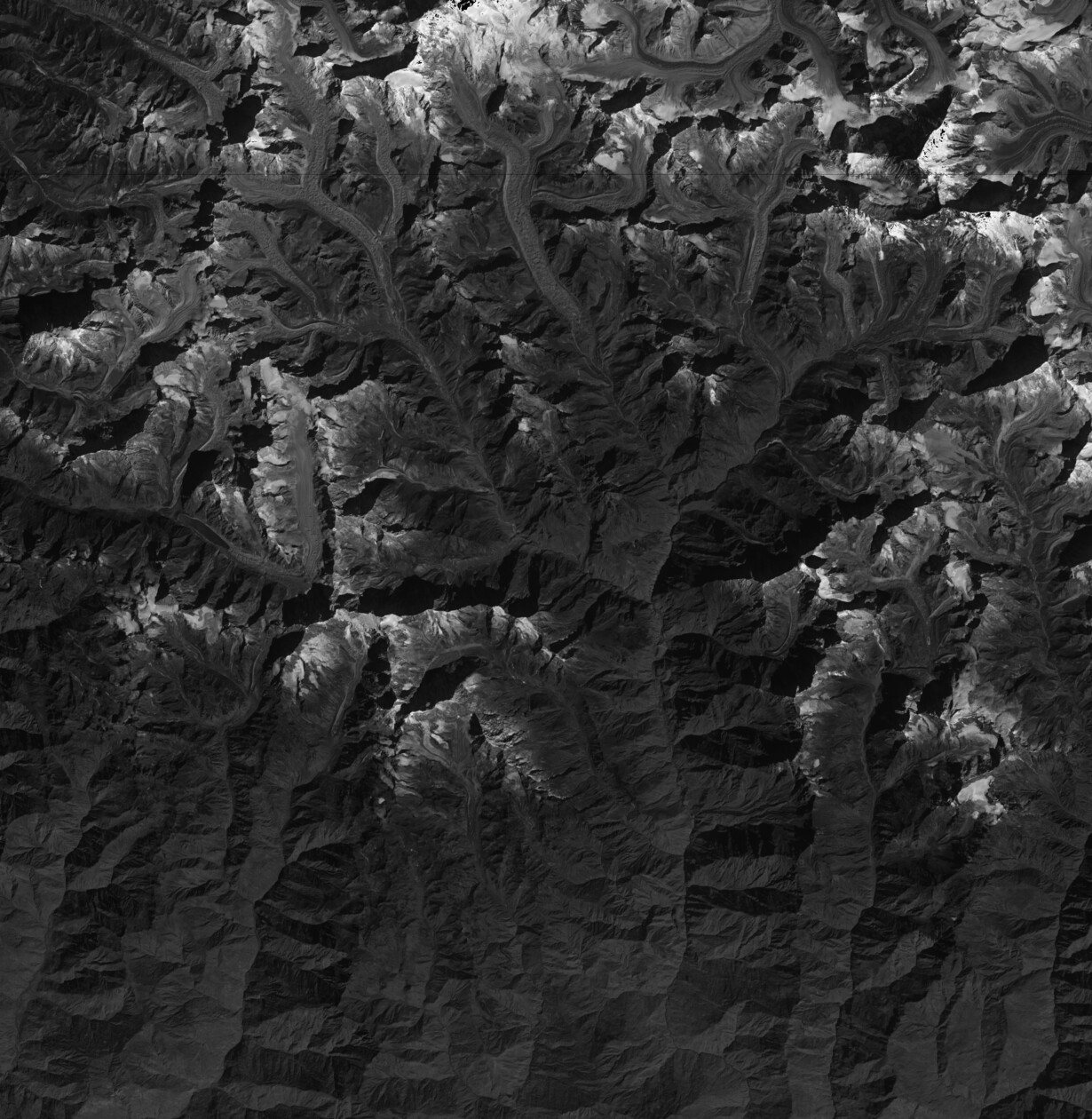}
  \caption{Example ASTER image from Everest.}
\end{figure}

\subsubsection{Google Earth Studio}
GES is a web-based tool for exporting renderings of Google Earth imagery.
We use it to simulate a perfect satellite imaging scenario with no lighting variation and perfect camera intrinsics and extrinsics.
GES supports rendering imagery using a pinhole camera model.
In order to be commensurate with the CTX imagery, we choose a \qty{5}{\degree} field of view and simulate orbital passes West to East over our target at an altitude of \qty{250}{\kilo\meter} for Shasta and Denali and \qty{300}{\kilo\meter} for Everest.
The images are distributed over an approximately \qty{175}{\kilo\meter} ground track running West to East and centered on the subject, again commensurate with the CTX imagery.
Denali and Everest images are at a \qty{400}{pixel} by \qty{400}{pixel} resolution and Shasta images are at a \qty{480}{pixel} by \qty{480}{pixel} resolution which gives an approximately \qty{60}{\meter} GSD for all three scenes.
Also unlike the other two datasets, GES images are RGB.

\begin{figure}[h]
  \centering
  \includegraphics[width=\linewidth]{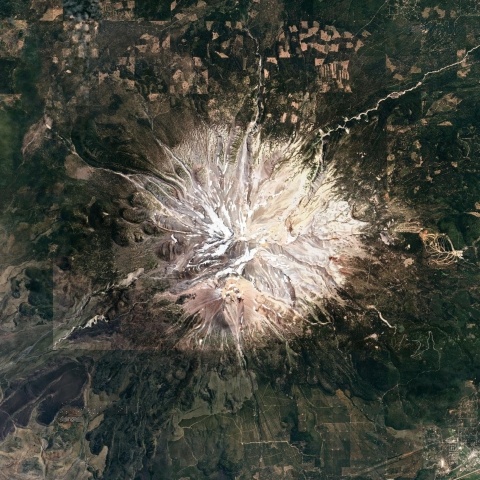}
  \caption{Example rendered image from GES of Mount Shasta.}
\end{figure}

\begin{figure*}[htbp]
  \centering

  {
    \def\ctxcolor{\cellcolor{blue!15}}
    \def\astercolor{\cellcolor{green!15}}
    \def\gescolor{\cellcolor{red!15}}
    \def\hirisecolor{\cellcolor{orange!15}}

    \def\ctxlabelcolor{blue!100}
    \def\asterlabelcolor{green!100}
    \def\geslabelcolor{red!100}
    \def\hiriselabelcolor{orange!100}
    \renewcommand{\imgbox}[1]{\includegraphics[width=\linewidth,keepaspectratio]{#1}}
    \renewcommand{\arraystretch}{1.5}

    \newcommand{\rowlabel}[2]{\rotatebox[origin=c]{90}{\textcolor{#1}{\large\textbf{#2}}}}
    \newcommand{\vertlabel}[1]{\rotatebox[origin=c]{90}{{\textbf{#1}}}}
    \begin{tabular}{lc>{\centering\arraybackslash}m{3.3cm}>{\centering\arraybackslash}m{3.3cm}>{\centering\arraybackslash}m{3.3cm}>{\centering\arraybackslash}m{3.3cm}r}
      & & \textbf{Gale Crater} & \textbf{Jezero Crater} & \textbf{Everest} & \textbf{Gunnbjørn Fjeld} & \\

      &
      \vertlabel{DTM} &
      \ctxcolor\imgbox{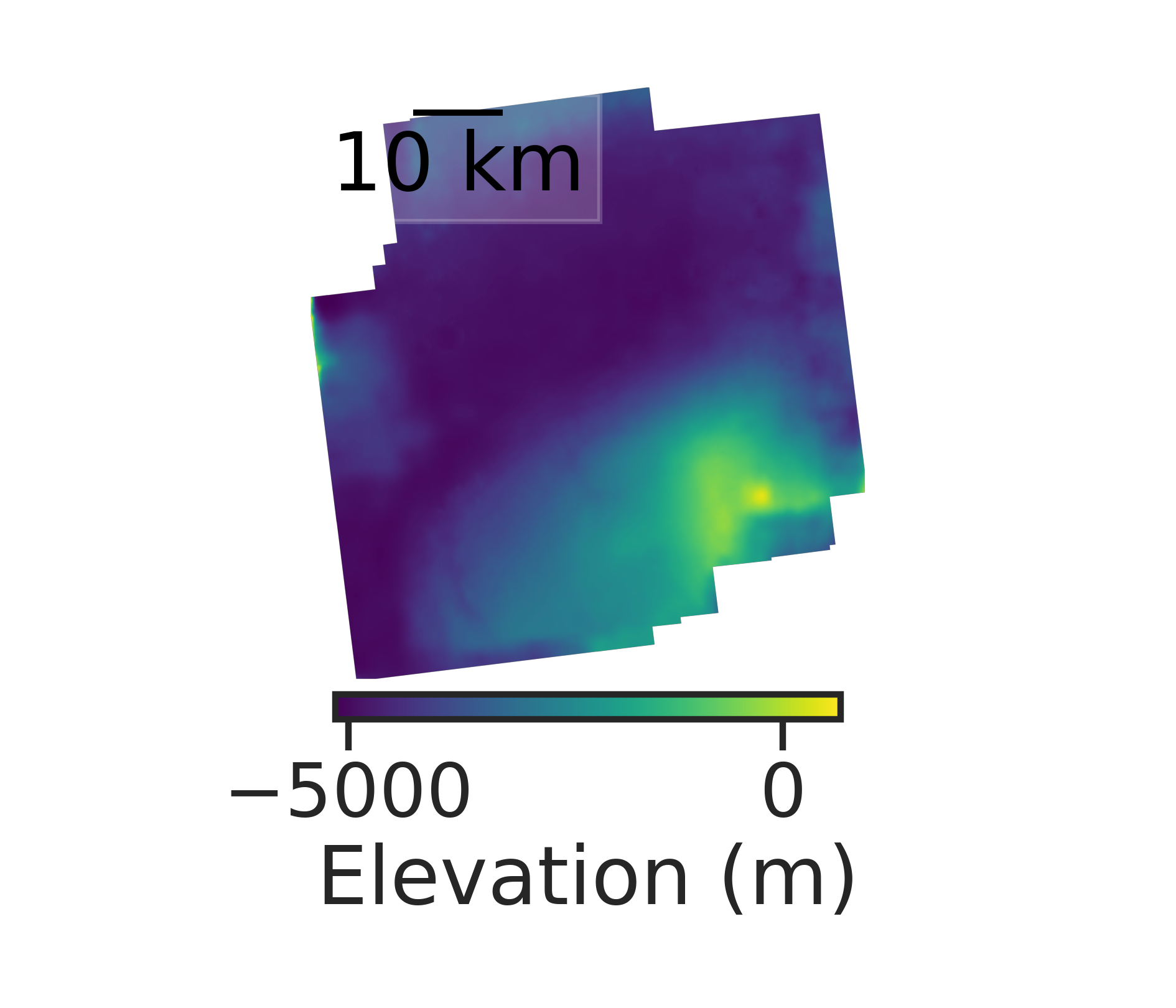} &
      \ctxcolor\imgbox{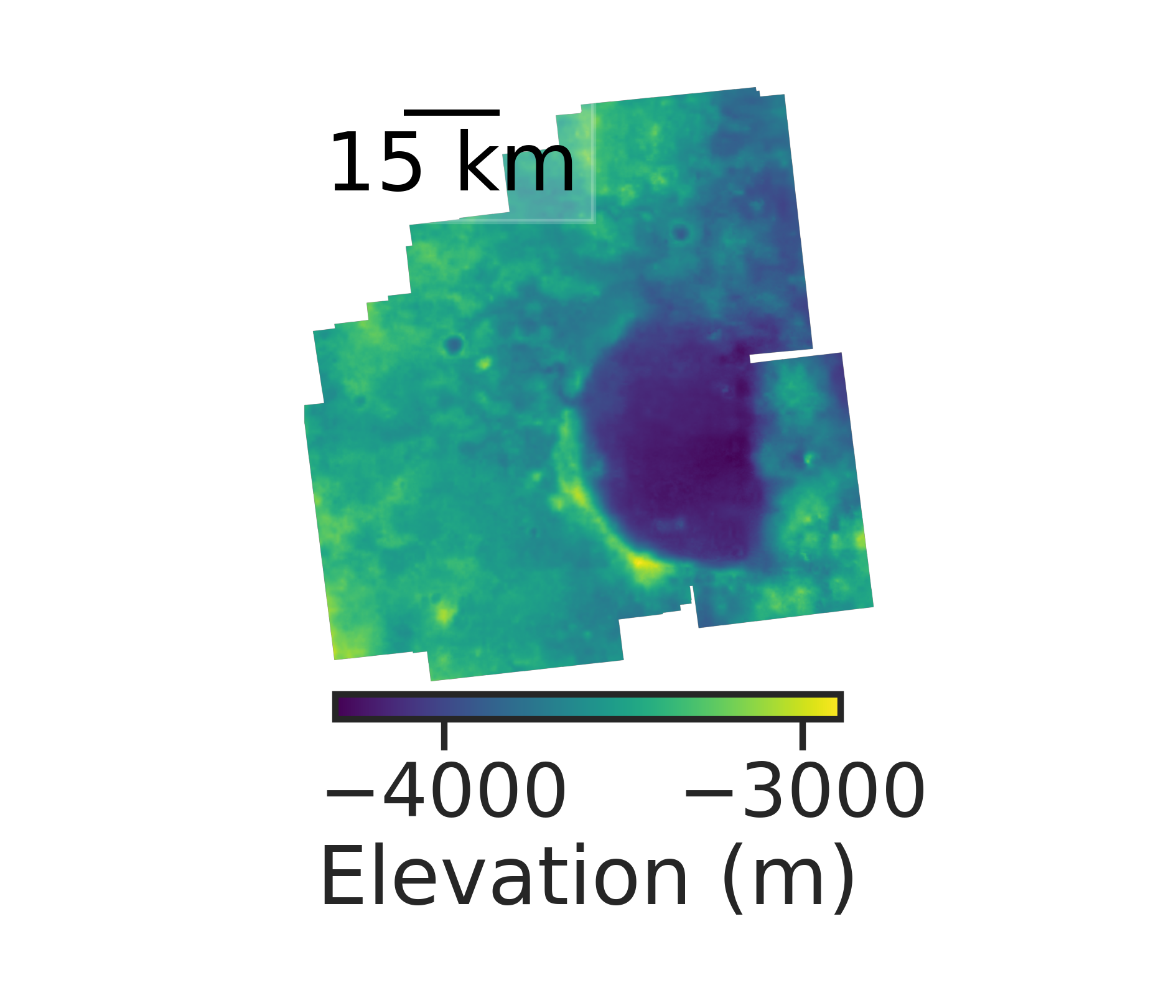} &
      \astercolor\imgbox{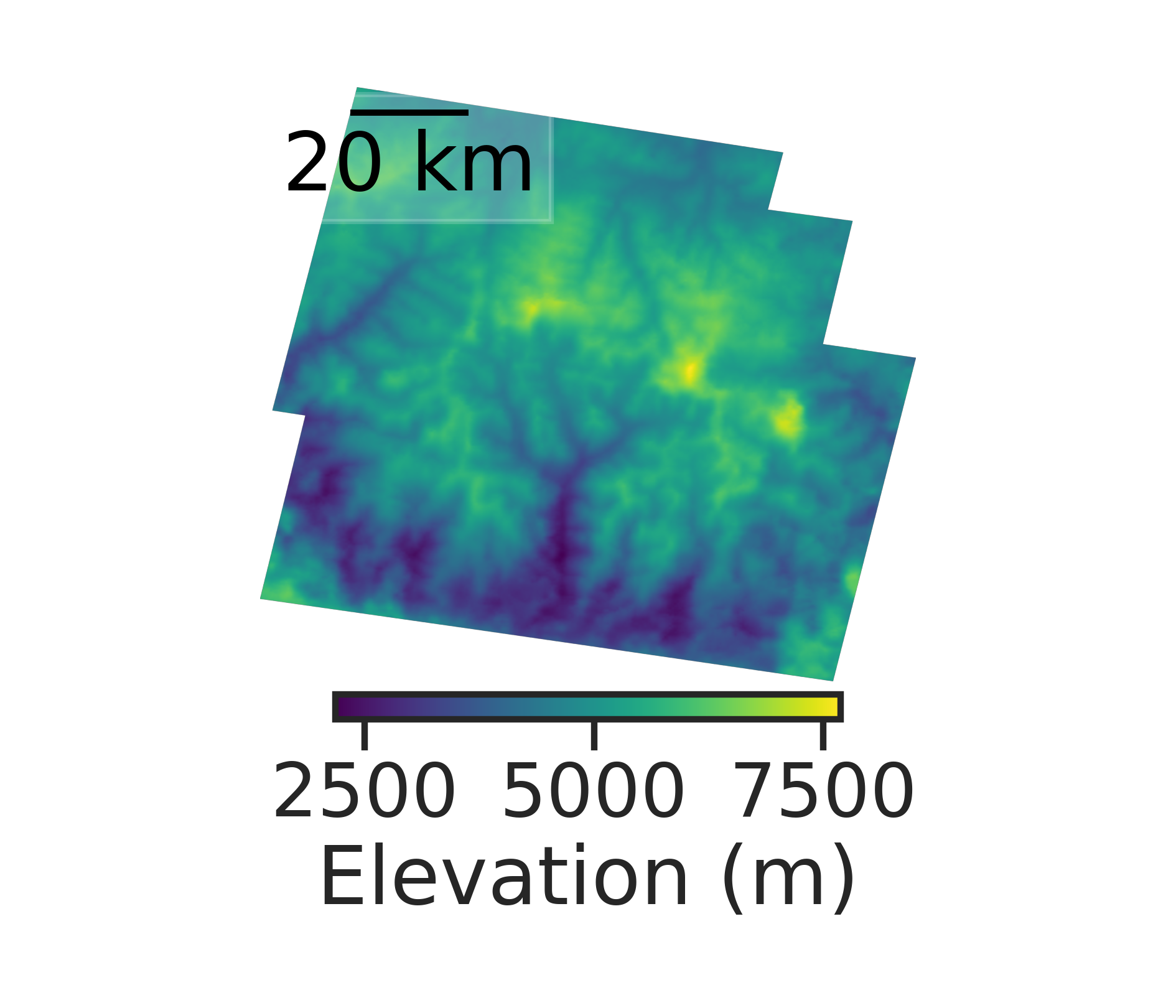} &
      \astercolor\imgbox{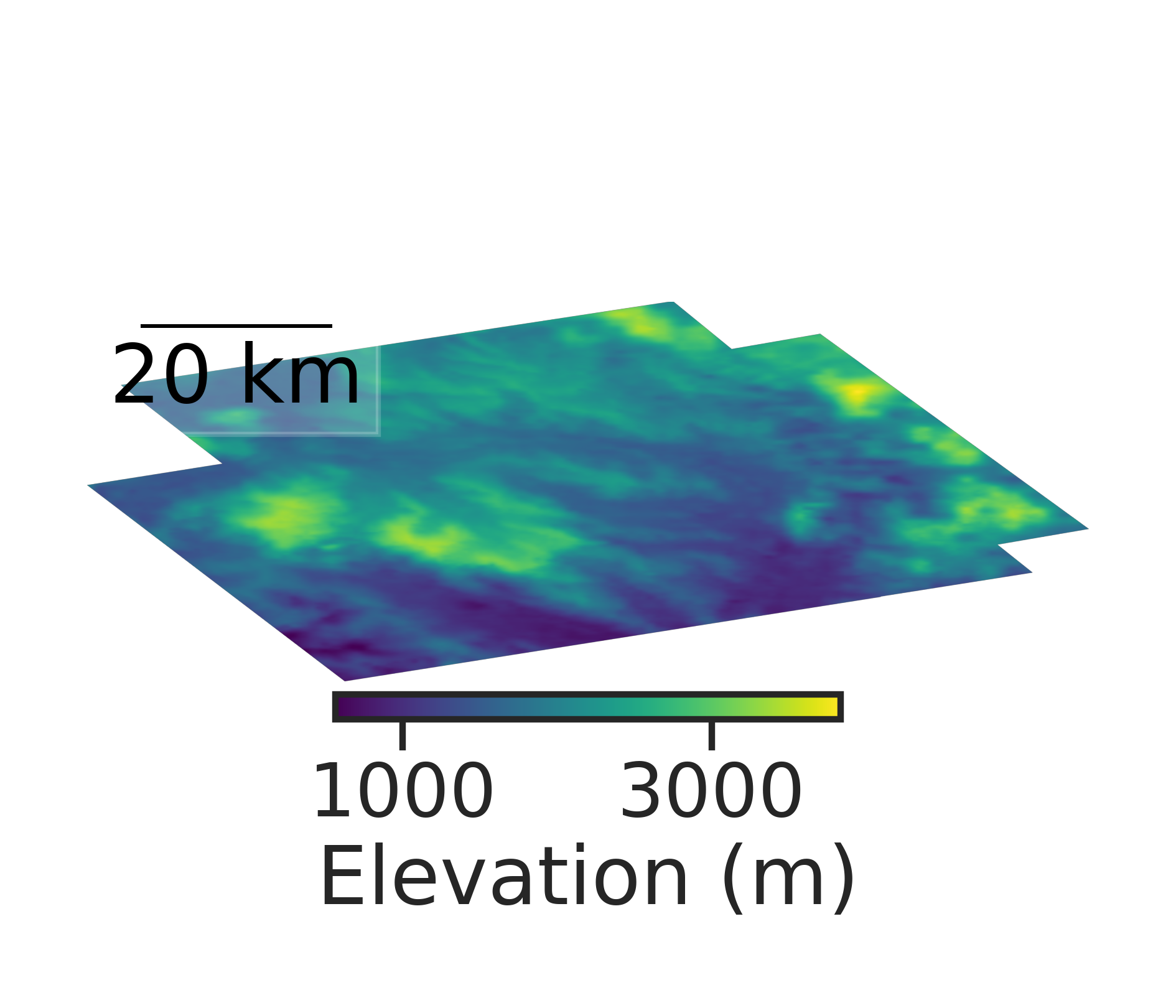} & \\

      \rowlabel{\ctxlabelcolor}{CTX} &
      \vertlabel{Error Std. Dev.} &
      \ctxcolor\imgbox{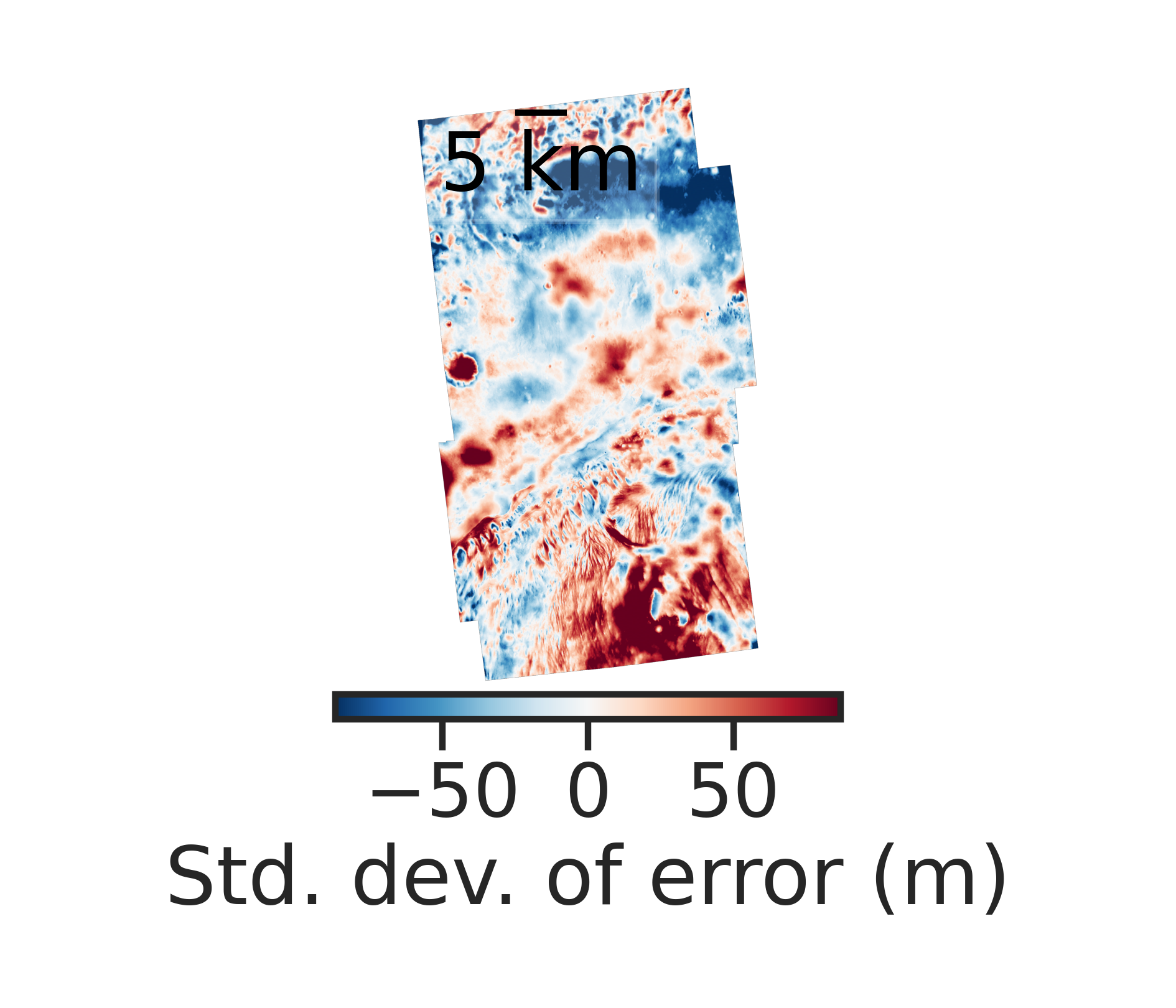} &
      \ctxcolor\imgbox{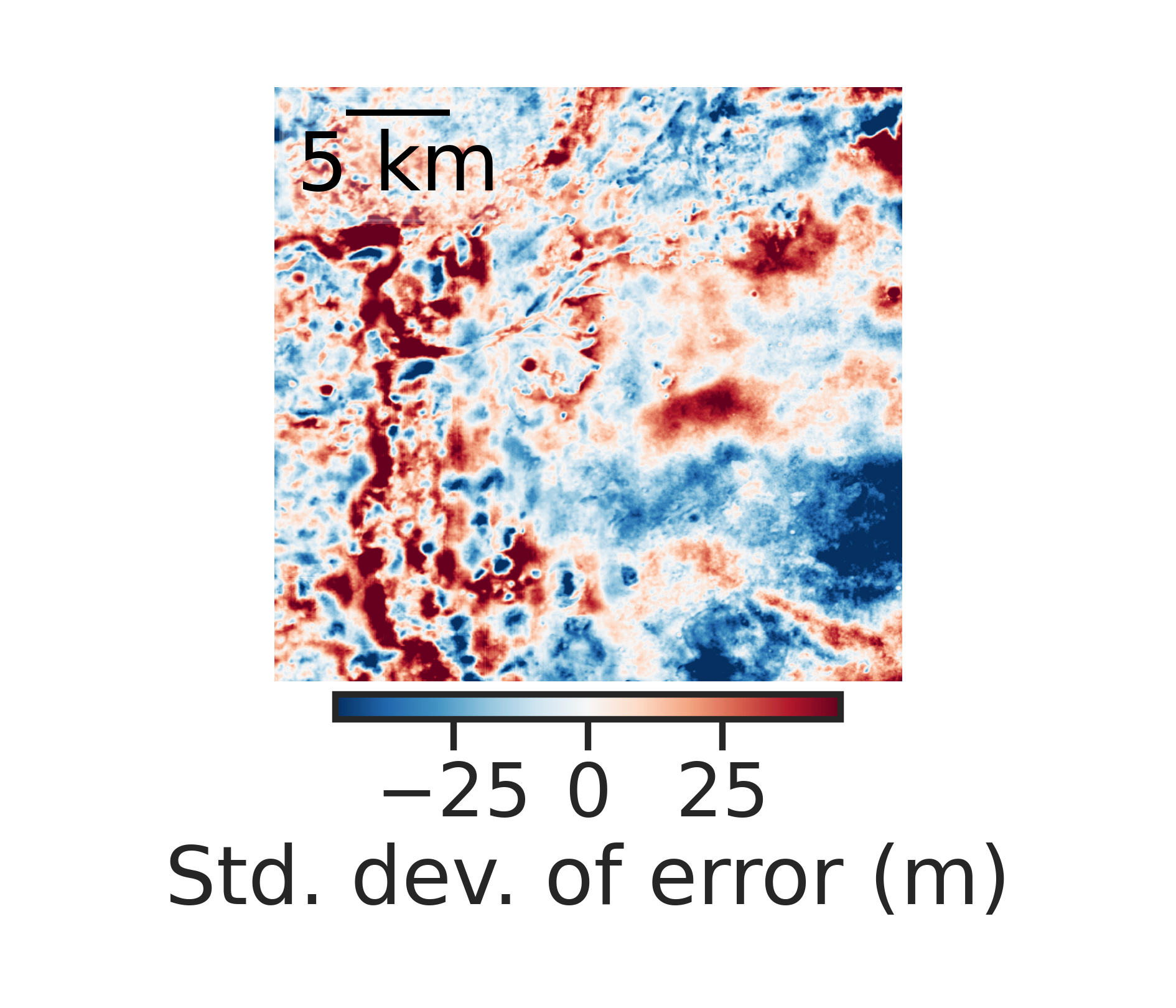} &
      \astercolor\imgbox{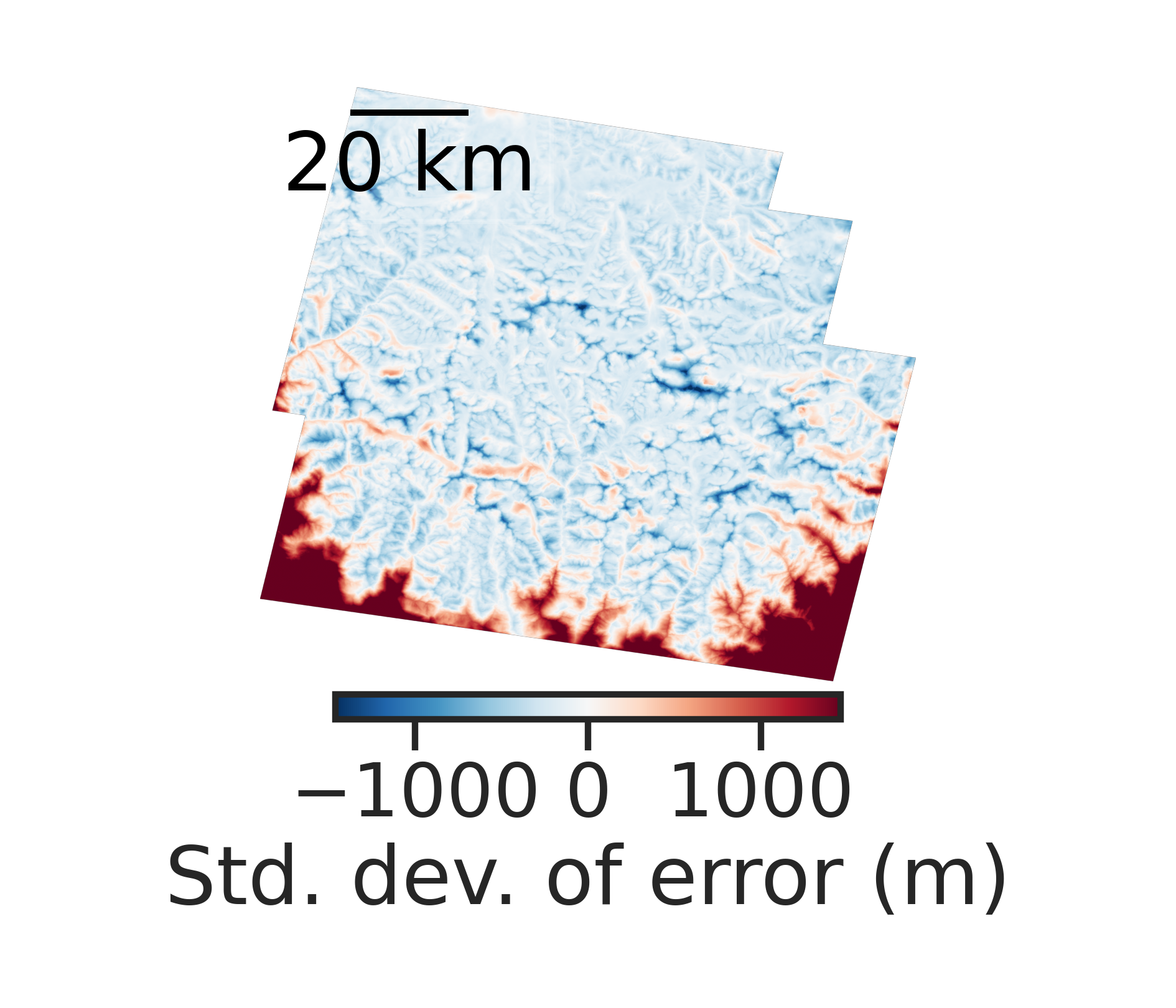} &
      \astercolor\imgbox{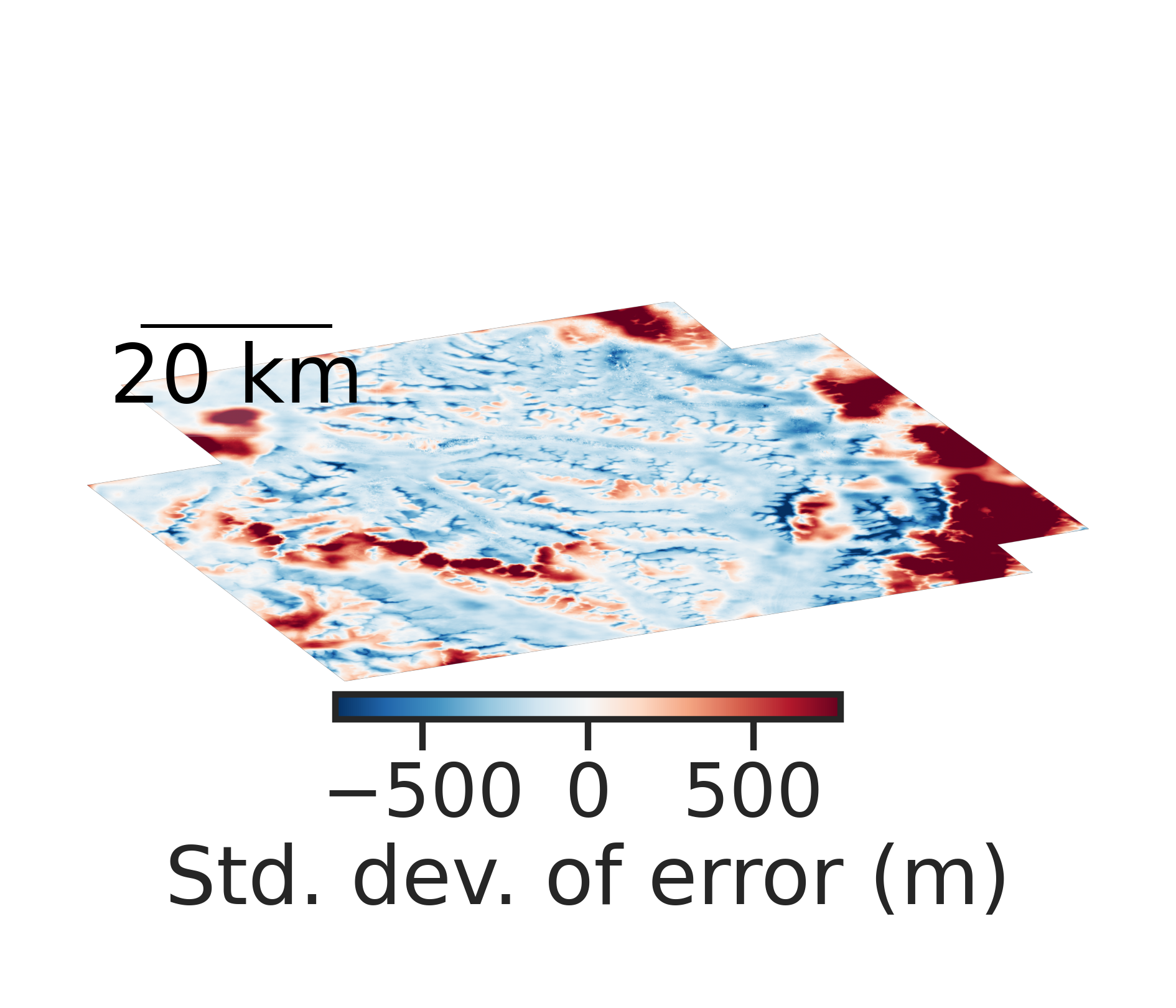} &
      \rowlabel{\asterlabelcolor}{ASTER} \\

      &
      \vertlabel{Error Hist.} &
      \ctxcolor\imgbox{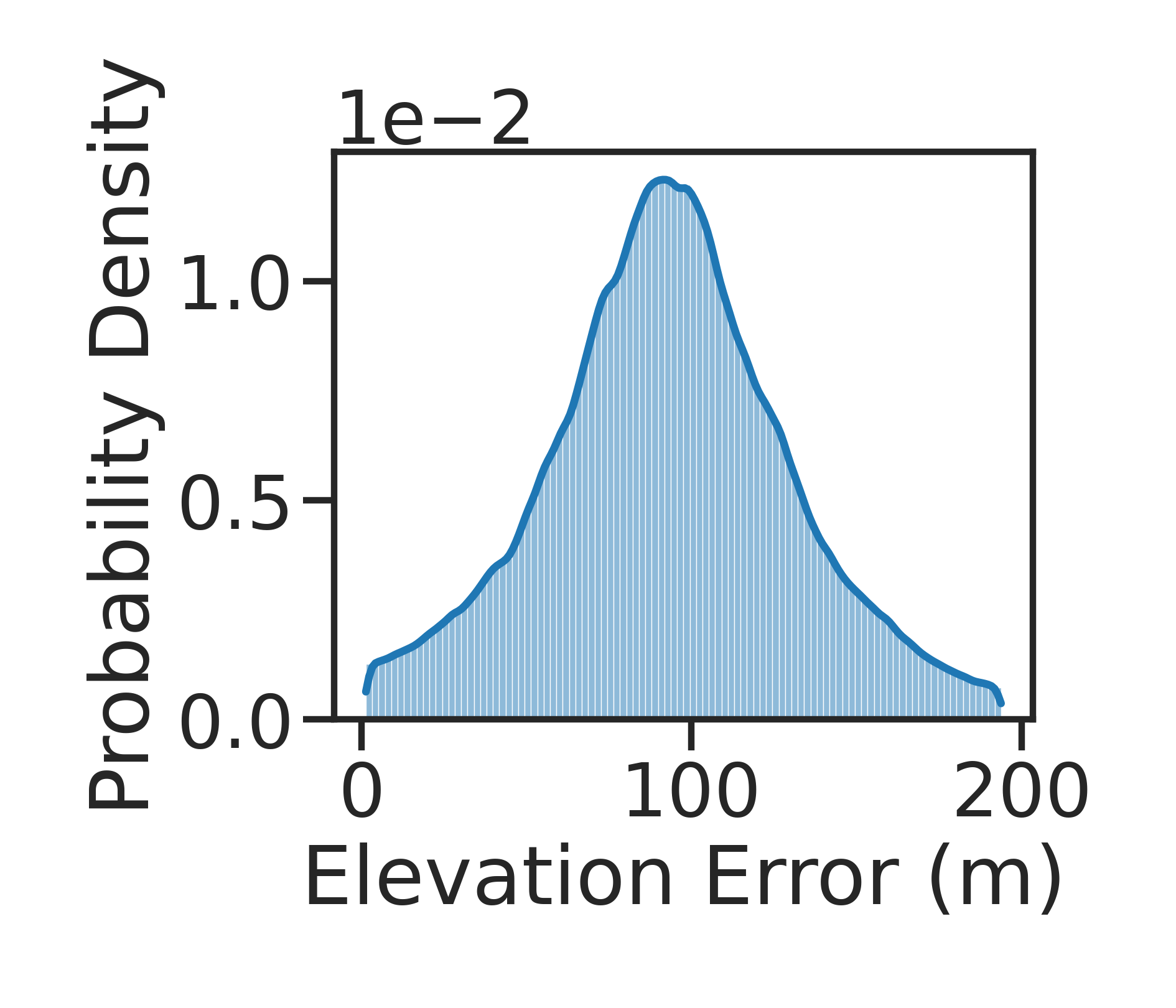} &
      \ctxcolor\imgbox{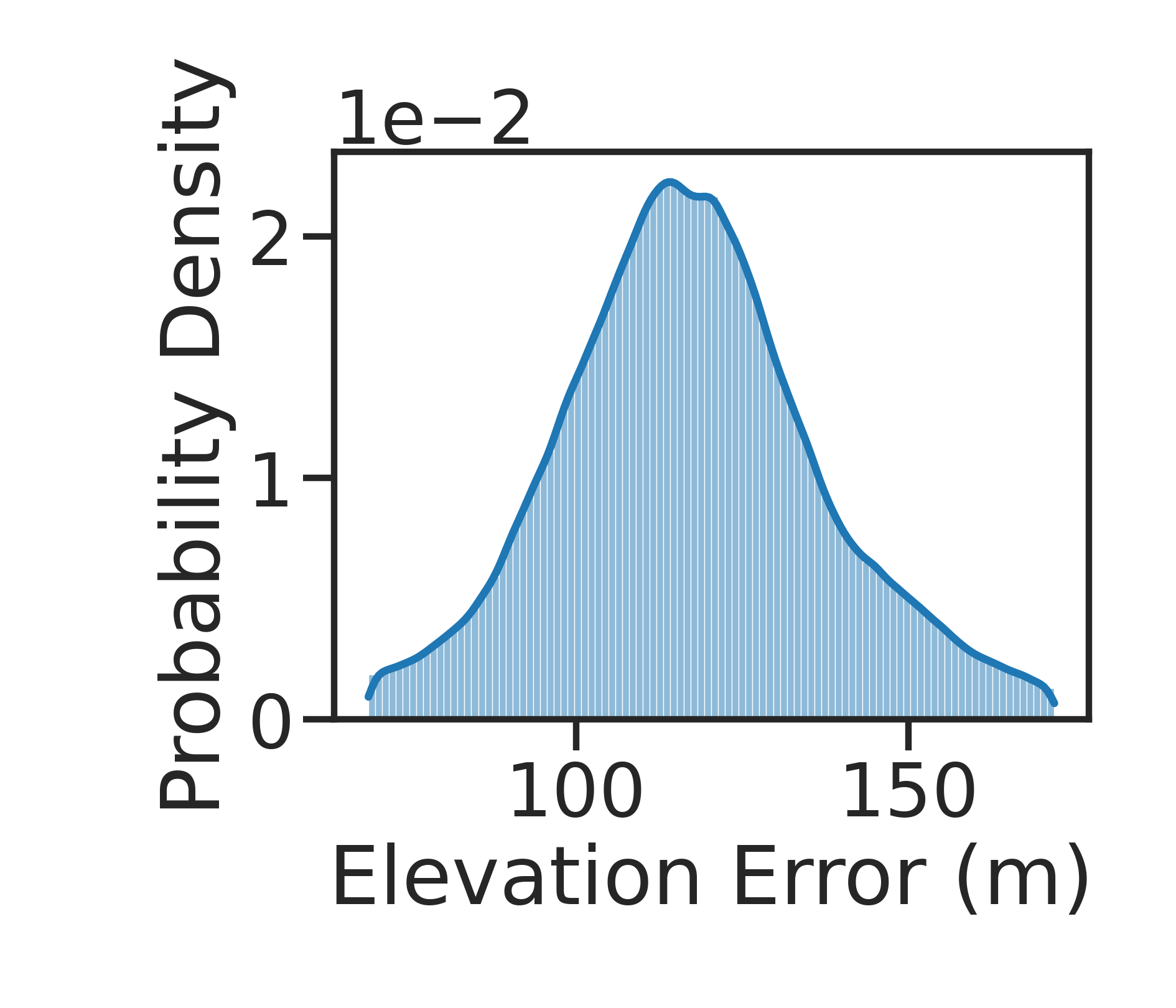} &
      \astercolor\imgbox{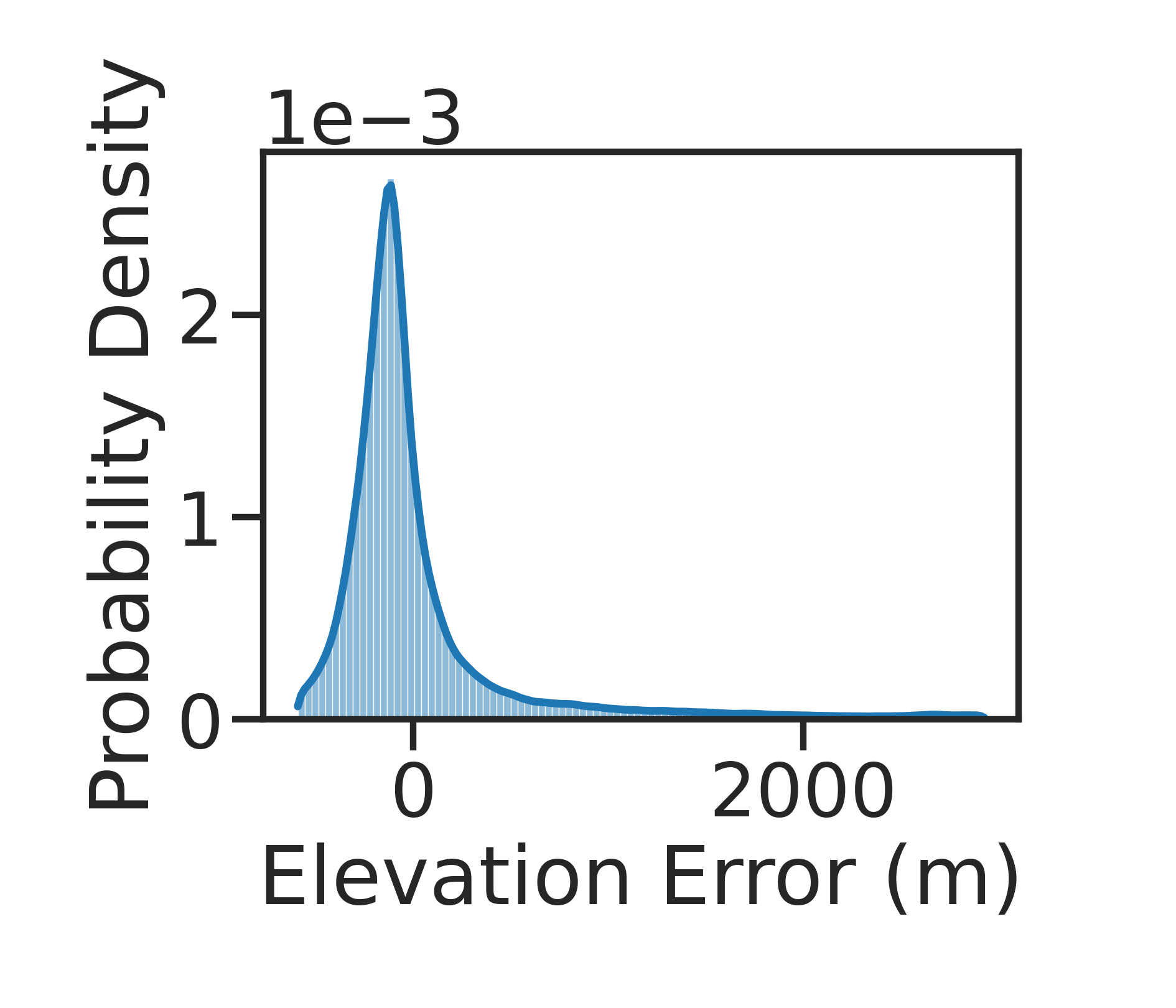} &
      \astercolor\imgbox{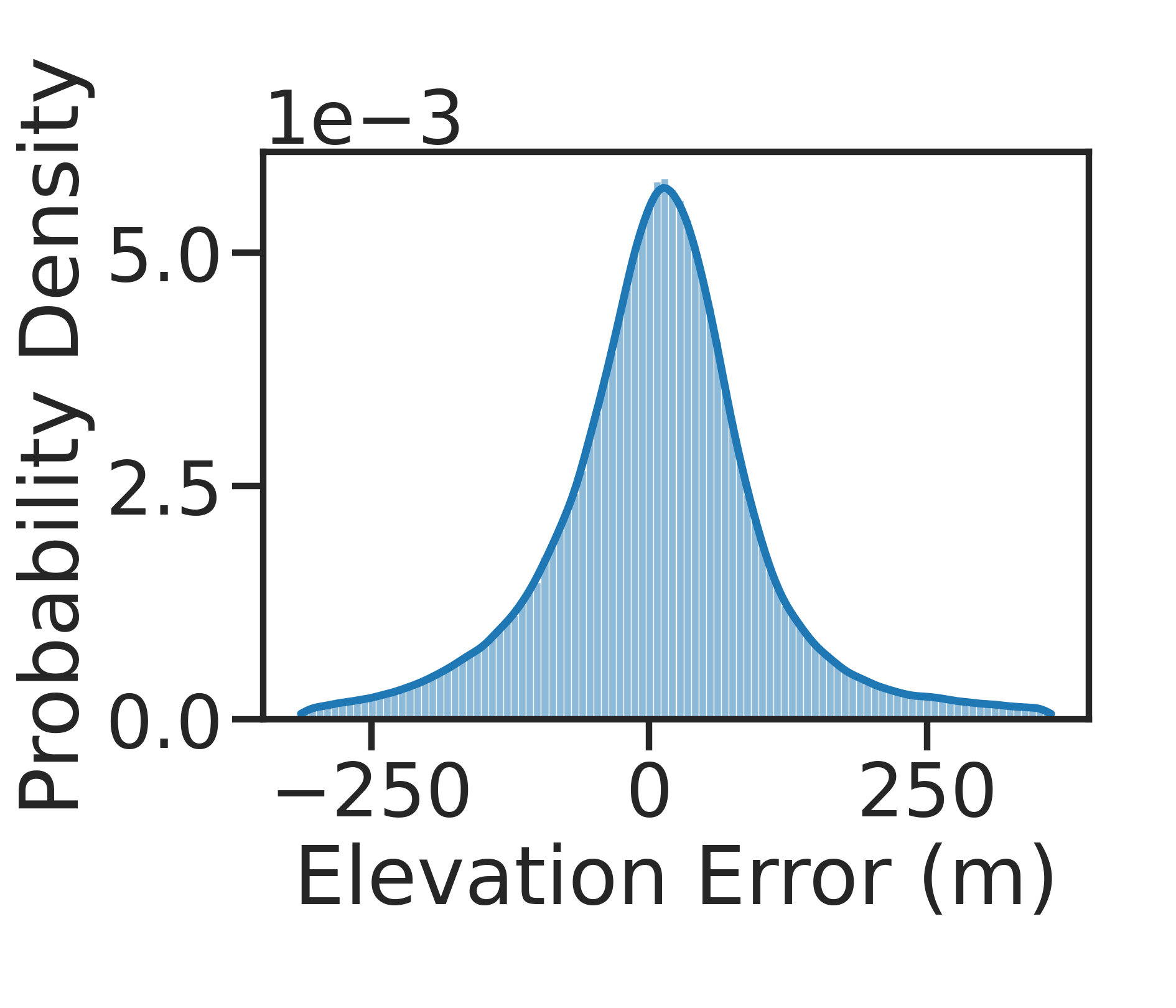} & \\

      & & \textbf{Denali} & \textbf{Shasta} & \textbf{Everest} & & \\

      &
      \vertlabel{DTM} &
      \gescolor\imgbox{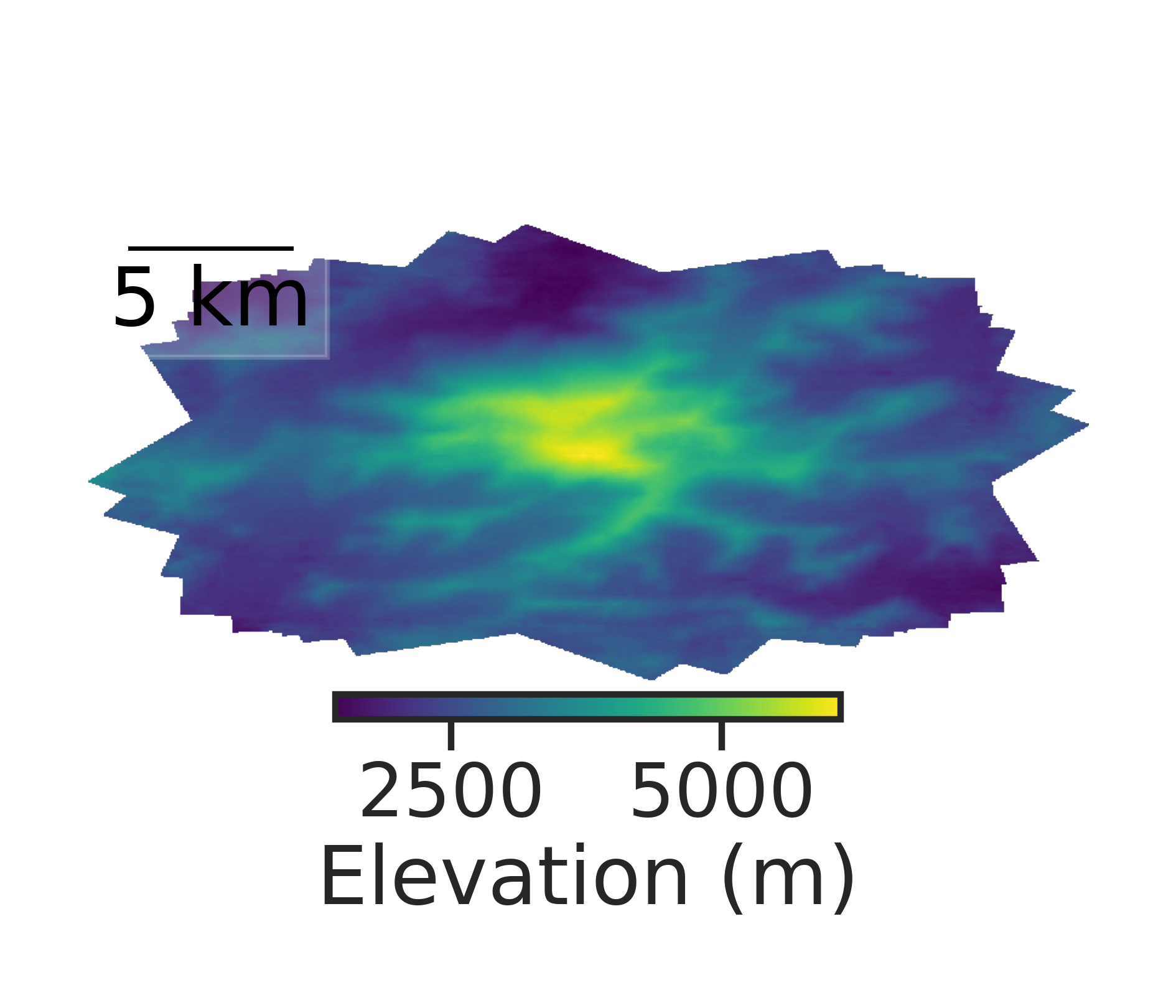} &
      \gescolor\imgbox{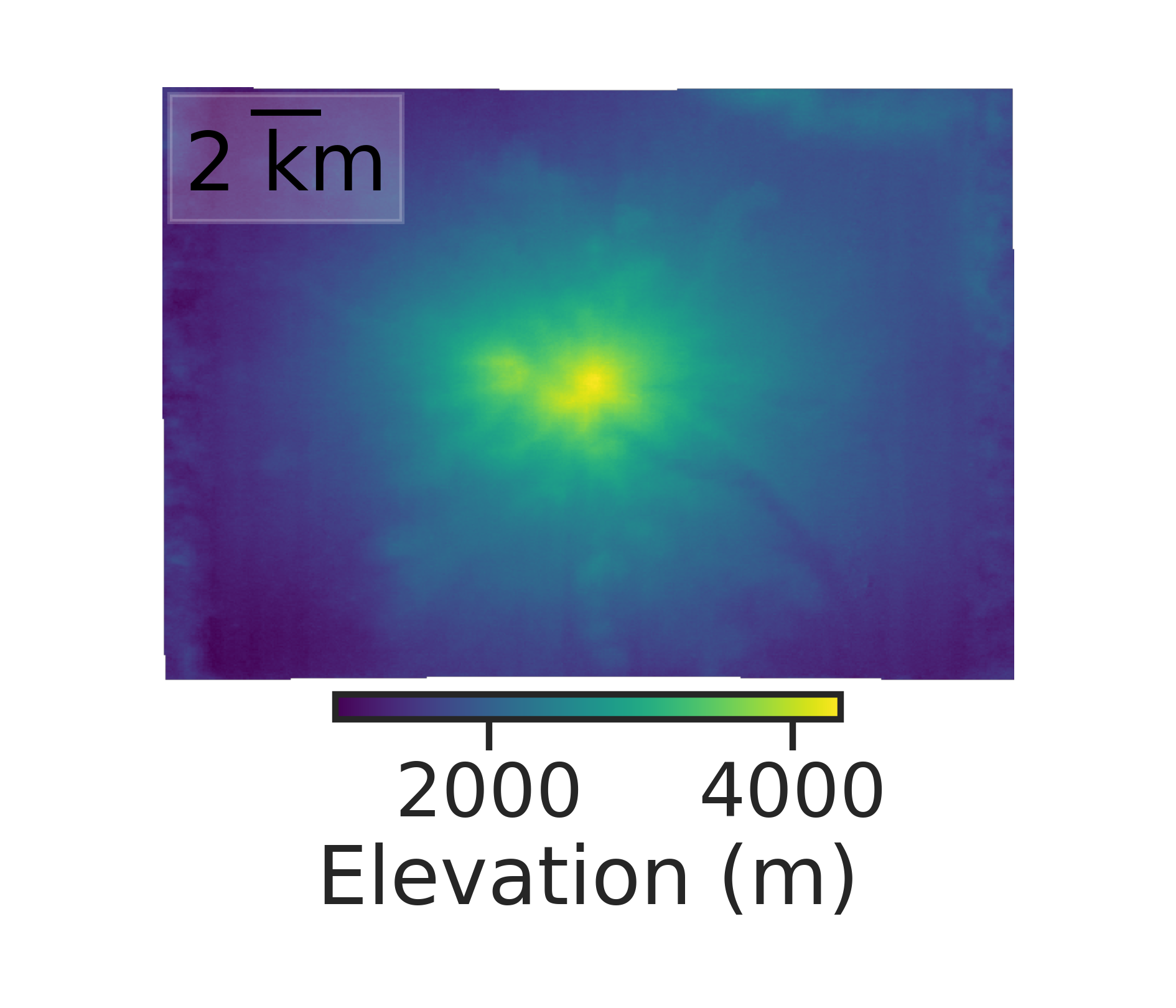} &
      \gescolor\imgbox{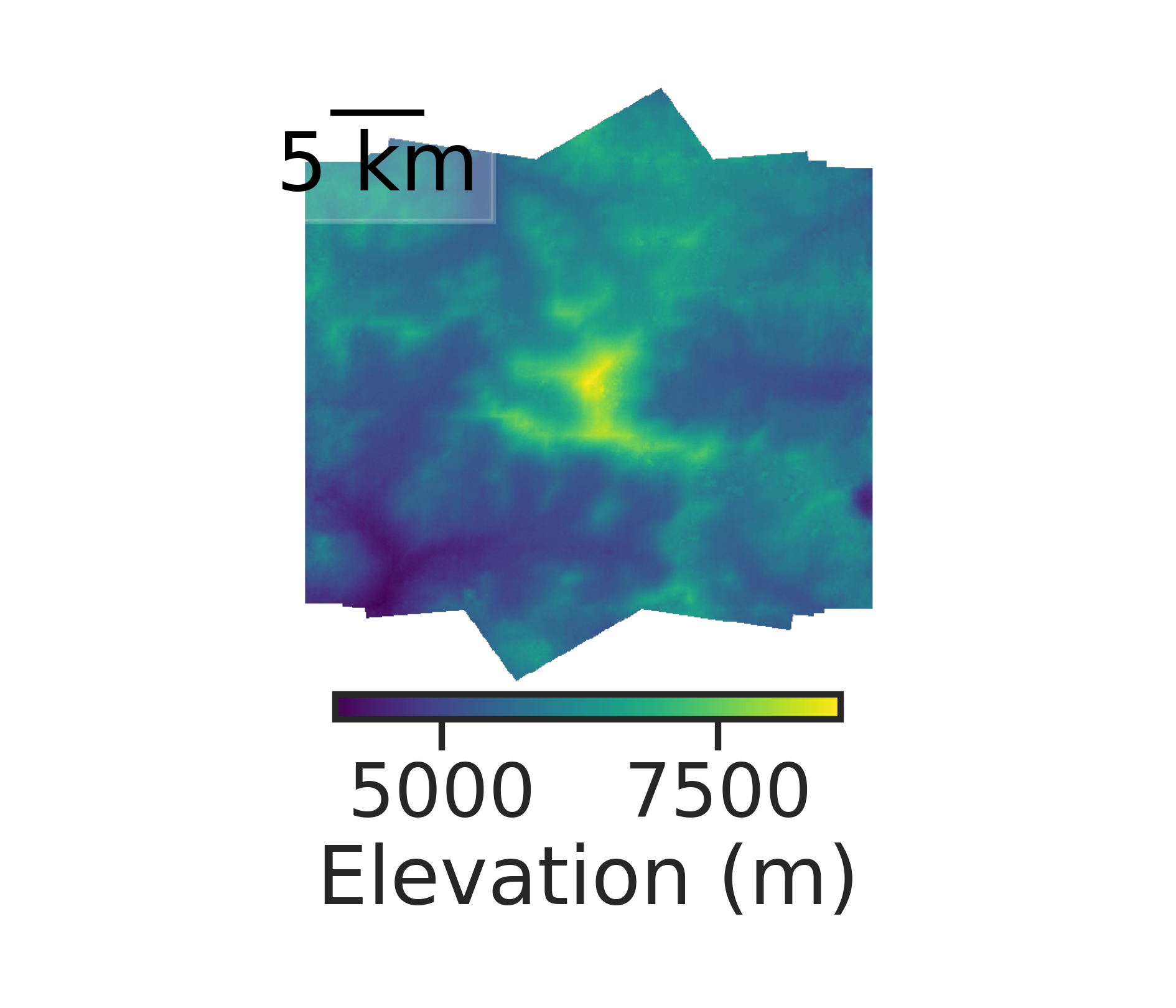} & & \\

      \rowlabel{\geslabelcolor}{GES} &
      \vertlabel{Error Std. Dev.} &
      \gescolor\imgbox{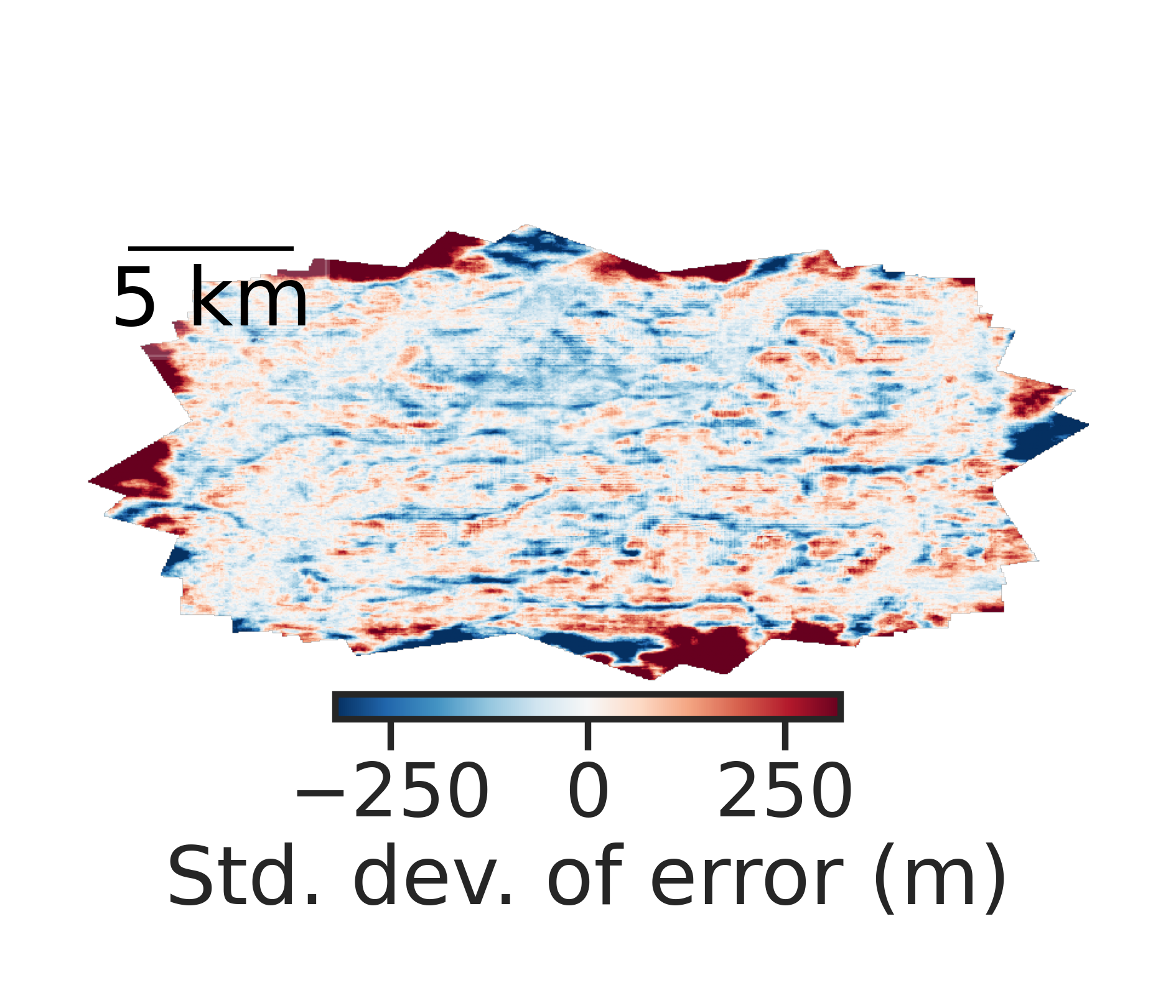} &
      \gescolor\imgbox{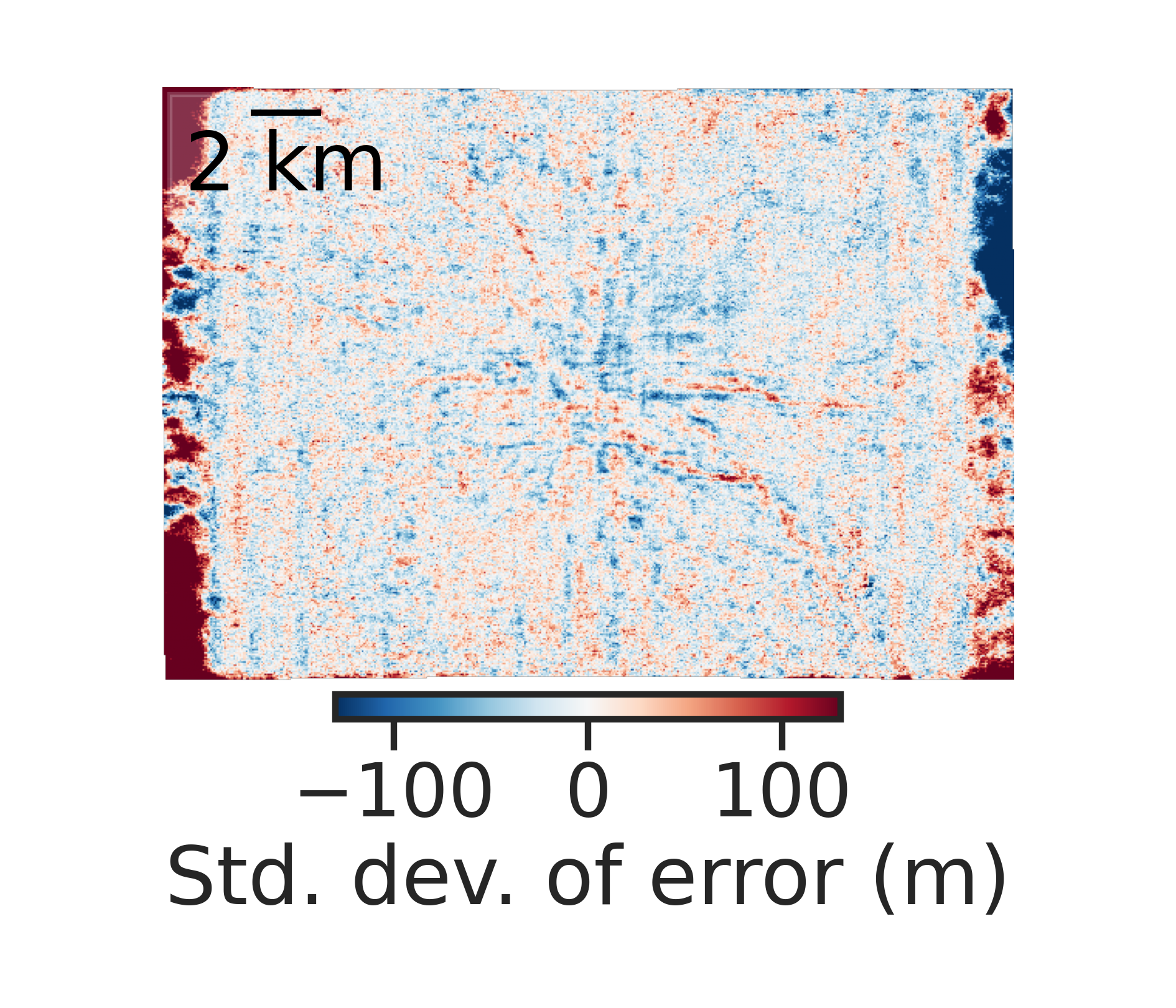} &
      \gescolor\imgbox{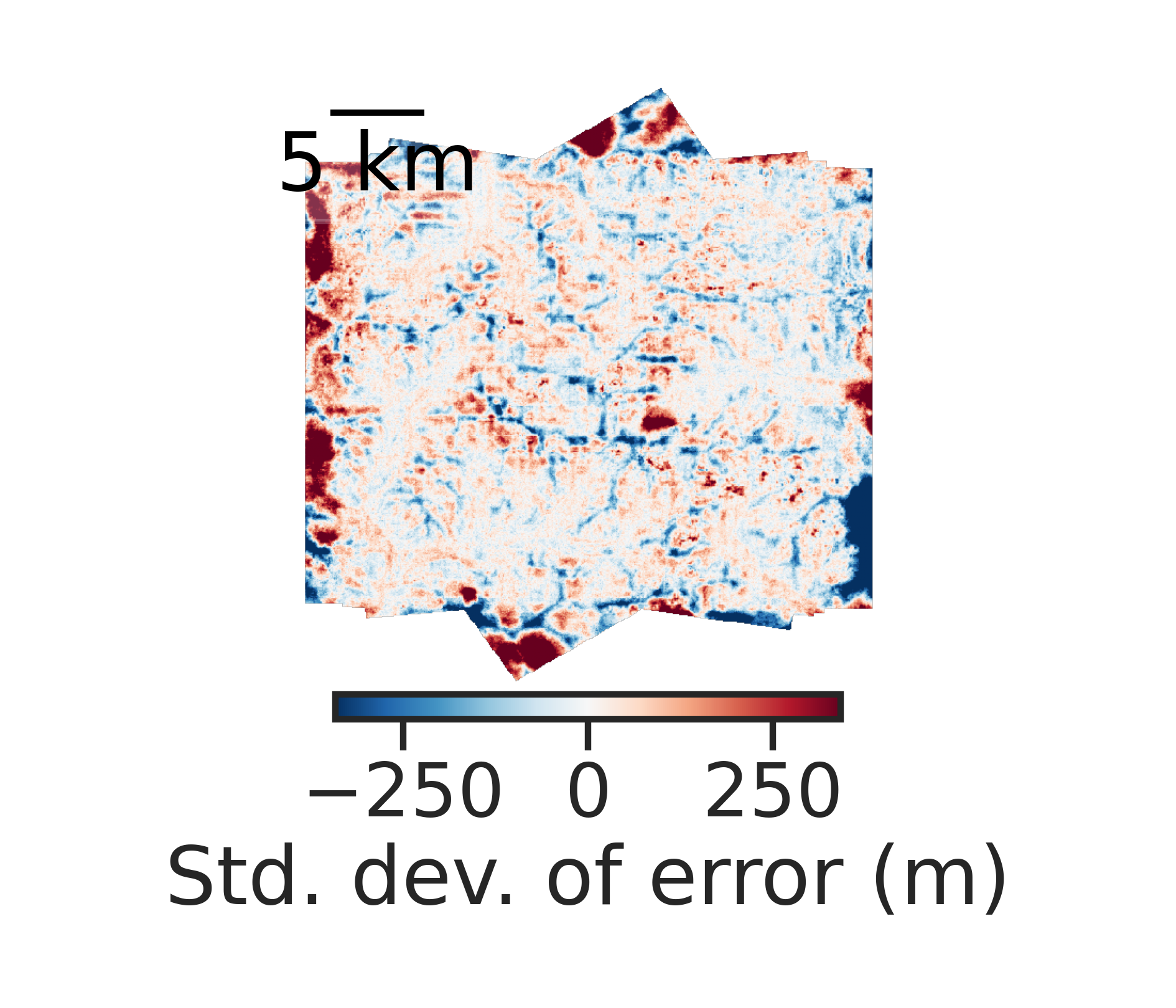} & & \\

      &
      \vertlabel{Error Hist.} &
      \gescolor\imgbox{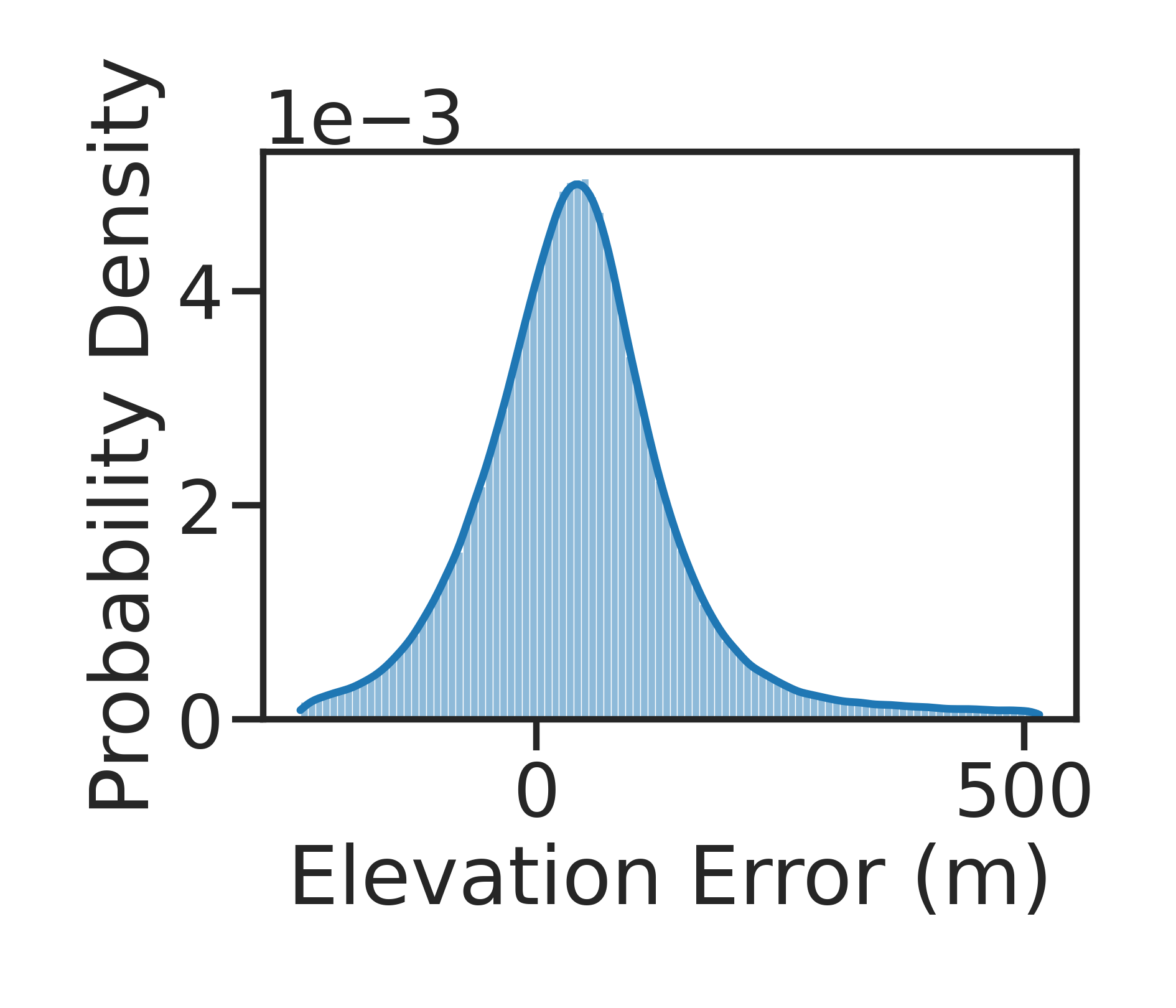} &
      \gescolor\imgbox{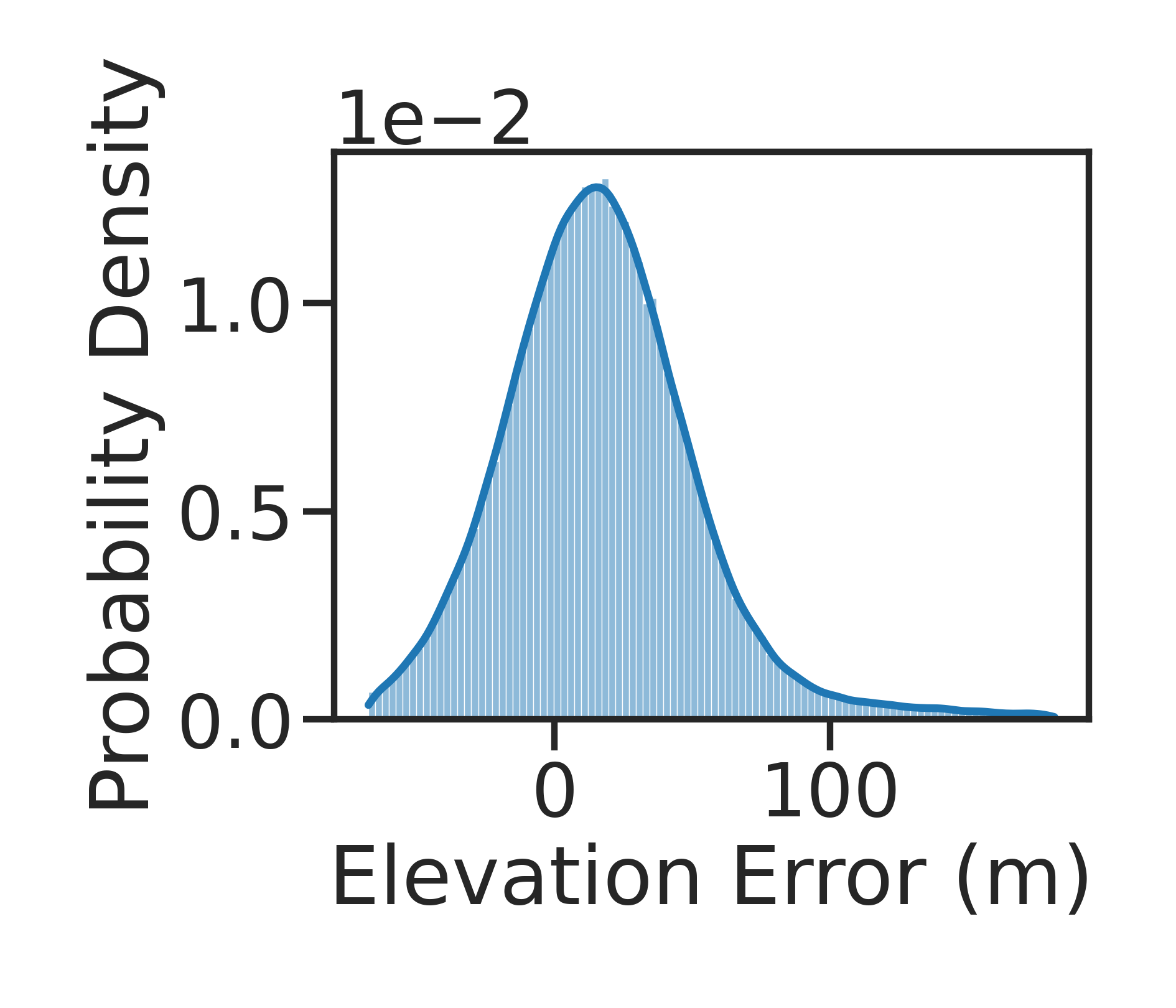} &
      \gescolor\imgbox{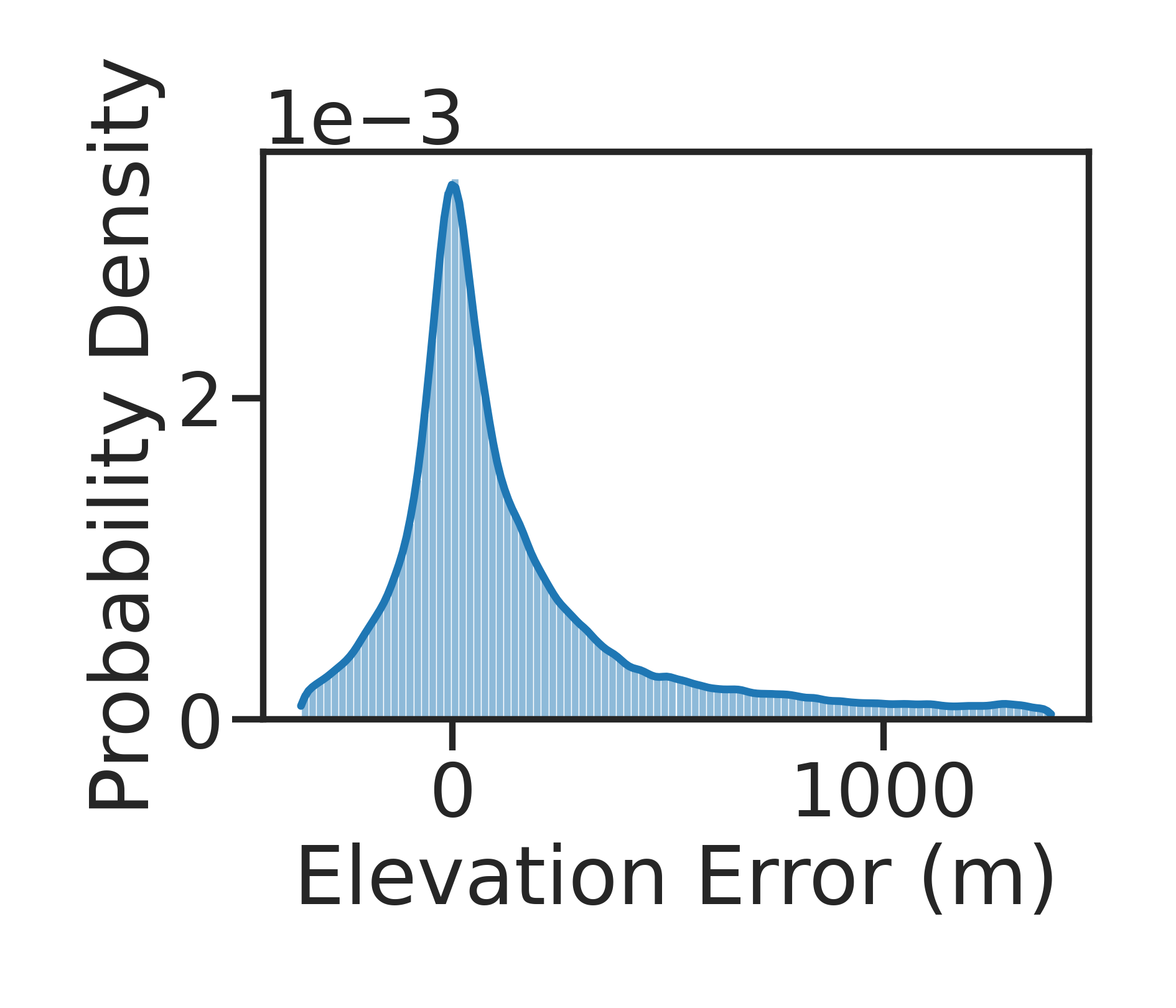} & & \\
    \end{tabular}

  }

  \caption{Results of NTM on 8 scenes from CTX, ASTER, and GES.
  Each column shows the generated terrain map, the standard deviation of the error between the terrain map and a ground truth DTM, and the histogram of the error.
  The first two columns show results for CTX images of Gale and Jezero craters on Mars, the next two columns show results for ASTER images of Everest and Greenland, and the last three columns show results for GES images of Denali, Shasta, and Everest.}
  \label{fig:dtm-results}
\end{figure*}

\subsubsection{Image Processing}
In order to use an image in NTM, like in any standard neural volume rendering method, we require a camera model which allows us to construct the locus for each pixel in the image.
For pinhole cameras, this is straightforward, and, since this model is available for GES, we use that directly.
For linescan cameras, such as CTX and ASTER, a common solution is to rely on an RPC model to construct loci.
However, RPC models are not available for CTX and ASTER images (and are often not available for satellite imagery of planetary bodies other than Earth).
Therefore, we utilize the linescan camera model directly.
To do this, we rely on SPICE~\footnote{https://naif.jpl.nasa.gov/naif/toolkit.html} to provide ephemerides for the MRO and Terra satellites such that we can obtain the camera position and orientation at the time of image acquisition.
We use the ISIS3~\cite{kelvin_rodriguez_integrated_2024} software to process the images from raw data records (which also performs radiometric calibration and removes some common imaging artifacts) and the ALE~\cite{paquette_abstraction_2023} library to create community sensor models (CSM) for each image.
These CSM can then be read by Knoten~\footnote{https://github.com/DOI-USGS/knoten} to construct the pixel loci.

\begin{table*}[htbp]
  \centering
  \begin{tabular}{lccccc}
\hline
DTM & Mean Error (m) & Std Dev (m) & GSD (m) & \# Training Images \\
\hline
Gale Crater (CTX) & \qty{93.17}{\meter} & \qty{36.61}{\meter} & \qty{6}{\meter} & 19 \\
Jezero Crater (CTX) & \qty{117.27}{\meter} & \qty{19.46}{\meter} & \qty{6}{\meter} & 49 \\
Everest (ASTER) & \qty{24.82}{\meter} & \qty{497.94}{\meter} & \qty{15}{\meter} & 8 \\
Gunnbjørn Fjeld (ASTER) & \qty{140.27}{\meter} & \qty{309.21}{\meter} & \qty{15}{\meter} & 4 \\
Denali (GES) & \qty{44.03}{\meter} & \qty{105.30}{\meter} & \qty{60}{\meter} & 31 \\
Everest (GES) & \qty{7.74}{\meter} & \qty{96.47}{\meter} & \qty{60}{\meter} & 31 \\
Shasta (GES) & \qty{16.86}{\meter} & \qty{35.00}{\meter} & \qty{60}{\meter} & 31 \\
\hline
\end{tabular}

  \caption{Summary of DTM errors compared to the reference for each dataset.
    The error is calculated as $h_{\text{DTM}} - h_{\text{ref}}$.
    The top and bottom \qty{2}{\percent} of the error distribution are not considered to exclude outliers.
    Also shown are the approximate GSD for each instrument which serve as an approximate lower bound of the achievable accuracy of the DTM.
  }
  \label{tab:dtm-error-table}
\end{table*}

\begin{figure*}[htbp]
  \centering
  \renewcommand{\arraystretch}{1.5}
  \setlength{\tabcolsep}{4pt}
  \def\ctxcolor{\cellcolor{blue!15}}
  \def\astercolor{\cellcolor{green!15}}
  \def\gescolor{\cellcolor{red!15}}
  \begin{tabular}{>{\centering\arraybackslash}m{1cm}
      >{\centering\arraybackslash}m{3.3cm}
      >{\centering\arraybackslash}m{3.3cm}
      >{\centering\arraybackslash}m{3.3cm}
    >{\centering\arraybackslash}m{3.3cm}}
    & \ctxcolor\textbf{Gale (CTX)} & \ctxcolor\textbf{Jezero (CTX)} & \astercolor\textbf{Gunnbjørn Fjeld (ASTER)} & \astercolor\textbf{Everest (ASTER)} \\
    \rotatebox{90}{\textbf{NTM Novel View}}
    & \ctxcolor\includegraphics[width=3.3cm,height=3.3cm]{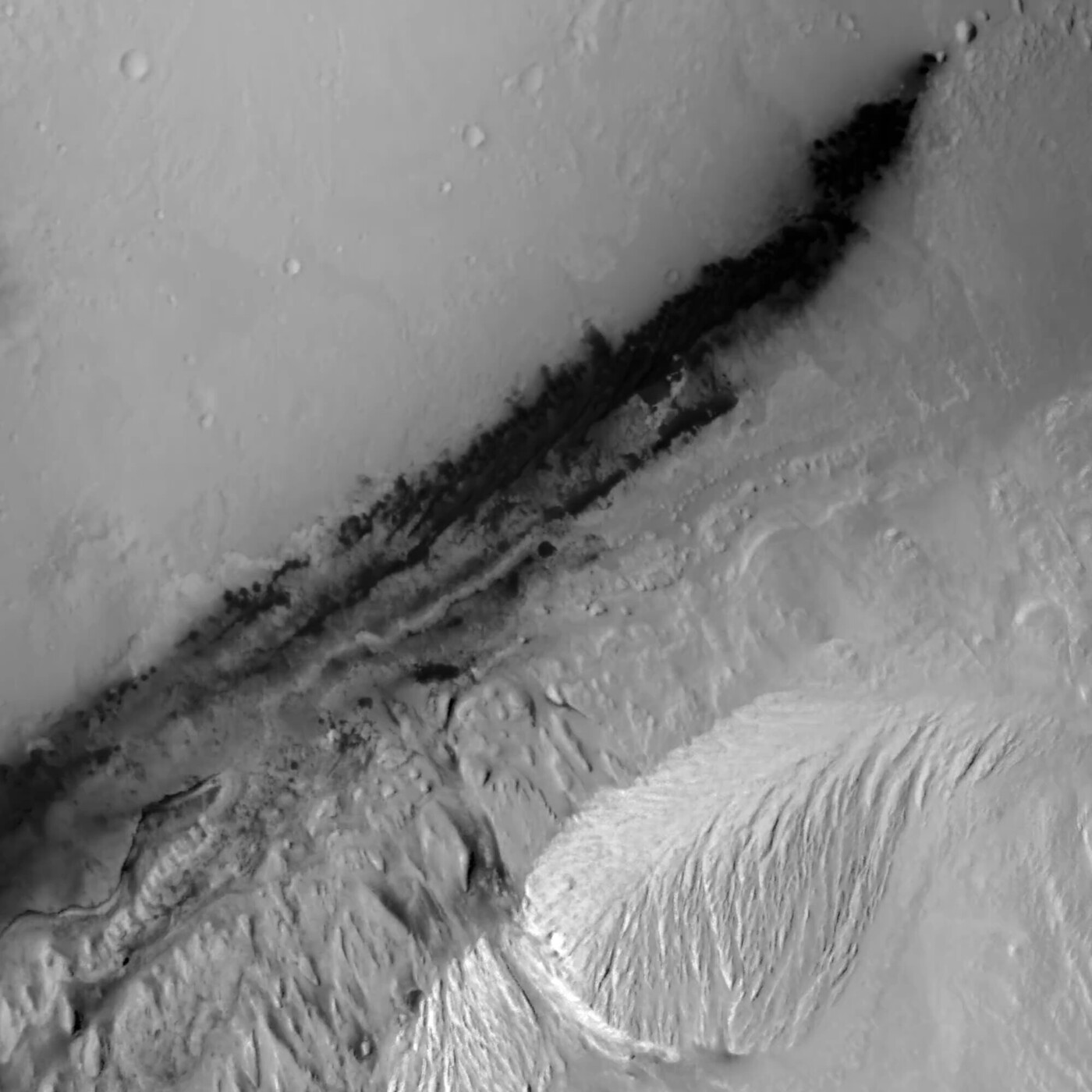}
    & \ctxcolor\includegraphics[width=3.3cm,height=3.3cm]{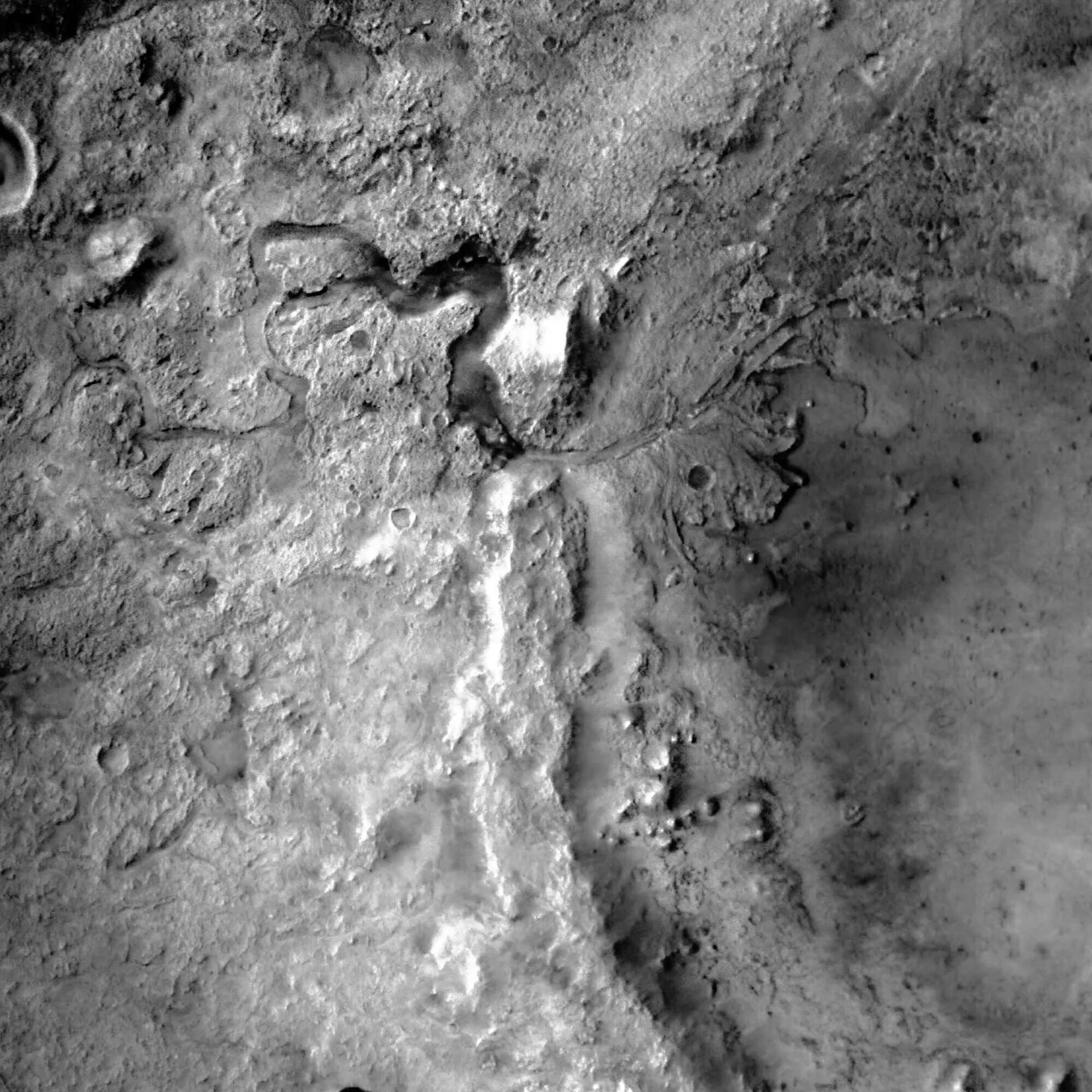}
    & \astercolor\includegraphics[width=3.3cm,height=3.3cm]{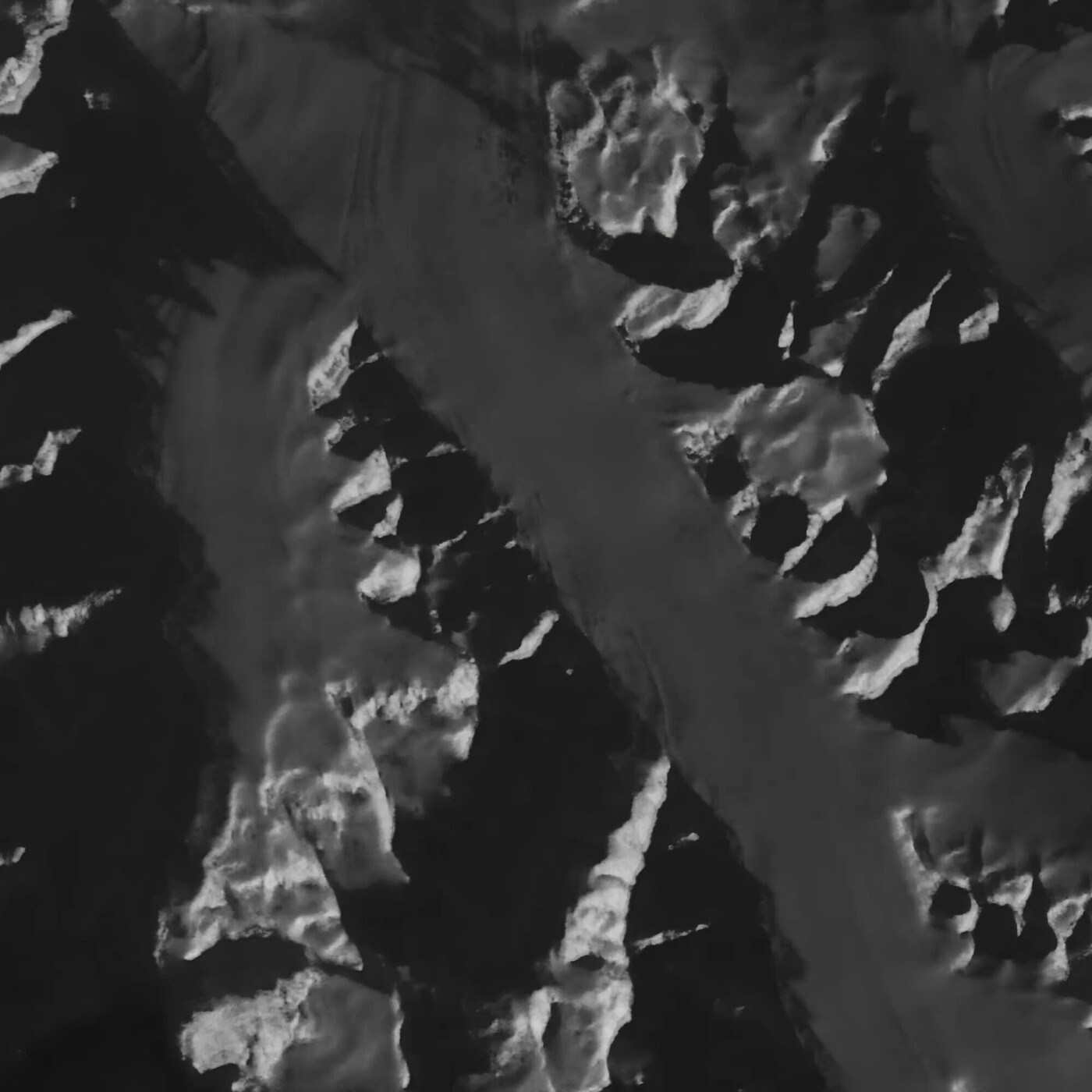}
    & \astercolor\includegraphics[width=3.3cm,height=3.3cm]{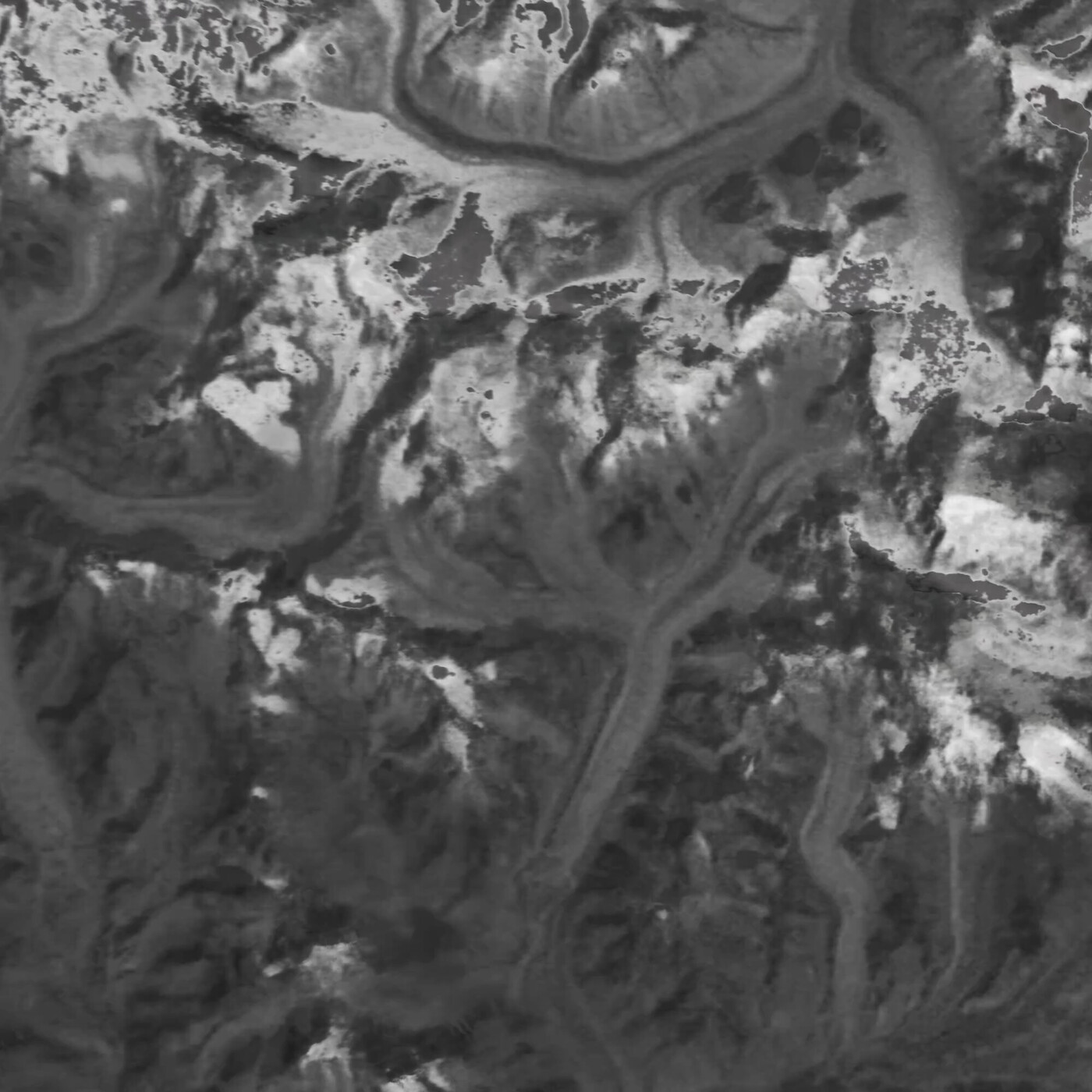} \\
    \rotatebox{90}{\textbf{NTM Texture}}
    & \ctxcolor\includegraphics[width=3.3cm,height=3.3cm]{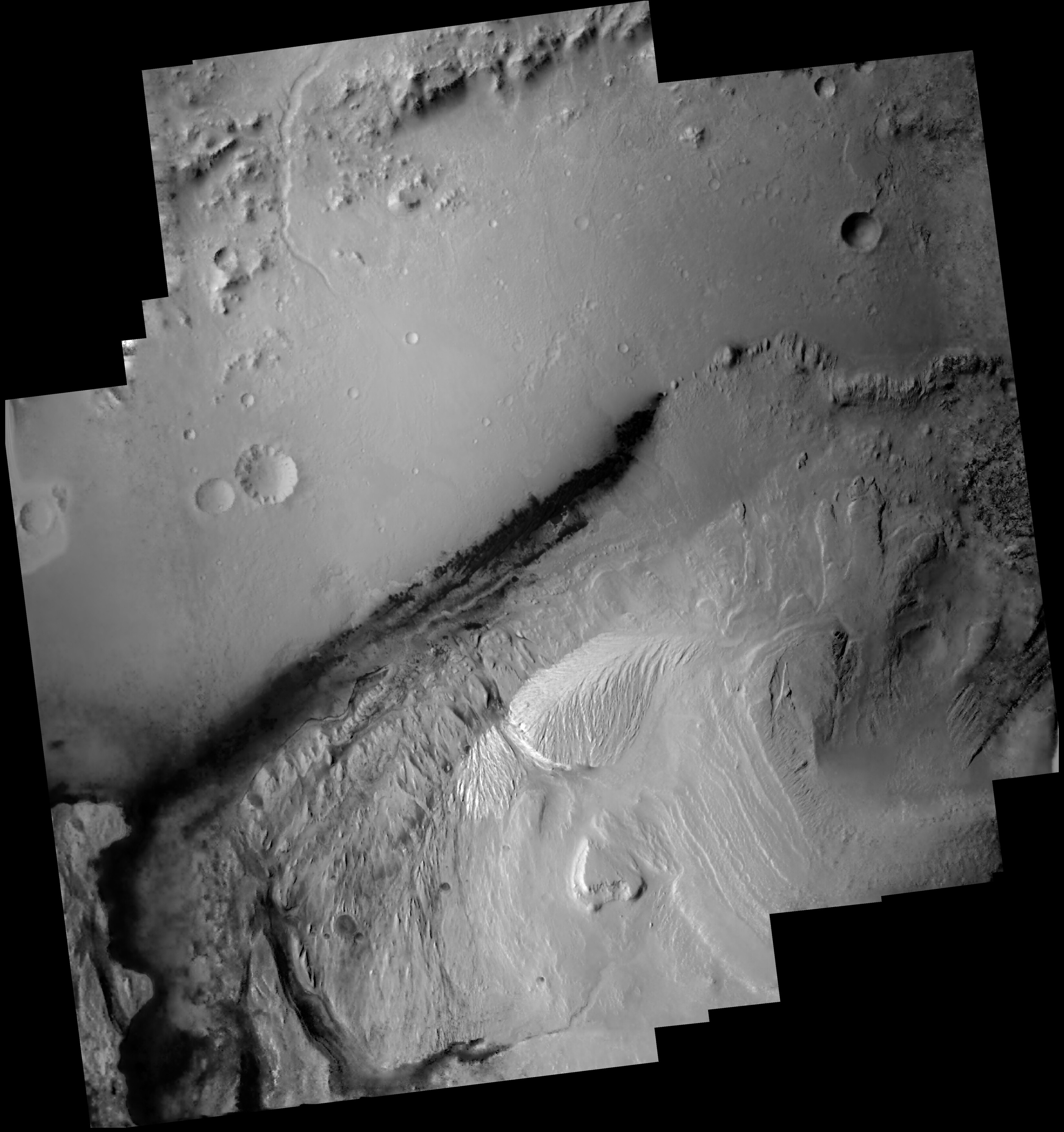}
    & \ctxcolor\includegraphics[width=3.3cm,height=3.3cm]{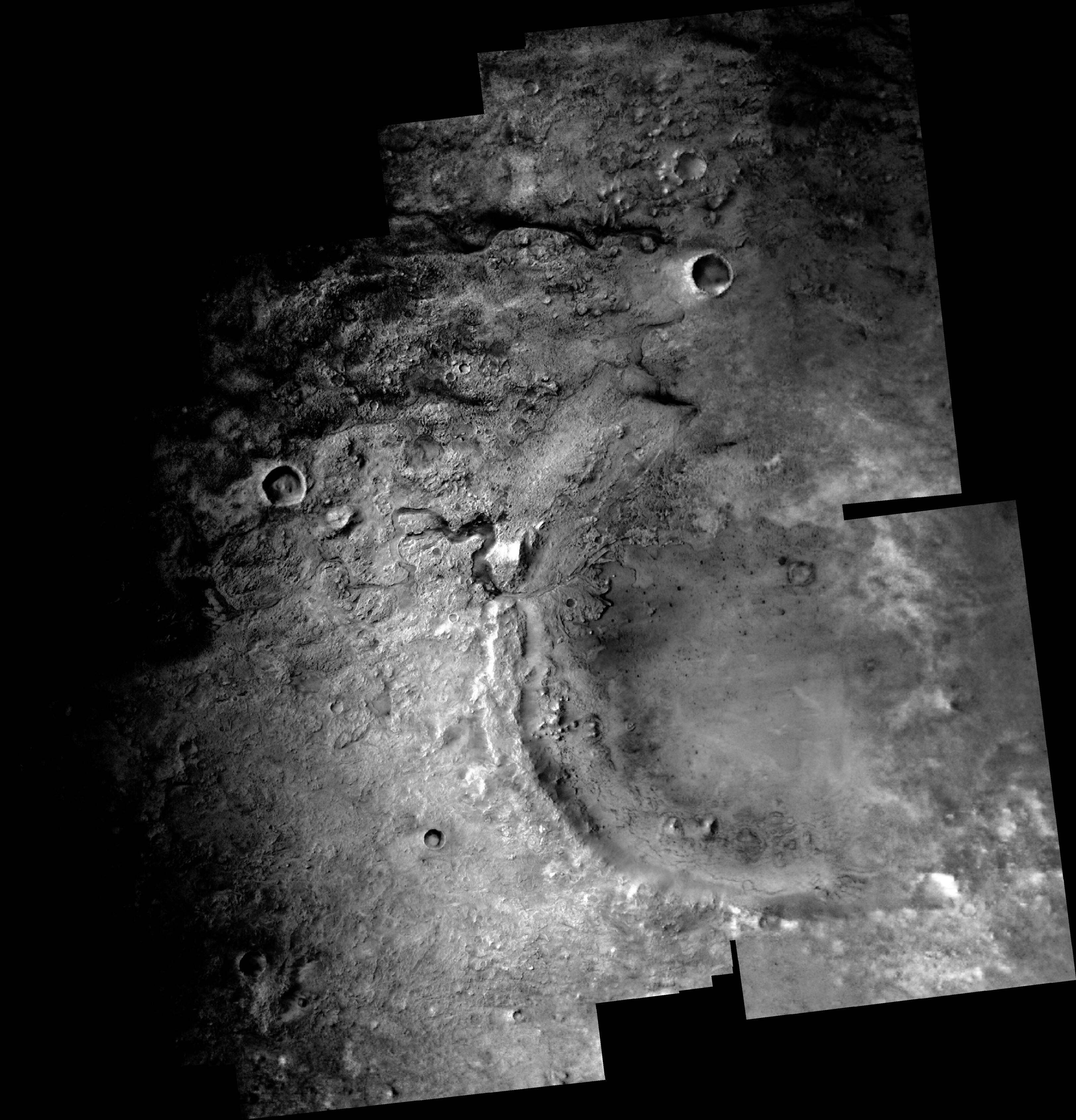}
    & \astercolor\includegraphics[width=3.3cm,height=3.3cm]{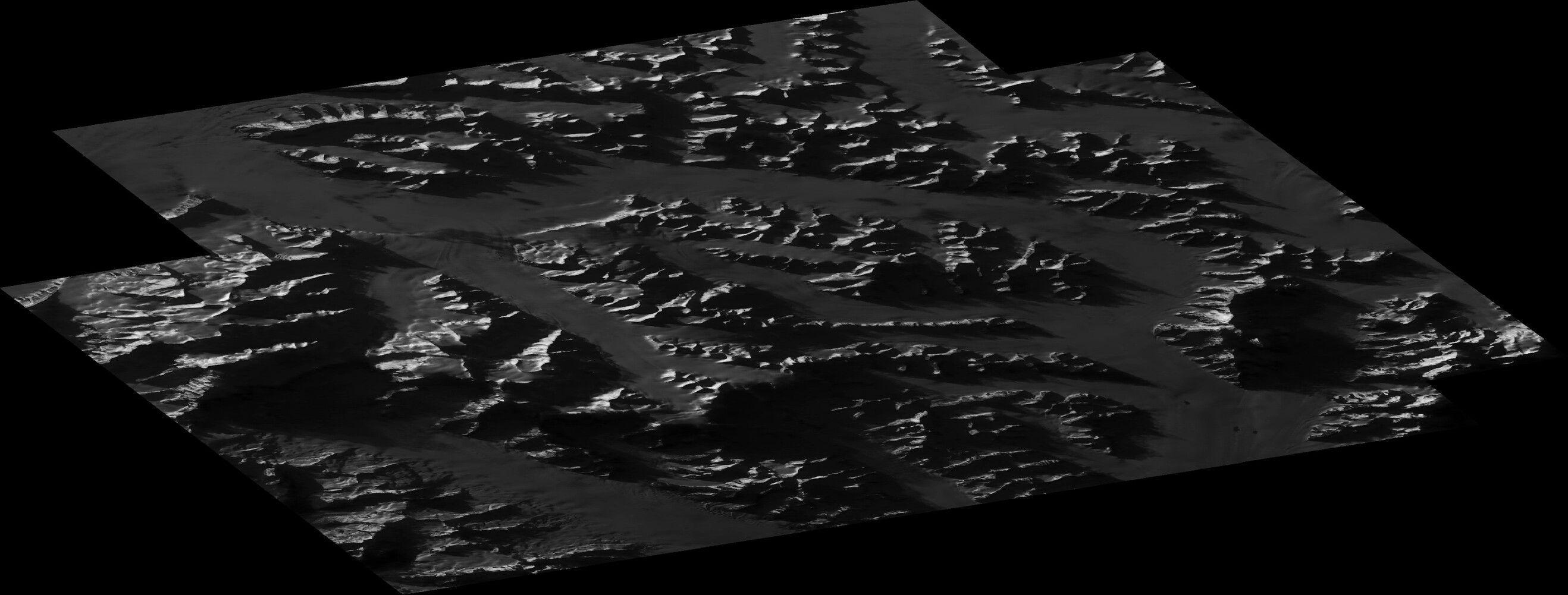}
    & \astercolor\includegraphics[width=3.3cm,height=3.3cm]{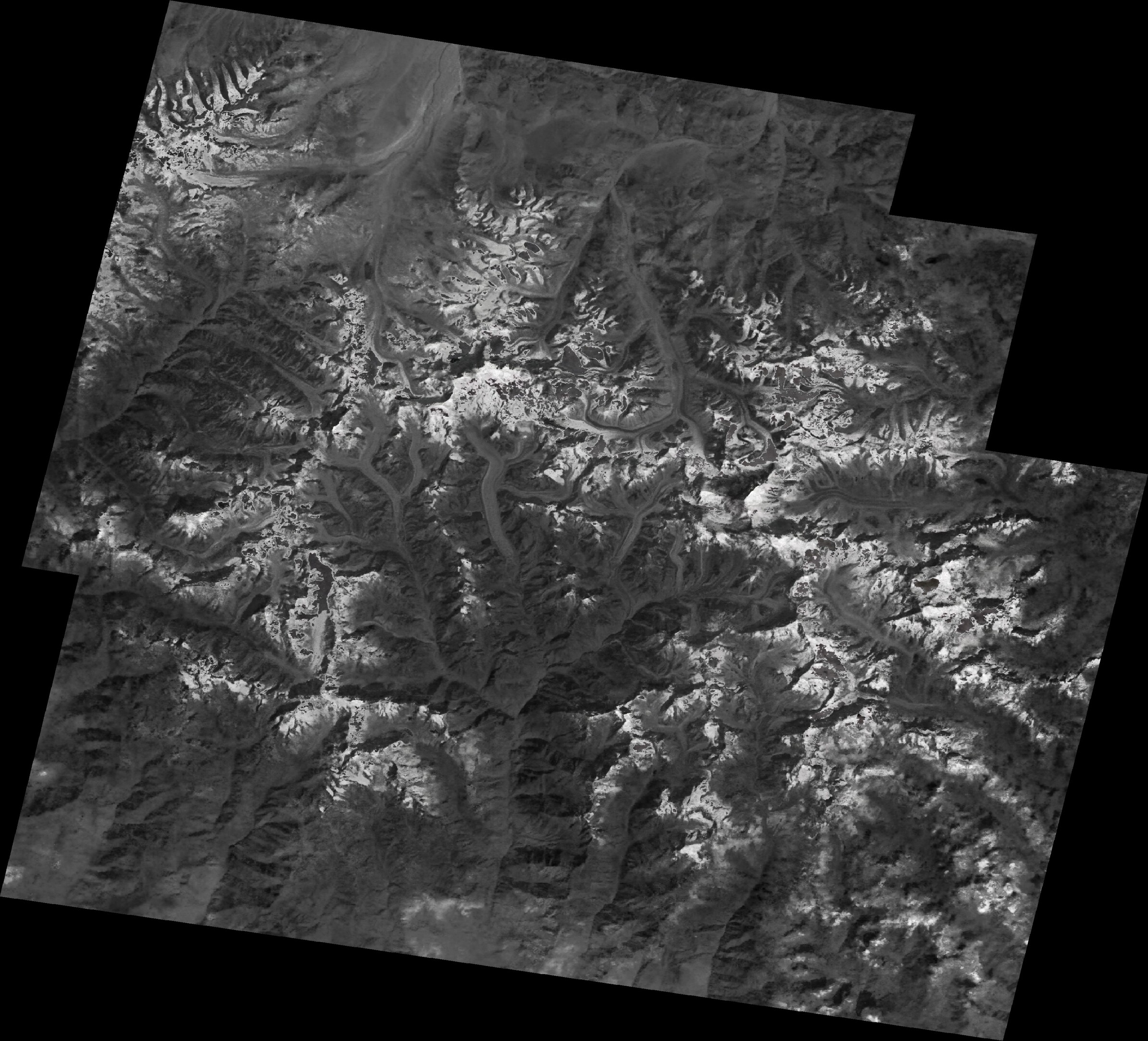} \\
    & \gescolor\textbf{Denali (GES)} & \gescolor\textbf{Everest (GES)} & \gescolor\textbf{Shasta (GES)} & \\
    \rotatebox{90}{\textbf{NTM Novel View}}
    & \gescolor\includegraphics[width=3.3cm,height=3.3cm]{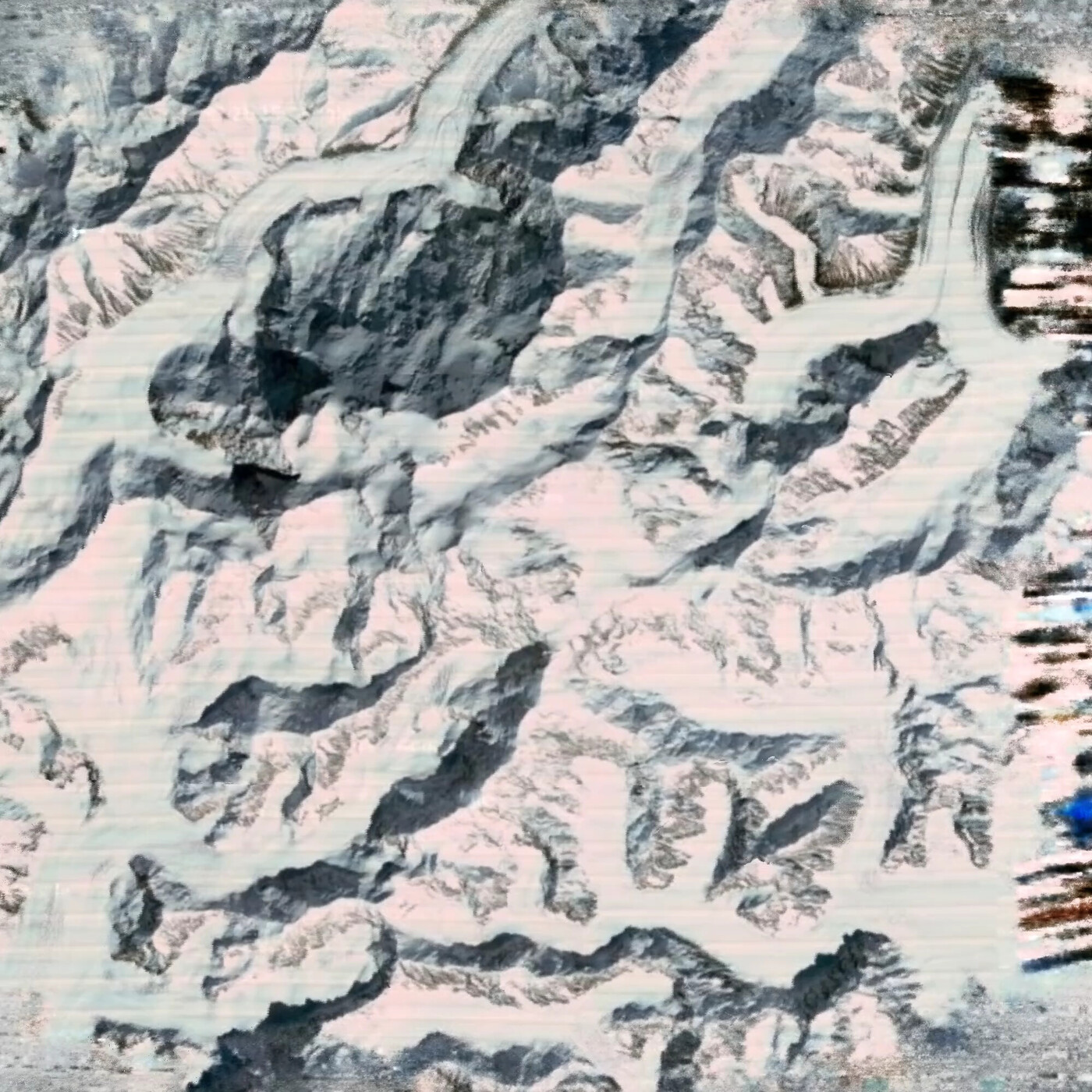}
    & \gescolor\includegraphics[width=3.3cm,height=3.3cm]{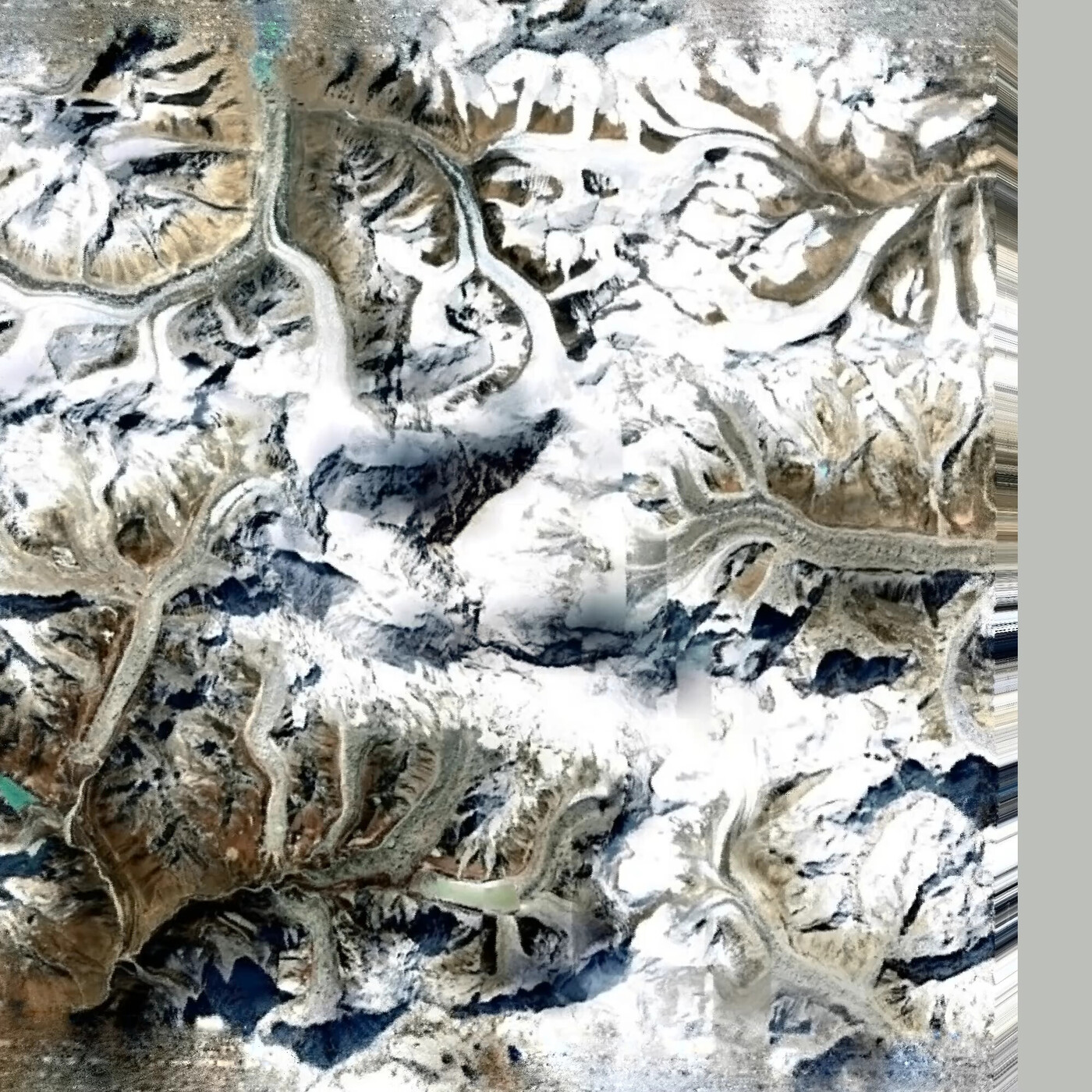}
    & \gescolor\includegraphics[width=3.3cm,height=3.3cm]{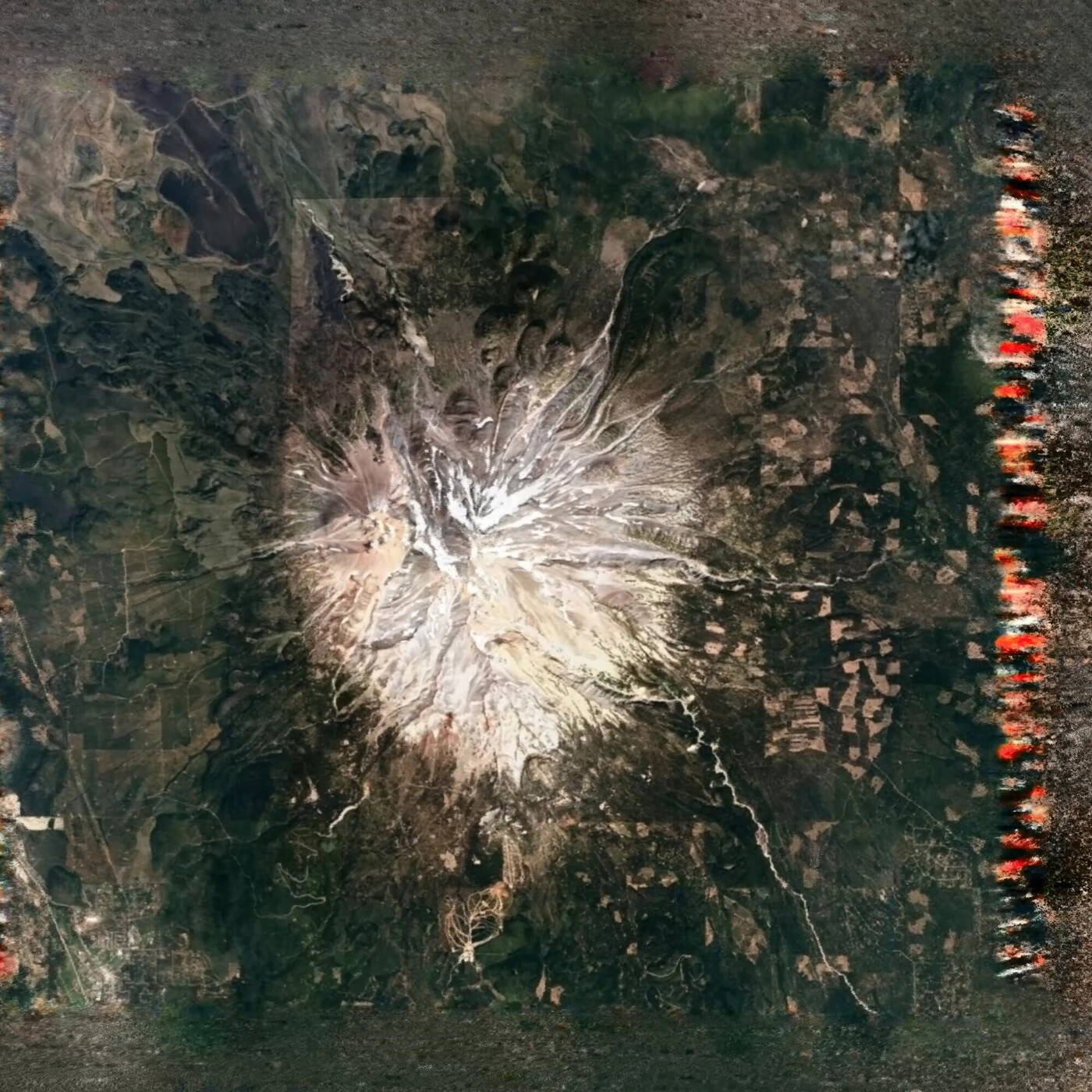}
    & \\
    \rotatebox{90}{\textbf{NTM Texture}}
    & \gescolor\includegraphics[width=3.3cm,height=3.3cm]{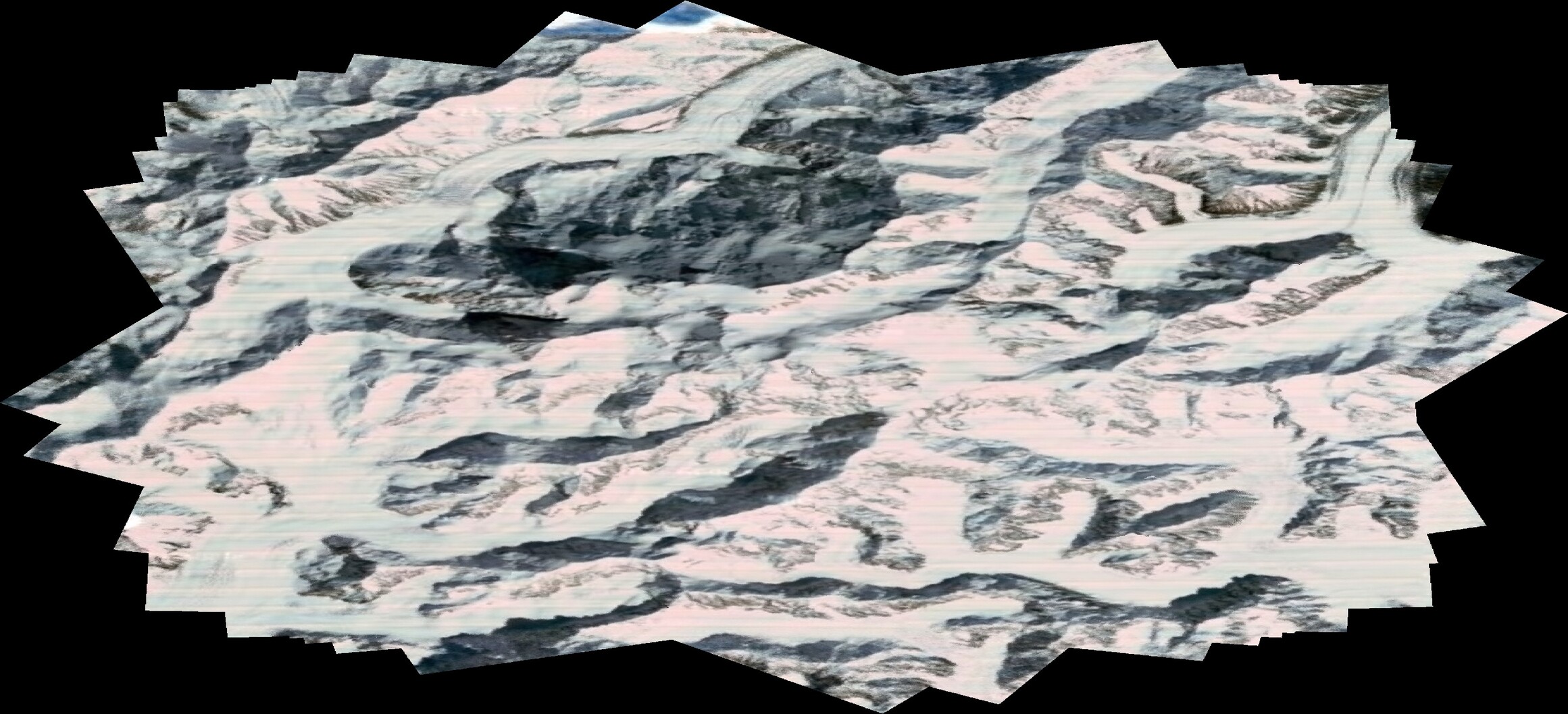}
    & \gescolor\includegraphics[width=3.3cm,height=3.3cm]{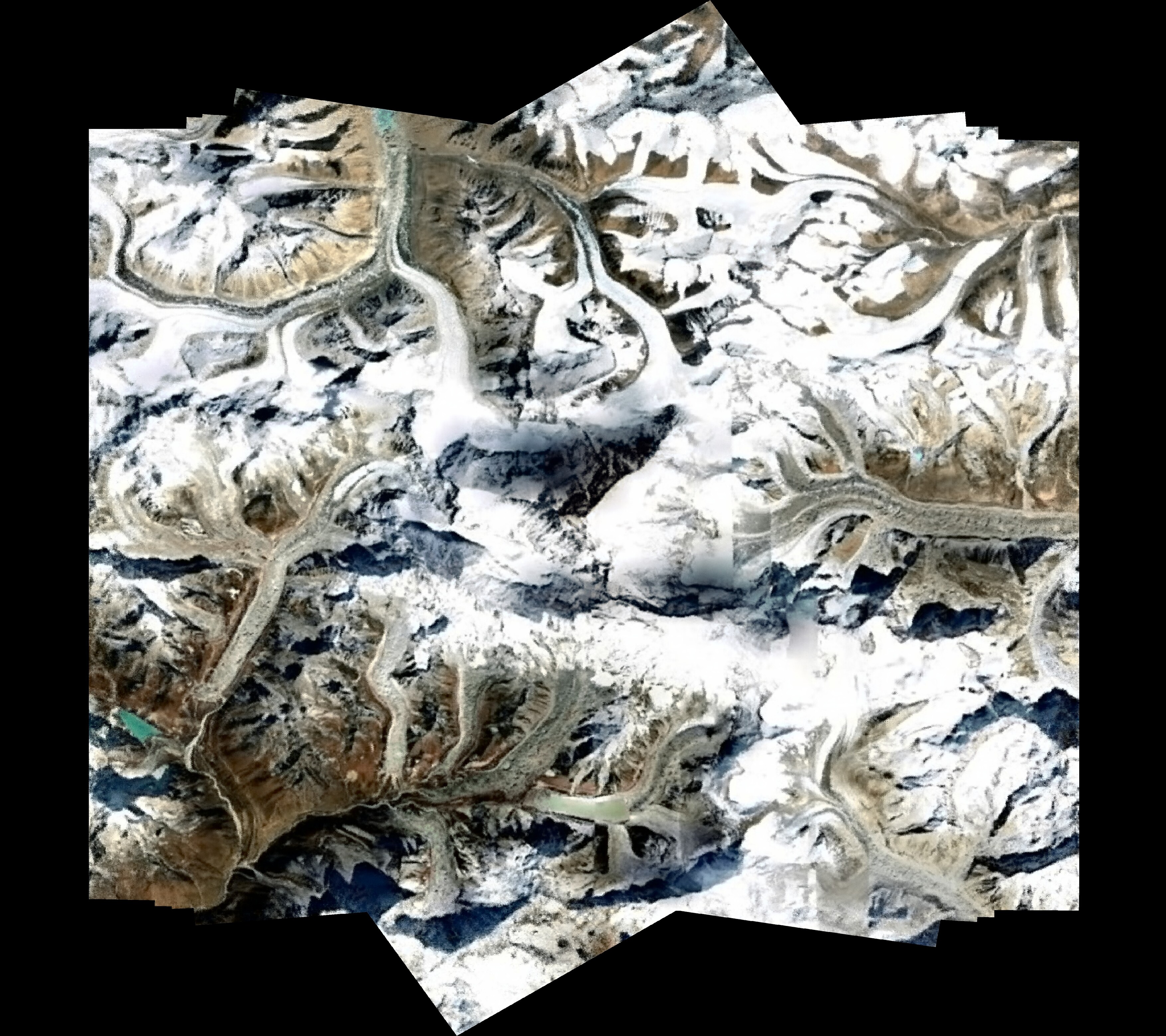}
    & \gescolor\includegraphics[width=3.3cm,height=3.3cm]{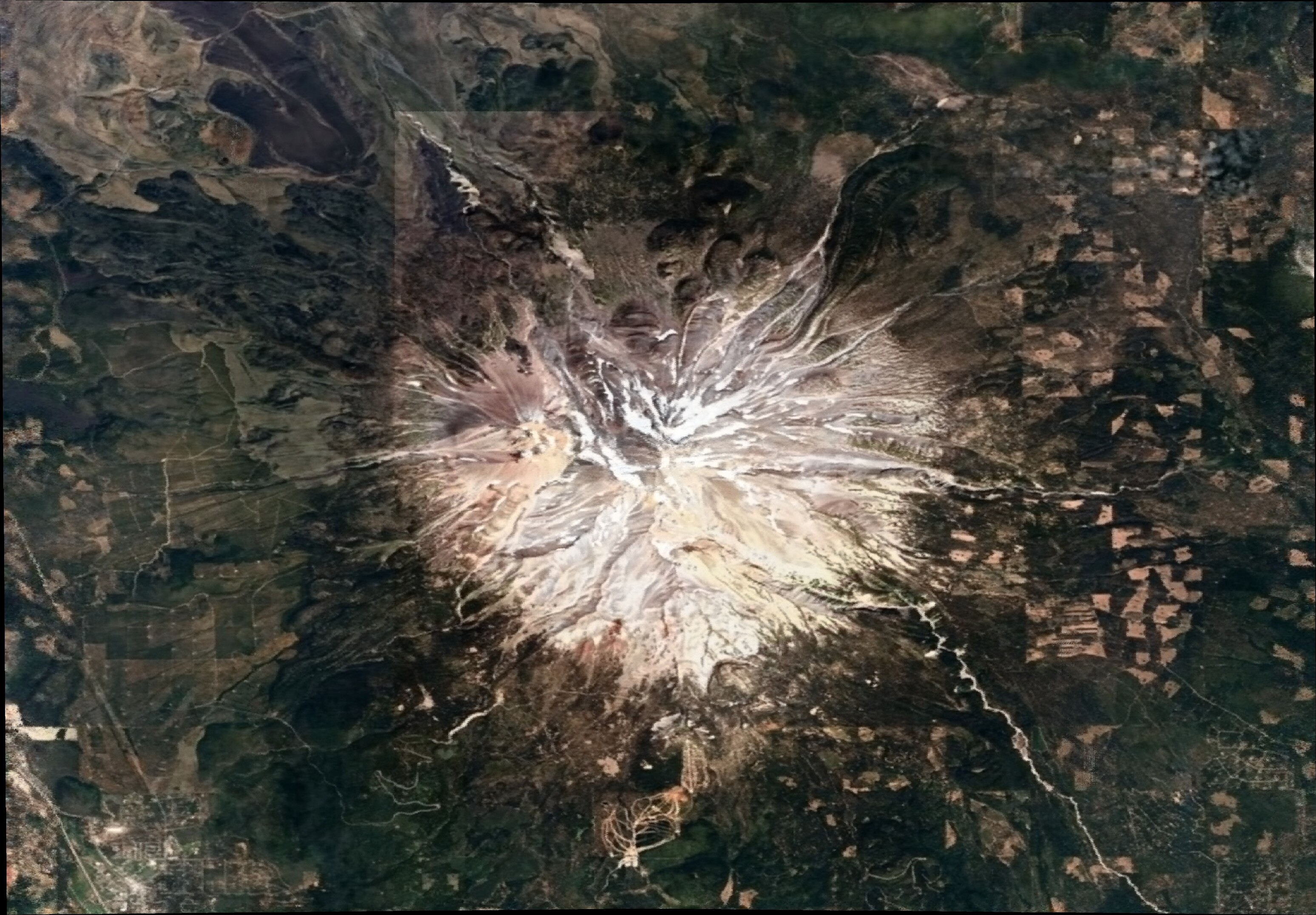}
    & \\
  \end{tabular}
  \caption{Novel view renderings and surface textures generated by NTM.
    The novel views are taken from \qty{30}{\degree} off nadir looking at the center of each scene.
  The textures are taken over the whole scene.}
  \label{fig:qualitative-results}
\end{figure*}

\section{RESULTS AND ANALYSIS}\label{sec:results}
In order to test NTM, we trained on seven scenes using data from CTX, ASTER, and GES.
For the CTX camera, we selected Jezero and Gale craters on Mars.
These are the two most recent landing sites for NASA's Mars rovers and therefore ample imaging data and high quality reference DTMs~\cite{calef_iii_msl_2016,fergason_mars_2020} are available for both areas.
For ASTER, we selected Mount Everest and Gunnbjørn Fjeld in Greenland, the former for its prominence and the latter for its seasonal invariability.
For GES, we selected Mount Shasta in California, Denali in Alaska, and Mount Everest.
For the Earth scenes, the reference DTMs used are derived from the global ASTER DEM~\cite{nasametiaistjapan_spacesystems_and_usjapan_aster_science_team_aster_2019}.
The Gale Crater scene was trained with 19 CTX images, the Jezero Crater scene with 49 CTX images, the Everest scene with 8 ASTER images, the Gunnbjørn Fjeld scene with 4 ASTER images.
For each scene, the images were chosen by selecting all images from the instrument that overlapped with the geographic coordinate of the feature of interest by a geospatial search conducted on the Mars Orbital Data Explorer~\footnote{https://ode.rsl.wustl.edu/mars/index.aspx} or the NASA Earthdata Search~\footnote{https://search.earthdata.nasa.gov/}.
Images were filtered using metadata to exclude those with significant cloud cover or acquisition errors from the final training datasets.
For the GES data, 31 rendered images were used for each scene.
The specific images used in each real scene are listed by their identifiers in Table~\ref{tab:image-ids}.
\par
We trained NTM on each scene for 100k iterations with the Adam optimizer at a learning rate of \qty{3e-4}{}.
The proposal networks in our sampler were trained at a learning rate of \qty{1e-2}{}.
We used a batch size of 2048 and trained on all images in each dataset.
The network was trained on an NVIDIA RTX 6000 Ada and training takes approximately two hours for each scene.
\par
DTM and surface textures are extracted from the network by regularly sampling within the predefined bounding box for each dataset.
These outputs are then georeferenced for analysis.
In order to consistently evaluate each scene, approximate footprints for all images used to train a scene are calculated.
The union of these is used to create a mask and clip the raw DTM sampled from the network.
\par
Quantitative results for the comparison of each generated DTM and the reference are summarized in Table~\ref{tab:dtm-error-table} and visualized in Fig.~\ref{fig:dtm-results}.
For overall quality assessment of the generated DTM, we focus on the standard deviation of the error since the mean deviation can be easily corrected with a sparse depth measurement and correct relative terrain geometry is more important for navigation applications.
We observe that for instruments with a higher GSD, we are able to achieve a lower standard deviation of the error.
In fact, we are often within a single order of magnitude of the ground sample distance, as can be seen in the Gale and Jezero Crater scenes.
The standard deviation is largest on the scenes produced with ASTER data.
This may be explained by the the fact that comparatively few images were used to train these scenes.
This effect is echoed in the CTX scene results, where the Jezero Crater scene, which was trained with more than twice the number of images as the Gale Crater scene, has a significantly lower standard deviation.
Looking at the spatial variation of the error in the ASTER scenes, we also note that there are significant artifacts along the edges of the scene that may also impact the standard deviation of the error.
Examining the mean error, we note that for our synthetic GES scenes, for which we have perfect camera intrinsics and extrinsics and for which the illumination is consistent across all images, the mean error is lower than the GSD of the data.
From this, we expect that the source of this error is due to the presence of uncertainty in the camera model or ephemerides in the real data, or due to transient phenomena in the images.
\par
Qualitative rendering results can be seen in Fig.~\ref{fig:qualitative-results}.
An advantage of using a neural volume rendering approach is that novel views of the scene can be rendered.
We demonstrate this by rendering \qty{30}{\degree} off-nadir perspective camera views of each scene using a \qty{10}{\degree} field of view.
These can be compared against the full texture which is georeferenced and clipped in the same manner as the DTM.

\subsection{Systematic Terrain Error in Extreme Lighting Conditions}
\begin{figure}[h!]
  \centering
  \begin{subfigure}{0.22\textwidth}
    \centering
    \includegraphics[width=\linewidth]{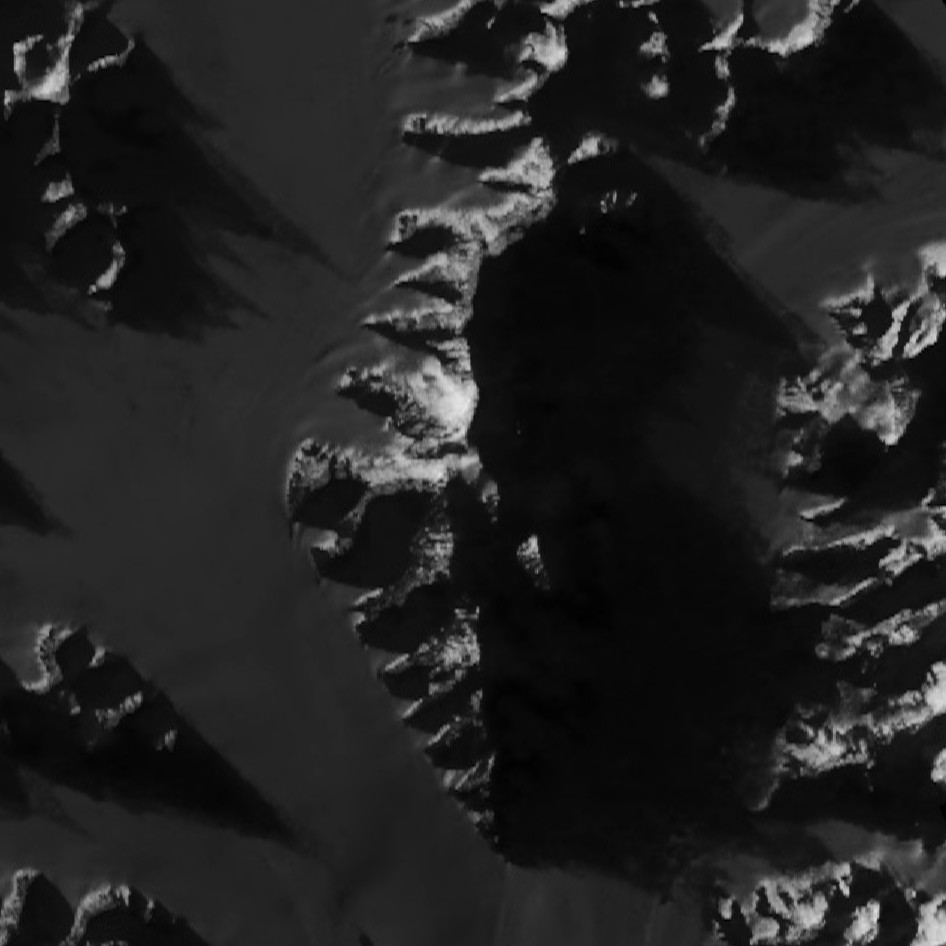}
    \caption{Texture}
    \label{fig:system-light-a}
  \end{subfigure}
  \hfill
  \begin{subfigure}{0.22\textwidth}
    \centering
    \includegraphics[width=\linewidth]{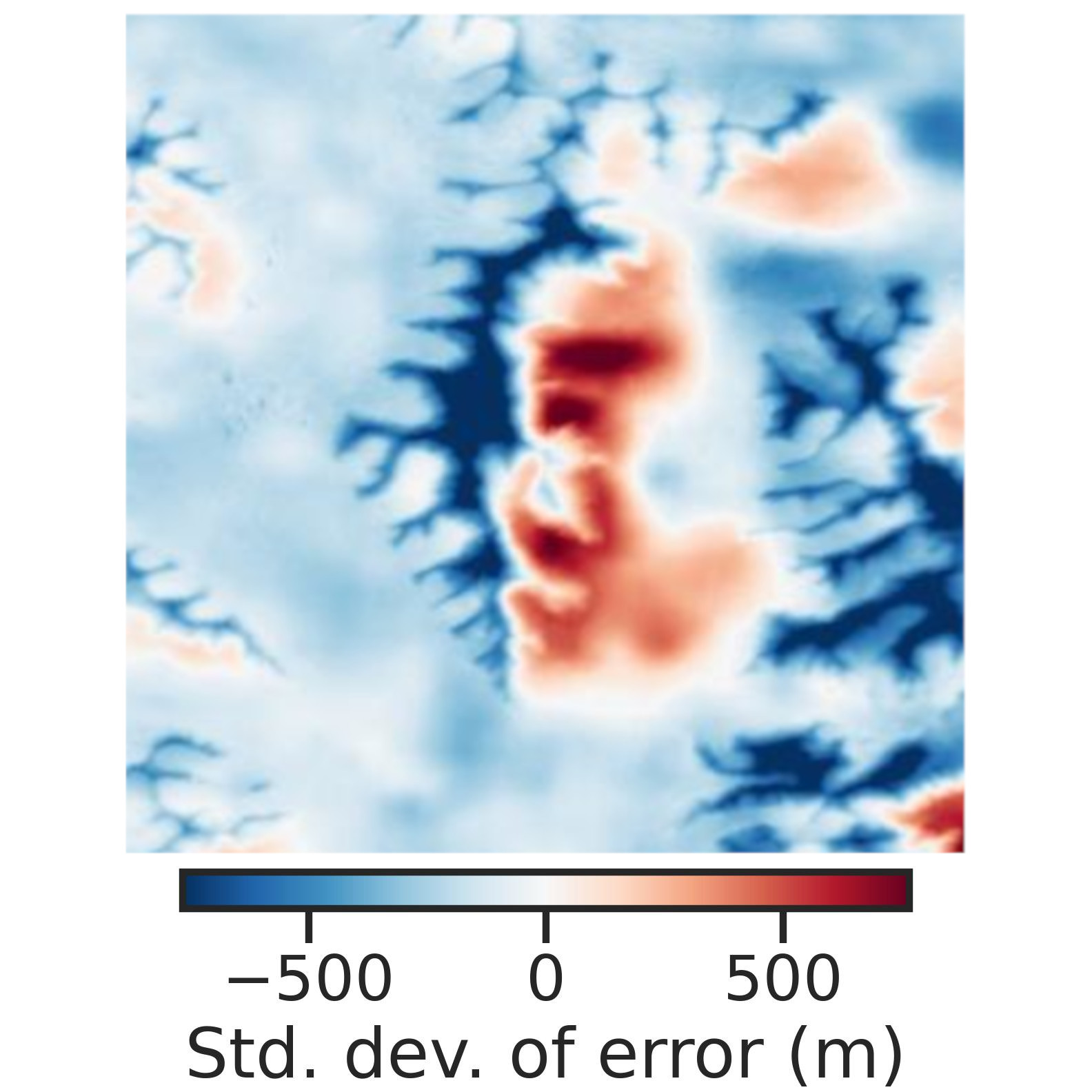}
    \caption{Error std. dev.}
    \label{fig:system-light-b}
  \end{subfigure}
  \caption{Systematic under and overestimation of terrain height in illuminated and shadowed areas respectively in the Gunnbjørn Fjeld scene Southeast corner mountain range.}
  \label{fig:system-light}
\end{figure}

Fig.~\ref{fig:system-light-a} shows the texture and error variance from the Gunnbjørn Fjeld scene.
While the terrain and lighting in this scene are relatively consistent between images due to the instruments sun-synchronous orbit and the glacial terrain, the solar azimuth is quite low, resulting in observable and extreme shadows.
We note that the magnitude of the error is significantly higher in areas that are in shadow.
This is very pronounced in the Southeast corner of the scene as seen in Fig.~\ref{fig:system-light-b}.
This suggests that integrating a lighting model into the rendering process could lead to improved results in these areas.
However, observations of a given area at different solar azimuths might not be available from a given instrument.
This might be addressed by integrating data from multiple instruments or by integrating additional frequency bands from the same instrument.

\subsection{Systematic Terrain Error in Areas with High Relief Gradient}
\begin{figure}[h!]
  \centering
  \includegraphics[width=\linewidth]{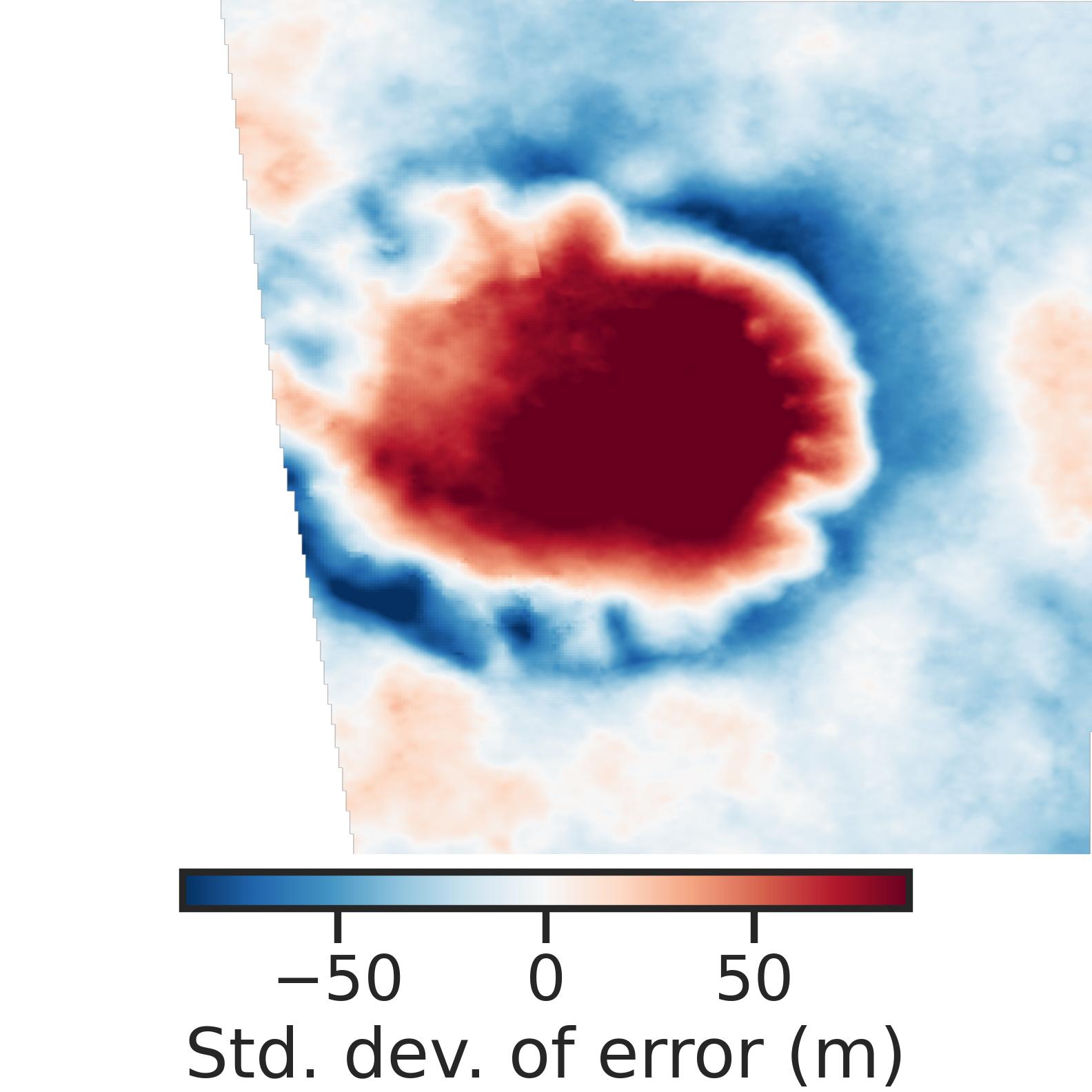}
  \caption{Systematic underestimation and overestimation of terrain height in small crater in West of Gale Crater scene.}
  \label{fig:small-crater}
\end{figure}

Another possible source of error is areas of the terrain that are changing quickly (i.e., have a high relief gradient).
In Fig.~\ref{fig:small-crater}, we show a small crater in the West of the Gale Crater scene.
This crater has a similar pattern of over and underestimation of terrain height as the Gunnbjørn Fjeld scene but is not shaded significantly in any of the input images.
We theorize that this error is due to a limitation of constraining solutions to only represent height maps, which prevents different colors from being assigned to the same planar coordinate at different heights.
For example, in the extreme case of the network being asked to represent a vertical wall --- which cannot be precisely represented by a height map --- it would be required to introduce a slope to the terrain to explain all observed colors, sacrificing terrain accuracy for a reduced loss.
Outside of such an extreme, higher verticality may still act in a similar fashion by requiring high-frequency, difficult to learn color changes in comparison to flatter areas.

\section{CONCLUSION}\label{sec:conclusion}
In this work, we outline a new procedure for generating DTMs from multi-view satellite imagery using volume rendering to train a neural representation of the terrain map.
Our method only requires camera intrinsics and extrinsics for each image and does not rely on any depth priors.
Furthermore, we do not require RPC models for images, meaning that our method is applicable to planetary bodies where prior terrain information required to fit these models is not available.
We demonstrate our method on both synthetic and real satellite imaging datasets from Earth and Mars.
We show that our method is able to produce accurate and coherent terrain maps across a variety of scenes when traditional volume rendering methods struggle.
This is a promising step toward a new method for generating DTMs for planetary exploration that can be used to support future missions to locations that are not well covered by existing orbital imaging datasets and for which no depth prior is available.

\subsection{Future Work}
Neural volume rendering approaches allow a great deal of flexibility in modeling which we have yet to fully explore.
For example, we argued that integrating an illumination model into our method may have the potential to increase the accuracy of recovered terrain.
This may be accomplished in a similar way to~\cite{mari_multi-date_2023}, although we may achieve better results since our model provides a clearer signal for surface intersection from which to perform solar raycasting.
Transient phenomena, such as clouds, dust storms, or seasonal changes can also be readily handled~\cite{martin-brualla_nerf_2021}.
Additionally, the color network can be swapped for an explicit albedo and radiance model to allow areas of the field to emit radiation in physically meaningful quantities.
This would allow for imaging data from multiple frequency bands to be utilized in training the same network.
Finally, allowing the network to bundle adjust data during training could compensate alignment errors in initial pose estimates derived from ephemerides for two different instruments, allowing imagery from both to be combined into single scene reconstruction.
Altogether, these changes would allow our method to be used to produce DTMs from images taken from a wide variety of instruments, thus maximizing data utilization.
\par
With regard to computation speed, our current implementation requires us to perform two backwards passes through the network in order to calculate the finite difference estimate used in~\ref{eq:sdf-opacity}.
This slows down training and inference significantly as it prevents the use of the more efficient fully-fused network kernels in~\cite{muller_tiny-cuda-nn_2021}, which are used in most implementations provided in Nerfstudio.
Future work will explore methods of addressing this issue either through the implementation of custom kernels to allow for this double backward pass, or by eliminating the need for this gradient.

\section*{APPENDIX}
Table~\ref{tab:image-ids} lists the unique identifiers for all images used in this work.
{
  \def\ctxcolor{\cellcolor{blue!15}}
  \def\astercolor{\cellcolor{green!15}}
  \begin{table*}[hbtp]
    \centering
    \caption{Image identifiers for all satellite images used in this work.}
    \label{tab:image-ids}
    {
      \renewcommand{\arraystretch}{2.0}
      \begin{tabular}{|c|c|>{\raggedright\arraybackslash}p{10cm}|}
        \hline
        \Large\textbf{Dataset} & \Large\textbf{Scene} & \Large\textbf{Image Identifiers} \\
        \hline
        \ctxcolor
        \multirow{2}{*}{\large\textbf{CTX}} &     \ctxcolor
        \large\textbf{Gale Crater} &
        \ctxcolor
        {{B01\_009861\_1753\_XI\_04S222W, B08\_012841\_1751\_XN\_04S222W, F01\_036128\_1754\_XN\_04S222W, F11\_040269\_1754\_XN\_04S222W, G02\_018854\_1754\_XN\_04S222W, G02\_018920\_1754\_XN\_04S222W, G17\_024735\_1755\_XN\_04S222W, G21\_026357\_1754\_XN\_04S222W, J08\_047772\_1752\_XN\_04S222W, J16\_051042\_1754\_XI\_04S222W, K01\_053732\_1752\_XN\_04S222W, K18\_060418\_1754\_XI\_04S222W, K19\_060840\_1753\_XN\_04S222W, K22\_061961\_1753\_XN\_04S222W, N01\_062673\_1754\_XI\_04S222W, N06\_064519\_1753\_XN\_04S222W, N17\_069031\_1752\_XN\_04S222W, P06\_003453\_1752\_XI\_04S222W, U07\_073567\_1752\_XN\_04S222W}}
        \\
        \cline{2-3}
        \ctxcolor & \ctxcolor \large\textbf{Jezero Crater} & \ctxcolor
        {{F06\_038240\_1991\_XN\_19N282W, F09\_039348\_1984\_XN\_18N282W, F19\_043106\_1991\_XN\_19N282W, F22\_044438\_1991\_XN\_19N282W, G13\_023313\_1985\_XN\_18N282W, G13\_023379\_1985\_XN\_18N282W, G14\_023524\_1985\_XN\_18N282W, G14\_023669\_1985\_XN\_18N282W, J09\_048341\_1991\_XN\_19N282W, J18\_052020\_1984\_XN\_18N283W, J20\_052455\_1984\_XN\_18N282W, J20\_052521\_1984\_XN\_18N282W, J20\_052666\_1984\_XN\_18N282W, J22\_053233\_1984\_XN\_18N282W, J22\_053299\_1983\_XN\_18N283W, J22\_053378\_1984\_XN\_18N282W, J22\_053444\_1983\_XN\_18N283W, J22\_053523\_1983\_XN\_18N283W, K01\_053589\_1983\_XN\_18N283W, K01\_053655\_1983\_XN\_18N283W, K01\_053734\_1984\_XN\_18N282W, K01\_053879\_1984\_XN\_18N282W, K01\_053945\_1984\_XN\_18N282W, K02\_054090\_1984\_XN\_18N282W, K16\_059708\_1985\_XN\_18N282W, K19\_060829\_1986\_XN\_18N283W, K22\_061963\_1988\_XN\_18N282W, K22\_062029\_1988\_XN\_18N282W, K22\_062095\_1988\_XN\_18N282W, K22\_062161\_1987\_XN\_18N282W, K23\_062174\_1986\_XN\_18N283W, K23\_062240\_1985\_XN\_18N283W, K23\_062319\_1986\_XN\_18N282W, K23\_062385\_1988\_XN\_18N282W, K23\_062451\_1988\_XN\_18N282W, K23\_062530\_1986\_XN\_18N282W, N01\_062662\_1988\_XN\_18N282W, N01\_062675\_1988\_XN\_18N282W, N01\_062741\_1988\_XN\_18N282W, N01\_062886\_1988\_XN\_18N282W, N01\_062952\_1988\_XN\_18N282W, N05\_064165\_1986\_XN\_18N282W, N05\_064231\_1982\_XN\_18N282W, N09\_065985\_1986\_XN\_18N282W, N13\_067291\_1986\_XN\_18N282W, P02\_001820\_1984\_XI\_18N282W, P04\_002743\_1987\_XI\_18N282W, U05\_073068\_1986\_XN\_18N282W, U07\_073635\_1986\_XN\_18N282W}}
        \\
        \hline
        \astercolor
        \multirow{2}{*}{\large\textbf{ASTER}} & \astercolor \large\textbf{Everest} & \astercolor
        {{AST\_L1A\_00302012006045809\_20250523121643\_872380, AST\_L1A\_00304182008050534\_20250523121643\_872377, AST\_L1A\_00304202025035146\_20250523121653\_872524, AST\_L1A\_00304242001050941\_20250523121643\_872365, AST\_L1A\_00305282005050453\_20250523121643\_872379, AST\_L1A\_00310232003045929\_20250523121643\_872367, AST\_L1A\_00311132020050523\_20250523121653\_872521, AST\_L1A\_00311252015050006\_20250523121653\_872518}}
        \\
        \cline{2-3}
        \astercolor & \astercolor \large\textbf{Greenland} & \astercolor
        {{AST\_L1A\_00306032024221846\_20250603184520\_2746981, AST\_L1A\_00306272024220425\_20250603184520\_2746986, AST\_L1A\_00307072024221427\_20250603184520\_2746992, AST\_L1A\_00307242024221134\_20250603184520\_2746993}}
        \\
        \hline
      \end{tabular}
    }
  \end{table*}
}

\bibliographystyle{IEEEtaes}
\bibliography{ntm}

\begin{thebibliography}{10}
\providecommand{\url}[1]{#1}
\csname url@samestyle\endcsname
\renewcommand{\newblock}{\par}
\providecommand{\bibinfo}[2]{#2}
\providecommand{\BIBentrySTDinterwordspacing}{\spaceskip=0pt\relax}
\providecommand{\BIBentryALTinterwordstretchfactor}{4}
\providecommand{\BIBentryALTinterwordspacing}{\spaceskip=\fontdimen2\font plus
\BIBentryALTinterwordstretchfactor\fontdimen3\font minus \fontdimen4\font\relax}
\providecommand{\BIBforeignlanguage}[2]{{%
\expandafter\ifx\csname l@#1\endcsname\relax
\typeout{** WARNING: IEEEtran.bst: No hyphenation pattern has been}%
\typeout{** loaded for the language `#1'. Using the pattern for}%
\typeout{** the default language instead.}%
\else
\language=\csname l@#1\endcsname
\fi
#2}}
\providecommand{\BIBdecl}{\relax}
\BIBdecl

\bibitem{balaram_ingenuity_2021}
\BIBentryALTinterwordspacing
J.~Balaram, M.~Aung, and M.~P. Golombek
\newblock  The ingenuity helicopter on the perseverance rover \newblock  vol. 217, no.~4, p.~56. [Online]. Available: \url{https://doi.org/10.1007/s11214-021-00815-w}
\BIBentrySTDinterwordspacing

\bibitem{lorenz_dragonfly_2018}
R.~D. Lorenz, E.~P. Turtle, J.~W. Barnes, and M.~G. Trainer
\newblock  Dragonfly: A rotorcraft lander concept for scientific exploration at titan \newblock  vol.~34, no.~3.

\bibitem{vaquero_eels_2024}
\BIBentryALTinterwordspacing
T.~S. Vaquero \emph{et~al.}
\newblock  {EELS}: Autonomous snake-like robot with task and motion planning capabilities for ice world exploration \newblock  vol.~9, no.~88, p. eadh8332, publisher: American Association for the Advancement of Science. [Online]. Available: \url{https://www.science.org/doi/full/10.1126/scirobotics.adh8332}
\BIBentrySTDinterwordspacing

\bibitem{de_la_croix_multi-agent_2024}
\BIBentryALTinterwordspacing
J.-P. de~la Croix \emph{et~al.}
\newblock  Multi-agent autonomy for space exploration on the {CADRE} lunar technology demonstration \newblock  In \emph{2024 {IEEE} Aerospace Conference}, pp. 1--14, {ISSN}: 1095-323X. [Online]. Available: \url{https://ieeexplore.ieee.org/abstract/document/10521425}
\BIBentrySTDinterwordspacing

\bibitem{schonfeld_summary_2023}
\BIBentryALTinterwordspacing
J.~Schonfeld
\newblock  Summary of the contracted deliveries of {NASA} payloads to the moon via commercial lunar payload services ({CLPS}) \newblock  In \emph{2023 {IEEE} International Conference on Systems, Man, and Cybernetics ({SMC})}, pp. 863--866, {ISSN}: 2577-1655. [Online]. Available: \url{https://ieeexplore.ieee.org/abstract/document/10394001}
\BIBentrySTDinterwordspacing

\bibitem{michaelson_terrain-relative_2024}
\BIBentryALTinterwordspacing
K.~A. Michaelson, F.~Wang, and R.~Zanetti
\newblock  Terrain-relative navigation with neuro-inspired elevation encoding \newblock  vol.~60, no.~3, pp. 3368--3378. [Online]. Available: \url{https://ieeexplore.ieee.org/document/10438923}
\BIBentrySTDinterwordspacing

\bibitem{jung_digital_2020}
\BIBentryALTinterwordspacing
Y.~Jung, S.~Lee, and H.~Bang
\newblock  Digital terrain map based safe landing site selection for planetary landing \newblock  vol.~56, no.~1, pp. 368--380. [Online]. Available: \url{https://ieeexplore.ieee.org/document/8701525}
\BIBentrySTDinterwordspacing

\bibitem{kim_autonomous_2004}
\BIBentryALTinterwordspacing
J.~Kim and S.~Sukkarieh
\newblock  Autonomous airborne navigation in unknown terrain environments \newblock  vol.~40, no.~3, pp. 1031--1045. [Online]. Available: \url{https://ieeexplore.ieee.org/document/1337472}
\BIBentrySTDinterwordspacing

\bibitem{bell_iii_calibration_2013}
J.~Bell~{III} \emph{et~al.}
\newblock  Calibration and performance of the mars reconnaissance orbiter context camera ({CTX}) \newblock  vol.~8, pp. 1--14.

\bibitem{tancik_nerfstudio_2023}
\BIBentryALTinterwordspacing
M.~Tancik \emph{et~al.}
\newblock  Nerfstudio: A modular framework for neural radiance field development \newblock  In \emph{Special Interest Group on Computer Graphics and Interactive Techniques Conference Conference Proceedings}, pp. 1--12. [Online]. Available: \url{http://arxiv.org/abs/2302.04264}
\BIBentrySTDinterwordspacing

\bibitem{wang_neus_2023}
\BIBentryALTinterwordspacing
P.~Wang, L.~Liu, Y.~Liu, C.~Theobalt, T.~Komura, and W.~Wang
\newblock  {NeuS}: Learning neural implicit surfaces by volume rendering for multi-view reconstruction \newblock . [Online]. Available: \url{http://arxiv.org/abs/2106.10689}
\BIBentrySTDinterwordspacing

\bibitem{dai_neural_2024}
\BIBentryALTinterwordspacing
A.~Dai, S.~Gupta, and G.~Gao
\newblock  Neural elevation models for terrain mapping and path planning \newblock . [Online]. Available: \url{http://arxiv.org/abs/2405.15227}
\BIBentrySTDinterwordspacing

\bibitem{beyer_ames_2018}
\BIBentryALTinterwordspacing
R.~A. Beyer, O.~Alexandrov, and S.~{McMichael}
\newblock  The ames stereo pipeline: {NASA}'s open source software for deriving and processing terrain data \newblock  vol.~5, no.~9, pp. 537--548, \_eprint: https://onlinelibrary.wiley.com/doi/pdf/10.1029/2018EA000409. [Online]. Available: \url{https://onlinelibrary.wiley.com/doi/abs/10.1029/2018EA000409}
\BIBentrySTDinterwordspacing

\bibitem{de_franchis_automatic_2014}
\BIBentryALTinterwordspacing
C.~de~Franchis, E.~Meinhardt-Llopis, J.~Michel, J.-M. Morel, and G.~Facciolo
\newblock  An automatic and modular stereo pipeline for pushbroom images \newblock  vol. {II}-3, pp. 49--56, conference Name: {ISPRS} Technical Commission {III} Symposium (Volume {II}-3) - 5\&ndash;7 September 2014, Zurich, Switzerland Publisher: Copernicus {GmbH}. [Online]. Available: \url{https://isprs-annals.copernicus.org/articles/II-3/49/2014/}
\BIBentrySTDinterwordspacing

\bibitem{schonberger_pixelwise_2016}
J.~L. Schönberger, E.~Zheng, M.~Pollefeys, and J.-M. Frahm
\newblock  Pixelwise view selection for unstructured multi-view stereo \newblock  In \emph{European Conference on Computer Vision ({ECCV})}.

\bibitem{schonberger_structure--motion_2016}
J.~L. Schönberger and J.-M. Frahm
\newblock  Structure-from-motion revisited \newblock  In \emph{Conference on Computer Vision and Pattern Recognition ({CVPR})}.

\bibitem{lee_extraction_2003}
\BIBentryALTinterwordspacing
H.-Y. Lee, T.~Kim, W.~Park, and H.~K. Lee
\newblock  Extraction of digital elevation models from satellite stereo images through stereo matching based on epipolarity and scene geometry \newblock  vol.~21, no.~9, pp. 789--796. [Online]. Available: \url{https://www.sciencedirect.com/science/article/pii/S0262885603000921}
\BIBentrySTDinterwordspacing

\bibitem{wang_epipolar_2011}
\BIBentryALTinterwordspacing
M.~Wang, F.~Hu, and J.~Li
\newblock  Epipolar resampling of linear pushbroom satellite imagery by a new epipolarity model \newblock  vol.~66, no.~3, pp. 347--355. [Online]. Available: \url{https://www.sciencedirect.com/science/article/pii/S0924271611000050}
\BIBentrySTDinterwordspacing

\bibitem{oh_piecewise_2010}
J.~Oh, W.~Lee, C.~Toth, D.~Grejner-Brzezinska, and C.~Lee
\newblock  A piecewise approach to epipolar resampling of pushbroom satellite images based on {RPC} \newblock  vol.~76, pp. 1353--1363.

\bibitem{zhang_ga-net_2019}
\BIBentryALTinterwordspacing
F.~Zhang, V.~Prisacariu, R.~Yang, and P.~H. Torr
\newblock  {GA}-net: Guided aggregation net for end-to-end stereo matching \newblock  In \emph{2019 {IEEE}/{CVF} Conference on Computer Vision and Pattern Recognition ({CVPR})}, pp. 185--194, {ISSN}: 2575-7075. [Online]. Available: \url{https://ieeexplore.ieee.org/document/8954424}
\BIBentrySTDinterwordspacing

\bibitem{gu_cascade_2020}
\BIBentryALTinterwordspacing
X.~Gu, Z.~Fan, Z.~Dai, S.~Zhu, F.~Tan, and P.~Tan
\newblock  Cascade cost volume for high-resolution multi-view stereo and stereo matching \newblock . [Online]. Available: \url{http://arxiv.org/abs/1912.06378}
\BIBentrySTDinterwordspacing

\bibitem{singh_cloud-gan_2018}
\BIBentryALTinterwordspacing
P.~Singh and N.~Komodakis
\newblock  Cloud-gan: Cloud removal for sentinel-2 imagery using a cyclic consistent generative adversarial networks \newblock  In \emph{{IGARSS} 2018 - 2018 {IEEE} International Geoscience and Remote Sensing Symposium}. {IEEE}, pp. 1772--1775. [Online]. Available: \url{https://ieeexplore.ieee.org/document/8519033/}
\BIBentrySTDinterwordspacing

\bibitem{pan_cloud_nodate}
H.~Pan
\newblock  Cloud removal for remote sensing imagery vai spatial attention generative adversarial network \newblock

\bibitem{wu_cloudformer_2022}
\BIBentryALTinterwordspacing
P.~Wu, Z.~Pan, H.~Tang, and Y.~Hu
\newblock  Cloudformer: A cloud-removal network combining self-attention mechanism and convolution \newblock  vol.~14, no.~23, p. 6132, number: 23 Publisher: Multidisciplinary Digital Publishing Institute. [Online]. Available: \url{https://www.mdpi.com/2072-4292/14/23/6132}
\BIBentrySTDinterwordspacing

\bibitem{christopoulos_cloudtran_2025}
\BIBentryALTinterwordspacing
D.~Christopoulos, V.~Ntouskos, and K.~Karantzalos
\newblock  {CloudTran}++: Improved cloud removal from multi-temporal satellite images using axial transformer networks \newblock  vol.~17, no.~1, p.~86, number: 1 Publisher: Multidisciplinary Digital Publishing Institute. [Online]. Available: \url{https://www.mdpi.com/2072-4292/17/1/86}
\BIBentrySTDinterwordspacing

\bibitem{alexandrov_multiview_2018}
\BIBentryALTinterwordspacing
O.~Alexandrov and R.~A. Beyer
\newblock  Multiview shape-from-shading for planetary images \newblock  vol.~5, no.~10, pp. 652--666, \_eprint: https://agupubs.onlinelibrary.wiley.com/doi/pdf/10.1029/2018EA000390. [Online]. Available: \url{https://onlinelibrary.wiley.com/doi/abs/10.1029/2018EA000390}
\BIBentrySTDinterwordspacing

\bibitem{langguth_shading-aware_2016}
F.~Langguth, K.~Sunkavalli, S.~Hadap, and M.~Goesele
\newblock  Shading-aware multi-view stereo \newblock  In \emph{Computer Vision – {ECCV} 2016}, B.~Leibe, J.~Matas, N.~Sebe, and M.~Welling, Eds. Springer International Publishing, pp. 469--485.

\bibitem{bosch_semantic_2019}
\BIBentryALTinterwordspacing
M.~Bosch, K.~Foster, G.~Christie, S.~Wang, G.~D. Hager, and M.~Brown
\newblock  Semantic stereo for incidental satellite images \newblock  In \emph{2019 {IEEE} Winter Conference on Applications of Computer Vision ({WACV})}. {IEEE}, pp. 1524--1532. [Online]. Available: \url{https://ieeexplore.ieee.org/document/8659180/}
\BIBentrySTDinterwordspacing

\bibitem{facciolo_automatic_2017}
\BIBentryALTinterwordspacing
G.~Facciolo, C.~De~Franchis, and E.~Meinhardt-Llopis
\newblock  Automatic 3d reconstruction from multi-date satellite images \newblock  In \emph{2017 {IEEE} Conference on Computer Vision and Pattern Recognition Workshops ({CVPRW})}. {IEEE}, pp. 1542--1551. [Online]. Available: \url{https://ieeexplore.ieee.org/document/8014932/}
\BIBentrySTDinterwordspacing

\bibitem{mildenhall_nerf_2020}
\BIBentryALTinterwordspacing
B.~Mildenhall, P.~P. Srinivasan, M.~Tancik, J.~T. Barron, R.~Ramamoorthi, and R.~Ng
\newblock  {NeRF}: Representing scenes as neural radiance fields for view synthesis \newblock . [Online]. Available: \url{http://arxiv.org/abs/2003.08934}
\BIBentrySTDinterwordspacing

\bibitem{muller_instant_2022}
\BIBentryALTinterwordspacing
T.~Müller, A.~Evans, C.~Schied, and A.~Keller
\newblock Instant neural graphics primitives with a multiresolution hash encoding. [Online]. Available: \url{https://arxiv.org/abs/2201.05989v2}
\BIBentrySTDinterwordspacing

\bibitem{lin_barf_2021}
\BIBentryALTinterwordspacing
C.-H. Lin, W.-C. Ma, A.~Torralba, and S.~Lucey
\newblock  {BARF}: Bundle-adjusting neural radiance fields \newblock . [Online]. Available: \url{http://arxiv.org/abs/2104.06405}
\BIBentrySTDinterwordspacing

\bibitem{martin-brualla_nerf_2021}
\BIBentryALTinterwordspacing
R.~Martin-Brualla, N.~Radwan, M.~S.~M. Sajjadi, J.~T. Barron, A.~Dosovitskiy, and D.~Duckworth
\newblock  {NeRF} in the wild: Neural radiance fields for unconstrained photo collections \newblock . [Online]. Available: \url{http://arxiv.org/abs/2008.02268}
\BIBentrySTDinterwordspacing

\bibitem{wang_neus2_2023}
\BIBentryALTinterwordspacing
Y.~Wang, Q.~Han, M.~Habermann, K.~Daniilidis, C.~Theobalt, and L.~Liu
\newblock  {NeuS}2: Fast learning of neural implicit surfaces for multi-view reconstruction \newblock . [Online]. Available: \url{http://arxiv.org/abs/2212.05231}
\BIBentrySTDinterwordspacing

\bibitem{logothetis_neural_2024}
\BIBentryALTinterwordspacing
F.~Logothetis, I.~Budvytis, and R.~Cipolla
\newblock  A neural height-map approach for the binocular photometric stereo problem \newblock  In \emph{2024 {IEEE}/{CVF} Winter Conference on Applications of Computer Vision ({WACV})}. {IEEE}, pp. 1557--1566. [Online]. Available: \url{https://ieeexplore.ieee.org/document/10484088/}
\BIBentrySTDinterwordspacing

\bibitem{derksen_shadow_2021}
\BIBentryALTinterwordspacing
D.~Derksen and D.~Izzo
\newblock  Shadow neural radiance fields for multi-view satellite photogrammetry \newblock . [Online]. Available: \url{http://arxiv.org/abs/2104.09877}
\BIBentrySTDinterwordspacing

\bibitem{mari_sat-nerf_2022}
\BIBentryALTinterwordspacing
R.~Mari, G.~Facciolo, and T.~Ehret
\newblock  Sat-{NeRF}: Learning multi-view satellite photogrammetry with transient objects and shadow modeling using {RPC} cameras \newblock  In \emph{2022 {IEEE}/{CVF} Conference on Computer Vision and Pattern Recognition Workshops ({CVPRW})}. {IEEE}, pp. 1310--1320. [Online]. Available: \url{https://ieeexplore.ieee.org/document/9857322/}
\BIBentrySTDinterwordspacing

\bibitem{mari_multi-date_2023}
\BIBentryALTinterwordspacing
R.~Marí, G.~Facciolo, and T.~Ehret
\newblock  Multi-date earth observation {NeRF}: The detail is in the shadows \newblock  In \emph{2023 {IEEE}/{CVF} Conference on Computer Vision and Pattern Recognition Workshops ({CVPRW})}. {IEEE}, pp. 2035--2045. [Online]. Available: \url{https://ieeexplore.ieee.org/document/10209050/}
\BIBentrySTDinterwordspacing

\bibitem{prosvetov_illuminating_2025}
\BIBentryALTinterwordspacing
A.~Prosvetov, A.~Govorov, M.~Pupkov, A.~Andreev, and V.~Nazarov
\newblock  Illuminating the moon: Reconstruction of lunar terrain using photogrammetry, neural radiance fields, and gaussian splatting \newblock  vol.~52, p. 100953. [Online]. Available: \url{https://linkinghub.elsevier.com/retrieve/pii/S2213133725000265}
\BIBentrySTDinterwordspacing

\bibitem{hansen_analyzing_2024}
\BIBentryALTinterwordspacing
M.~Hansen, C.~Adams, T.~Fong, and D.~Wettergreen
\newblock  Analyzing the effectiveness of neural radiance fields for geometric modeling of lunar terrain \newblock  In \emph{2024 {IEEE} Aerospace Conference}, pp. 1--12, {ISSN}: 1095-323X. [Online]. Available: \url{https://ieeexplore.ieee.org/abstract/document/10521163}
\BIBentrySTDinterwordspacing

\bibitem{van_kints_neural_2024}
\BIBentryALTinterwordspacing
E.~Van~Kints
\newblock  Neural radiance field methods for satellite imagery of polar climates \newblock  {ISBN}: 9798346895510. [Online]. Available: \url{https://www.proquest.com/docview/3152433988/abstract/E723FE7B79FA4F93PQ/1}
\BIBentrySTDinterwordspacing

\bibitem{li_nerfacc_2023}
\BIBentryALTinterwordspacing
R.~Li, M.~Tancik, and A.~Kanazawa
\newblock  {NerfAcc}: A general {NeRF} acceleration toolbox \newblock . [Online]. Available: \url{http://arxiv.org/abs/2210.04847}
\BIBentrySTDinterwordspacing

\bibitem{nichols_aster_1994}
\BIBentryALTinterwordspacing
D.~Nichols, A.~Kahle, H.~Fujisada, H.~Tsu, and Y.~Yamaguchi
\newblock  {ASTER} instrument design and science objectives \newblock  publisher: {UNKNOWN}. [Online]. Available: \url{https://ntrs.nasa.gov/citations/20210004760}
\BIBentrySTDinterwordspacing

\bibitem{kelvin_rodriguez_integrated_2024}
\BIBentryALTinterwordspacing
{Kelvin Rodriguez} and {Astrogeology Science Center}
\newblock  Integrated software for imagers and spectrometers ({ISIS}) 8.3.0 \newblock . [Online]. Available: \url{https://code.usgs.gov/astrogeology/isis/-/tags/8.3.0}
\BIBentrySTDinterwordspacing

\bibitem{paquette_abstraction_2023}
\BIBentryALTinterwordspacing
A.~C. Paquette
\newblock  Abstraction layer for ephemerides \newblock . [Online]. Available: \url{https://code.usgs.gov/astrogeology/ale/-/releases/0.9.0}
\BIBentrySTDinterwordspacing

\bibitem{calef_iii_msl_2016}
\BIBentryALTinterwordspacing
F.~J. Calef~{III} and T.~Parker
\newblock  {MSL} gale merged orthophoto mosaic \newblock . [Online]. Available: \url{https://astrogeology.usgs.gov/search/map/mars_msl_gale_merged_dem_1m}
\BIBentrySTDinterwordspacing

\bibitem{fergason_mars_2020}
\BIBentryALTinterwordspacing
R.~L. Fergason, D.~M. Galuszka, T.~M. Hare, D.~P. Mayer, and B.~L. Redding
\newblock  Mars 2020 terrain relative navigation {CTX} {DTM} mosaic \newblock . [Online]. Available: \url{https://astrogeology.usgs.gov/search/map/Mars/Mars2020/JEZ_ctx_B_soc_008_DTM_MOLAtopography_DeltaGeoid_20m_Eqc_latTs0_lon0}
\BIBentrySTDinterwordspacing

\bibitem{nasametiaistjapan_spacesystems_and_usjapan_aster_science_team_aster_2019}
\BIBentryALTinterwordspacing
{NASA/METI/AIST/Japan Spacesystems and U.S./Japan ASTER Science Team}
\newblock  {ASTER} global digital elevation model v003 \newblock . [Online]. Available: \url{https://www.earthdata.nasa.gov/data/catalog/lpcloud-astgtm-003}
\BIBentrySTDinterwordspacing

\bibitem{muller_tiny-cuda-nn_2021}
\BIBentryALTinterwordspacing
T.~Müller
\newblock  tiny-cuda-nn \newblock . [Online]. Available: \url{https://github.com/NVlabs/tiny-cuda-nn}
\BIBentrySTDinterwordspacing

\end{thebibliography}

\begin{IEEEbiography}[{\includegraphics*[width=1in,height=1.25in,keepaspectratio]{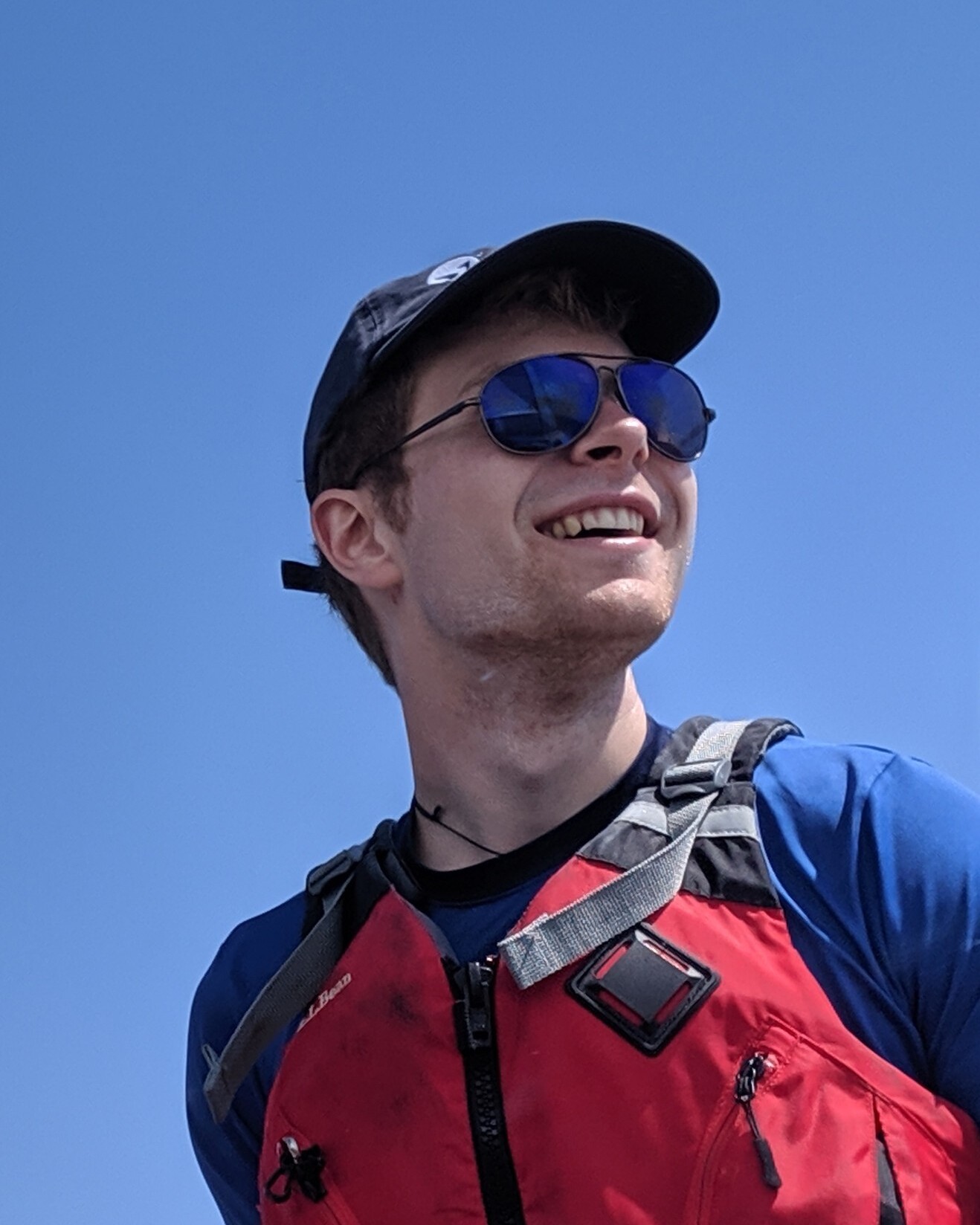}}]{Josef X. Biberstein}
  received his SB and SM degrees from MIT in the department of Aeronautics and Astronautics. He is currently in the doctoral program at MIT where his work focuses on control, autonomy, and neural inverse graphics with applications in the space domain. He has completed several prior internships at JPL and has contributed work to the Ingenuity and A-PUFFER missions. In his spare time, he is an avid outdoorsman.
\end{IEEEbiography}

\begin{IEEEbiography}[{\includegraphics*[width=1in,height=1.25in,keepaspectratio]{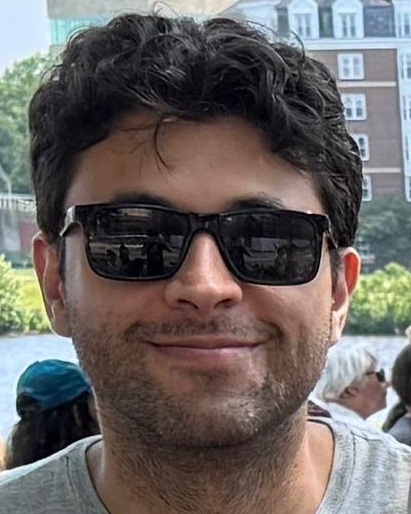}}]{Guilherme Cavalheiro}{\space}
  is a doctoral student at the department of Aeronautics and Astronautics at MIT, from where he also received his M.S. degree.
  With a background in Aeronautical Engineering from his B.S. degree from the Technological Institute of Aeronautics (ITA) in Brazil, his current interests lies in inverse graphics and generative diffusion, as well as their applications in robotics.
\end{IEEEbiography}

\begin{IEEEbiography}[{\includegraphics*[width=1in,height=1.25in,keepaspectratio]{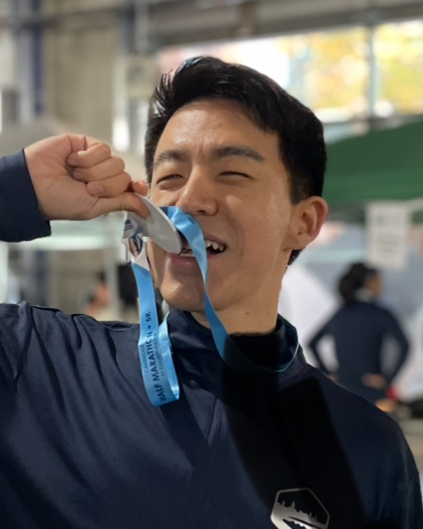}}]{Juyeop Han} received the B.S. degree in Mechanical Engineering from Seoul National University (SNU), Seoul, South Korea, in 2021.
  He also received the M.S. degree in Aerospace Engineering from Korea Advanced Institute of Science and Technology (KAIST), Daejeon, South Korea, in 2023.
  He is currently in the doctoral program at MIT where his work focuses on robot perception, motion planning, and uncertainty quantification for robotics.
\end{IEEEbiography}

\begin{IEEEbiography}[{\includegraphics*[width=1in,height=1.25in,keepaspectratio]{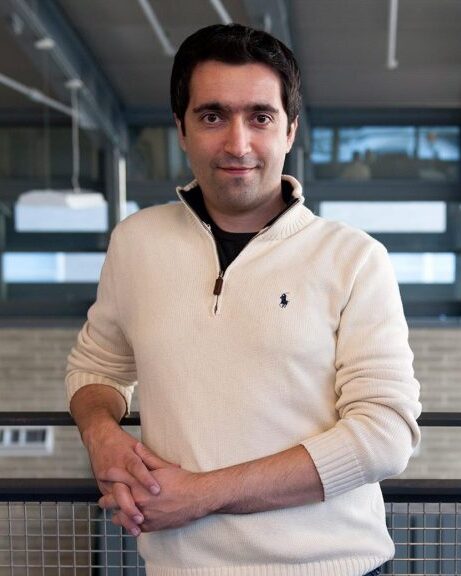}}]{Sertac Karaman} (Member, IEEE) received the S.M. degree in mechanical engineering and the Ph.D. degree in electrical engineering and computer science from the Massachusetts Institute of Technology (MIT), Cambridge, MA, USA, in 2009 and 2012, respectively.
 He is currently an Associate Professor of Aeronautics and Astronautics with MIT.
 He studies the applications of probability theory, stochastic processes, stochastic geometry, formal methods, and optimization for the design and analysis of high-performance cyber-physical systems.
 The application areas of his research include driverless cars, autonomous aerial vehicles, distributed aerial surveillance systems, air traffic control, certification and verification of control systems software, and many others.
 \par
 Dr. Karaman was a recipient of the IEEE Robotics and Automation Society Early Career Award in 2017, the Office of Naval Research Young Investigator Award in 2017, the Army Research Office Young Investigator Award in 2015, and the National Science Foundation Faculty Career Development (CAREER) Award in 2014.
\end{IEEEbiography}
\end{document}